\documentclass[acmsmall]{acmart}

\usepackage[normalem]{ulem}
 \usepackage{amsmath}
  \usepackage{multirow}
\usepackage{epstopdf}
  \usepackage{graphicx}
\usepackage{appendix}
\usepackage[normalem]{ulem}
\useunder{\uline}{\ul}{}
 \usepackage{enumitem}
 \useunder{\uline}{\ul}{}
\AtBeginDocument{%
  \providecommand\BibTeX{{%
    \normalfont B\kern-0.5em{\scshape i\kern-0.25em b}\kern-0.8em\TeX}}}
    
\usepackage{etoolbox}

\newcommand{\ms}[2]{{#1}\scriptsize{$\pm$#2}}
\newcommand{\msone}[2]{\bf {#1}\scriptsize{$\pm$#2}}
\newcommand{\mstwo}[2]{\underline{{#1}\scriptsize{$\pm$#2}}}
\usepackage[utf8]{inputenc} 
\usepackage[T1]{fontenc}    
\usepackage{hyperref}       
\usepackage{url}            
\usepackage{booktabs}       
\usepackage{amsfonts}       
\usepackage{nicefrac}       
\usepackage{microtype}      
\usepackage{xcolor}         
\usepackage{comment}
\usepackage{amsmath}

\usepackage{algorithm}
\usepackage{algorithmic,esint}
\usepackage{enumitem}
\usepackage{graphicx}
\usepackage{newfloat}
\usepackage{listings}
\usepackage{subfig}
\usepackage{adjustbox}
\usepackage{wrapfig}
\newtheorem{theorem}{Theorem}

\newtheorem{assumption}{Assumption}
\newtheorem{remark}[theorem]{Remark}

\newcommand{\model}{DIDA }
\newcommand{\modelnosp}{DIDA}
\newcommand{\modelp}{\textbf{I-DIDA} }
\newcommand{\modelpnosp}{\textbf{I-DIDA}}
\newcommand{\red}[1]{\textcolor{black}{#1}}
\newcommand{\ie}{{\it i.e.}}
\usepackage{tabularx}

  
\author{Zeyang Zhang}
\affiliation{%
  \institution{Tsinghua University}  \country{China}}
\email{zy-zhang20@mails.tsinghua.edu.cn}

\author{Xin Wang*}
\affiliation{%
  \institution{Tsinghua University}  \country{China}}
\email{xin\_wang@tsinghua.edu.cn}

\author{Ziwei Zhang}
\affiliation{%
  \institution{Tsinghua University}  \country{China}}
\email{zwzhang@tsinghua.edu.cn}

\author{Haoyang Li}
\affiliation{%
  \institution{Tsinghua University}  \country{China}}
\email{lihy18@mails.tsinghua.edu.cn}

\author{Wenwu Zhu*}
\affiliation{%
  \institution{Tsinghua University}  \country{China}}
\email{wwzhu@tsinghua.edu.cn}
\thanks{*Correspondence should be addressed to these authors.}

\begin{document}
\fancyhead{}

\title{Out-of-Distribution Generalized Dynamic Graph Neural Network with Disentangled Intervention and Invariance Promotion}

\begin{abstract}

Dynamic graph neural networks (DyGNNs) have demonstrated powerful predictive abilities by exploiting graph structural and temporal dynamics. 
However, the existing DyGNNs fail to handle distribution shifts, which naturally exist in dynamic graphs, mainly because the patterns exploited by DyGNNs may be variant with respect to labels under distribution shifts. 
In this paper, we propose \underline Disentangled \underline Intervention-based \underline Dynamic graph \underline Attention networks with \underline Invariance Promotion (\modelpnosp) to handle spatio-temporal distribution shifts in dynamic graphs by discovering and utilizing {\it invariant patterns}, \ie, structures and features whose predictive abilities are stable across distribution shifts. 
Specifically, we first propose a disentangled spatio-temporal attention network to capture the variant and invariant patterns. 
By utilizing the disentangled patterns, we design a spatio-temporal intervention mechanism to create multiple interventional distributions and an environment inference module to infer the latent spatio-temporal environments, and minimize the variance of predictions among these intervened distributions and environments, so that our model can make predictions based on invariant patterns with stable predictive abilities under distribution shifts.
Extensive experiments demonstrate the superiority of our method over state-of-the-art baselines under distribution shifts.
Our work is the first study of spatio-temporal distribution shifts in dynamic graphs, to the best of our knowledge.
\end{abstract}



\maketitle

\section{Introduction}
Dynamic graphs widely exist in real-world applications, including financial networks~\cite{nascimento2021dynamic,zhang2021dyngraphtrans}, social networks~\cite{berger2006framework,greene2010tracking}, 
traffic networks~\cite{peng2021dynamic,peng2020spatial},
{\it etc}. Distinct from static graphs, dynamic graphs can represent temporal structure and feature patterns, which are more complex yet common in reality. Besides the ubiquitous applications of Graph neural networks(GNNs) in various fields~\cite{cai2023user,chen2020neural,ma2023kr,yang2021hgat,zhang2022efraudcom,zitnik2018modeling,li2022graph,li2023preference,xie2021graph,wang2021combining,wang2022imbalanced, wang2022adagcl, bi2023predicting} due to their strong representation abilities of structural information, Dynamic graph neural networks (DyGNNs) have been recently proposed to further consider the time dimension and simultaneously tackle the highly complex structural and temporal information over dynamic graphs, which have achieved remarkable progress in 
many predictive tasks~\cite{skarding2021foundations,zhu2022learnable}. 

Nevertheless, the existing DyGNNs fail to handle spatio-temporal distribution shifts, which naturally exist in dynamic graphs for various reasons such as survivorship bias~\cite{brown1992survivorship}, selection bias~\cite{berk1983introduction,zhu2021shift}, trending~\cite{kim2021reversible}, {\it etc}. For example, in financial networks, external factors like period or market would affect the correlations between the payment flows and transaction illegitimacy~\cite{pareja2020evolvegcn}. Trends or communities also affect interaction patterns in coauthor networks~\cite{jin2021community} and recommendation networks~\cite{wang2022causal}. If DyGNNs highly rely on spatio-temporal patterns which are variant under distribution shifts, they will inevitably fail
to generalize well to the unseen test distributions.  

To address this issue, in this paper, we study the problem of handling spatio-temporal distribution shifts in dynamic graphs through discovering and utilizing {\it invariant patterns}, \ie, structures and features whose predictive abilities are stable across distribution shifts, which remain unexplored in the literature. 
However, this problem is highly non-trivial with the following challenges: 
\begin{itemize}[leftmargin = 0.5cm]
    \item How to discover the complex variant and invariant spatio-temporal patterns in dynamic graphs, which include both graph structures and node features varying through time?
    \item How to handle spatio-temporal distribution shifts in a principled manner with discovered variant and invariant patterns?
\end{itemize}

To tackle these challenges, \red{we propose a novel method named Disentangled Intervention-based Dynamic Graph Attention Networks with Invariance Promotion (\modelpnosp)}. Our proposed method handles distribution shifts well by discovering and utilizing invariant spatio-temporal patterns with stable predictive abilities.
Specifically, we first propose a disentangled spatio-temporal attention network to capture the variant and invariant patterns in dynamic graphs, which enables each node to attend to all its historic neighbors through a disentangled attention message-passing mechanism.
Then, inspired by causal inference literatures~\cite{pearl2000models,glymour2016causal}, we propose a spatio-temporal intervention mechanism to create multiple intervened distributions by sampling and reassembling variant patterns across neighborhoods and time, such that spurious impacts of variant patterns can be eliminated. 
To tackle the challenges that i) variant patterns are highly entangled across nodes and ii) directly generating and mixing up subsets of structures and features to do intervention is computationally expensive, we approximate the intervention process with summarized patterns obtained by the disentangled spatio-temporal attention network instead of original structures and features. 
Lastly, we propose an invariance regularization term to minimize prediction variance in multiple intervened distributions. \red{We further leverage variant patterns to enhance the invariance properties of the captured invariant patterns in the training process, by inferring the latent spatio-temporal environments and minimizing the prediction variance among these environments.} In this way, our model can capture and utilize invariant patterns with stable predictive abilities to make predictions under distribution shifts.
\red{Extensive experiments on one synthetic dataset and four real-world datasets, including node classification and link prediction tasks,  demonstrate the superiority of our proposed method over state-of-the-art baselines under distribution shifts.}
The contributions of our work are summarized as follows:

\begin{itemize}[leftmargin=0.5cm]
    \item \red{We propose Disentangled Intervention-based Dynamic Graph Attention Networks with Invariance Promotion (\modelpnosp)}, which can handle spatio-temporal distribution shifts in dynamic graphs. This is the first study of spatio-temporal distribution shifts in dynamic graphs, to the best of our knowledge.
    
    \item We propose a disentangled spatio-temporal attention network to capture variant and invariant graph patterns. We further design a spatio-temporal intervention mechanism to create multiple intervened distributions and an invariance regularization term based on causal inference theory to enable the model to focus on invariant patterns under distribution shifts. 
    
    \item \red{ We further promote the invariance property by minimizing the prediction variance among the latent environments inferred by the variant patterns.}
    
    \item \red{Experiments on one synthetic dataset and several real-world datasets demonstrate the superiority of our method over state-of-the-art baselines. }
    
\end{itemize}
\red{This manuscript is an extension of our paper published at NeurIPS 2022~\cite{zhang2022dynamic}. Compared with the conference version, we make significant contributions from the following aspects:}

\begin{itemize}
    \item \red{The newly proposed \modelp model is able to learn invariant patterns on dynamic graphs via enforcing sample-level and environment-level prediction invariance among the latent spatio-temporal patterns so as to improve the generalization ability of dynamic graph neural networks under spatio-temporal distribution shifts.}
    \item \red{The newly proposed environment-level invariance regularization can inherently boost the invariance property of the invariant patterns in the training process without adding extra time and memory complexity.}
    \item \red{\modelp jointly integrates spatio-temporal intervention mechanism and environment inference into a unified framework, so that the model can focus on invariant patterns to make predictions.}
    \item \red{More extensive experiments demonstrate that \modelp is able to show significant improvements over the state-of-the-art baseline methods and the original model proposed in the earlier conference paper.}
\end{itemize}

\red{The rest of this paper is organized as follows. We introduce the problem formulation and notations in Section \ref{sec:problem}. In Section \ref{sec:method}, we describe the details of our proposed framework. We present the experimental results in Section \ref{sec:exp} and review the related work in Section \ref{sec:related}. Finally, we conclude our work in Section \ref{sec:conclusion}. 
}

\begin{figure*}[t]
\centering
\includegraphics[width=1\linewidth]{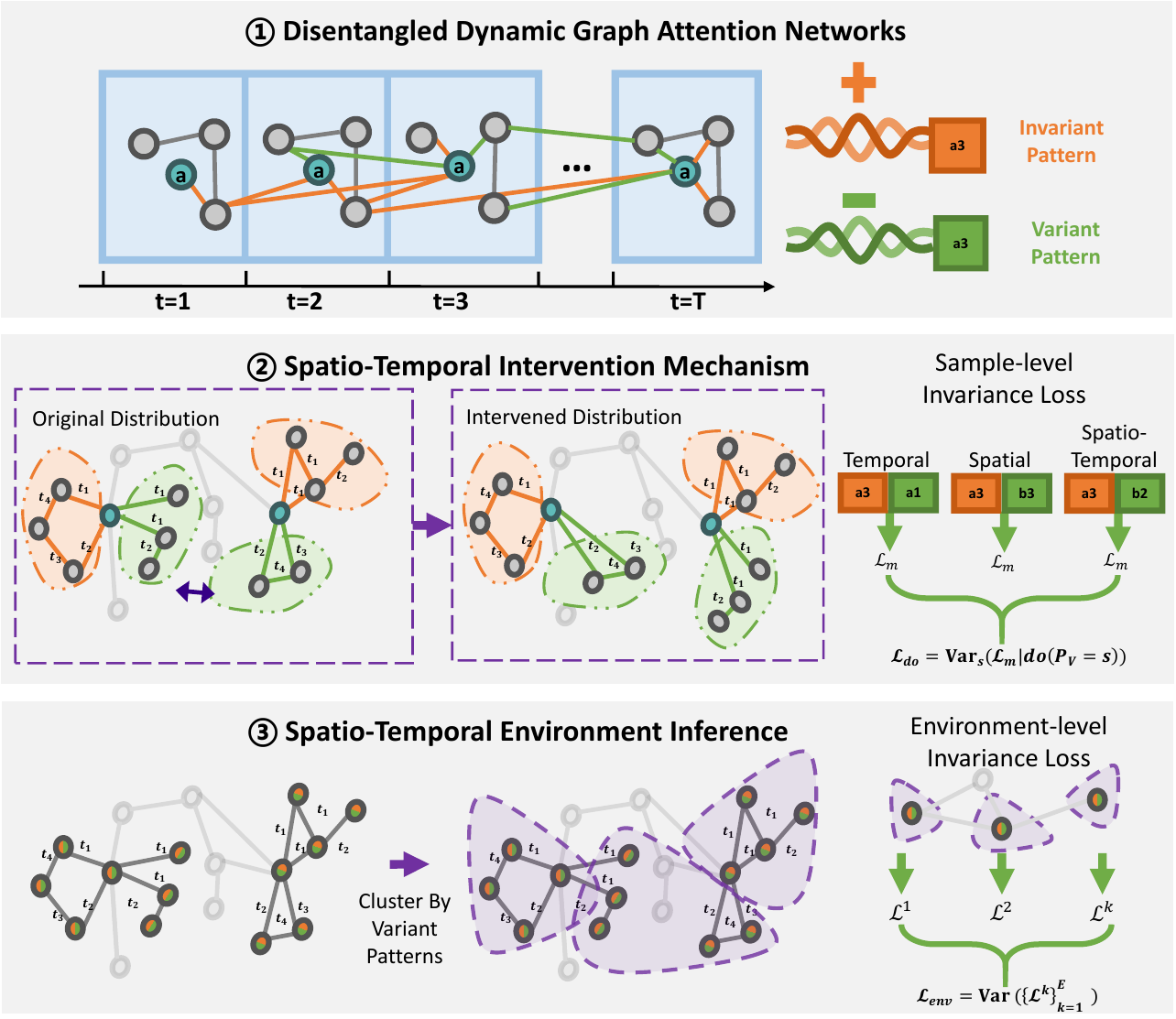}    
\caption{The framework of our proposed method \modelpnosp: 1. (Top) For a given dynamic graph with multiple timestamps, the disentangled dynamic graph attention networks first obtain summarizations of high-order invariant and variant patterns by disentangled spatio-temporal message passing. \red{2. (Middle) Then the spatio-temporal intervention mechanism creates multiple intervened distributions by sampling and reassembling variant patterns across space and time for each node. By utilizing the samples from the intervened distributions, the sample-level invariance loss is calculated to optimize the model so that it can focus on invariant patterns to make predictions. 3. (Bottom) Finally, the spatio-temporal environment inference module infers the environments by clustering the variant patterns, and an environment-level invariance loss is proposed to promote the invariance of the invariant patterns. In this way, the method can make predictions based on the invariant spatio-temporal patterns which have stable predictive abilities across distributions, and therefore handle the problem of distribution shifts on dynamic graphs. (Best viewed in color)}}
\label{fig:framework}
\end{figure*}

\section{Problem Formulation and Notations}\label{sec:problem}
In this section, we introduce the dynamic graph and prediction tasks, and formulate the problem of spatio-temporal distribution shift in dynamic graphs. \red{The notations adopted in this paper are summarized in Table \ref{tab:notation}.} 

\begin{table}
\caption{The summary of notations.}\label{tab:notation}
\centering
\adjustbox{max width = \linewidth}{
\begin{tabular}{c|l}
\toprule
\textbf{Notations}                                                  & \textbf{Descriptions}                                                                   \\ \midrule
$\mathcal{G}=(\mathcal{V},\mathcal{E})$                    & A graph with the node set and the edge set                                         \\
$\mathcal{G}^t=(\mathcal{V}^t,\mathcal{E}^t)$              & Graph slice at time $t$                                                        \\
$\mathcal{G}^{1:t},Y^t, \mathbf{G}^{1:t},\mathbf{Y}^t$     & The graph trajectory, label and their corresponding random variable across times                \\
$\mathcal{G}_v^{1:t},y^t, \mathbf{G}_v^{1:t},\mathbf{y}^t$ & Ego-graph trajectory, the node's label and their corresponding random variable \\
$f(\cdot),g(\cdot)$                                        & The predictor functions                                                                     \\
$P,\mathbf{P}$                                             & A pattern and its corresponding random variable                                  \\
$m(\cdot)$                                                 & A function to select structures and features from ego-graph trajectories         \\
do$(\cdot)$                                                & The do-calculus in causal inference                                                                    \\
$\phi(\cdot)$                                              & A function to find invariant patterns                                            \\
d                                                          & The dimensionality of node representation                                      \\
$\mathbf{q},\mathbf{k},\mathbf{v}$                         & The query, key, and value vector                                                    \\
$\mathcal{N}^t(u)$                                         & The dynamic neighborhood of node $u$ at time $t$                                            \\
$\mathbf{m}_I,\mathbf{m}_V,\mathbf{m}_f$                   & The structural mask of invariant and variant patterns, and the featural mask           \\
$\mathbf{z}_I^t(u),\mathbf{z}_V^t(u)$                      & Summarizations of invariant and variant patterns for node $u$ at time $t$      \\
Agg$_I(\cdot)$, Agg$_V(\cdot)$                             & Aggregation functions for invariant and variant patterns                       \\
$\mathbf{h}_u^t$                                           & Hidden embeddings for node $u$ at time $t$                                     \\
$\ell$                                                     & The loss function                                                                  \\
\red{$\mathcal{L},\mathcal{L}_m,\mathcal{L}_{do}$  }               & \red{The task loss, mixed loss, sample-level invariance loss}            
\\
\red{$\mathcal{L}^k,\mathcal{L}^{env}$  }               & \red{The $k$-th environment loss and the environment-level invariance loss}            
\\
\red{$K$}               & \red{The number of environments}
\\
\red{$\mathcal{K},k(u^t)$}               & \red{The environment set and the environment for the node $u$ at time $t$.}
\\
\bottomrule
\end{tabular}
}
\end{table}
\subsection{Dynamic Graph}
Dynamic Graph. Consider a graph $\mathcal{G}$ with the node set $\mathcal{V}$ and the edge set $\mathcal{E}$. A dynamic graph can be defined as 
$\mathcal{G}=(\{\mathcal{G}^{t}\}_{t=1}^{T})$, 
where $T$ is the number of time stamps, $\mathcal{G}^t=(\mathcal{V}^t,\mathcal{E}^t)$ is the graph slice at time stamp $t$,  $\mathcal{V}=\bigcup_{t=1}^{T} \mathcal{V}^{t}$, $\mathcal{E}=\bigcup_{t=1}^{T} \mathcal{E}^{t}$. We use $\mathbf{G}^t$ to denote a random variable of $\mathcal{G}^t$.
\subsection{Prediction Tasks}
For dynamic graphs, the prediction task can be summarized as using past graphs to make predictions, \ie
$p(\mathbf{Y}^{t}|\mathbf{G}^1,\mathbf{G}^2,\dots,\mathbf{G}^t) = p(\mathbf{Y}^{t}|\mathbf{G}^{1:t})$
, where label $\mathbf{Y}^t$ can be node properties or occurrence of links between nodes at time $t+1$. In this paper, we mainly focus on node-level tasks, which are commonly adopted in dynamic graph literatures~\cite{skarding2021foundations,zhu2022learnable}. Following ~\cite{wu2022handling,huang2020graph}, we factorize the distribution of graph trajectory into ego-graph trajectories, \ie 
$p(\mathbf{Y}^t \mid \mathbf{G}^{1:t})=\prod_v p(\mathbf{y}^t \mid \mathbf{G}_v^{1:t})$. 
An ego-graph induced from node $v$ at time $t$ is composed of the adjacency matrix including all edges in node $v$'s $L$-hop neighbors at time $t$, \ie, $\mathcal{N}^t_v$, and the features of nodes in $\mathcal{N}^t_v$.
The optimization objective is to learn an optimal predictor with empirical risk minimization.
\begin{equation}
\min_\theta \mathbb{E}_{(y^t,\mathcal{G}_v^{1:t}) \sim p_{tr}(\mathbf{y}^t,\mathbf{G}_v^{1:t})} \mathcal{L}(f_\theta(\mathcal{G}_v^{1:t}),y^{t}),
\label{eq:objective}
\end{equation}
where $f_\theta$ is a learnable dynamic graph neural networks,
We use $\mathbf{G}^{1:t}_v$,$\mathbf{y}^t$ to denote the random variable of the ego-graph trajectory and its label, and $\mathcal{G}^{1:t}_v$,$y^t$ refer to the respective instances.

\subsection{Spatio-Temporal Distribution Shift}
However, the optimal predictor trained with the training distribution may not generalize well to the test distribution when there exists a distribution shift problem.
In the literature of dynamic graphs, researchers are devoted to capturing laws of network dynamics which are stable in systems~\cite{wang2021inductive,qiu2020temporal,huang2020motif,zhou2018dynamic,trivedi2019dyrep}. Following them, we assume the conditional distribution is the same $p_{tr}(\mathbf{Y}^t|\mathbf{G}^{1:t}) = p_{te}(\mathbf{Y}^t|\mathbf{G}^{1:t})$, and only consider the covariate shift problem where $p_{tr}(\mathbf{G}^{1:t}) \neq p_{te}(\mathbf{G}^{1:t})$. Besides the temporal distribution shift which naturally exists in time-varying data~\cite{gagnon2022woods,kim2021reversible,du2021adarnn,venkateswaran2021environment,lu2021diversify} and the structural distribution shift in non-euclidean data~\cite{wu2022discovering,wu2022handling,ding2021closer}, there exists a much more complex spatio-temporal distribution shift in dynamic graphs. For example, the distribution of ego-graph trajectories may vary across periods or communities.

\section{Methodologies}\label{sec:method}
\red{In this section, we introduce our Disentangled Intervention-based Dynamic Graph Attention Networks with Invariance Promotion (\modelpnosp) to handle spatio-temporal distribution shift in dynamic graphs.} First, we propose a disentangled dynamic graph attention network to extract invariant and variant spatio-temporal patterns. \red{Then we propose a spatio-temporal intervention mechanism to create multiple intervened data distributions, coupled with an invariance loss to minimize the prediction variance among intervened distributions. Finally, we propose an environmental invariance regularization to promote the quality of invariant patterns, and optimize the model with both invariance regularizations to encourage the model to rely on invariant patterns to make predictions.}

\subsection{Handling Spatio-Temporal Distribution Shift}
\subsubsection{Spatio-Temporal Pattern}
In recent decades of development of dynamic graphs, some scholars endeavor to conclude insightful patterns of network dynamics to reflect how real-world networks evolve through time~\cite{kovanen2011temporal,benson2016higher,paranjape2017motifs,zitnik2019evolution}. For example, the laws of triadic closure describe that two nodes with common neighbors (patterns) tend to have future interactions in social networks~\cite{coleman1994foundations,huang2015triadic,zhou2018dynamic}. Besides structural information, node attributes are also an important part of the patterns, e.g., social interactions can be also affected by gender and age ~\cite{kovanen2013temporal}. Instead of manually concluding patterns, we aim at learning the patterns using DyGNNs so that the more complex spatio-temporal patterns with mixed features and structures can be mined in dynamic graphs. Therefore, we define the spatio-temporal pattern used for node-level prediction as a subset of ego-graph trajectory,
\begin{equation}
P^t(v)=m_v^t(\mathcal{G}^{1:t}_v),
\end{equation}
where $m^t_v(\cdot)$ selects structures and attributes from the ego-graph trajectory. In ~\cite{zhou2018dynamic}, the pattern can be explained as an open triad with similar neighborhood, and the model tends to make link predictions to close the triad with $\hat{y}_{u,v}^{t}=f_\theta(P^t(u),P^t(v))$ based on the laws of triadic closure~\cite{simmel1950sociology}. DyGNNs aim at exploiting predictive spatio-temporal patterns to boost prediction ability. However, the predictive power of some patterns may vary across periods or communities due to spatio-temporal distribution shift. Inspired by the causal theory~\cite{pearl2000models,glymour2016causal}, we make the following assumption.

\begin{assumption} For a given task, there exists a predictor $f(\cdot)$, for samples ($\mathcal{G}^{1:t}_v$,$y^{t}$) from any distribution, there exists an invariant pattern $P^t_I(v)$ and a variant pattern $P^t_V(v)$  such that $y^t=f(P^t_I(v))+\epsilon$ and $P^t_I(v) = \mathcal{G}_v^{1:t} \backslash P^t_V(v)$, \ie, $\mathbf{y}^t \perp \mathbf{P}^t_V(v) \mid \mathbf{P}^t_I(v)$. 
\label{assum}
\end{assumption}

In the Assumption~\ref{assum}, $P^t_I(v) = \mathcal{G}_v^{1:t} \backslash P^t_V(v)$ denotes that the dynamic graph is composed of the invariant patterns and variant patterns. The assumption shows that invariant patterns $\mathbf{P}^t_I(v)$ are sufficiently predictive for label $y^t$ and can be exploited across periods and communities without adjusting the predictor, while the influence of variant patterns $\mathbf{P}^t_V(v)$ on $\mathbf{y}^t$ is shielded by the invariant patterns. 

\subsubsection{Training Objective}
Our main idea is that to obtain better generalization ability, the model should rely on invariant patterns instead of variant patterns, as the former is sufficient for prediction while the predictivity of the latter could be variant under distribution shift. Along this, our objective can be transformed to
\begin{equation}
\begin{aligned}
\min_{\theta_1,\theta_2} \mathbb{E}_{(y^{t},\mathcal{G}_v^{1:t}) \sim p_{tr}(\mathbf{y}^{t},\mathbf{G}_v^{1:t})} \mathcal{L}(f_{\theta_1}(\tilde{P}_I^t(v)),y^{t}) \\ s.t \quad \phi_{\theta_2}(\mathcal{G}_v^{1:t})=\tilde{P}_I^t(v), \mathbf{y}^t \perp \tilde{\mathbf{P}}^t_V(v) \mid \tilde{\mathbf{P}}^t_I(v),
\end{aligned}
\label{eq:objective2}
\end{equation}
where $f_{\theta_1}(\cdot)$ make predictions based on the invariant patterns, $\phi_{\theta_2}(\cdot)$ aims at finding the invariant patterns. \red{However, the objective is challenging due to 1) the invariant and variant patterns are not labeled, and the model should be optimized to distinguish these patterns, 2) the properties of invariance and sufficiency should be achieved by specially designed mechanisms so that the model can rely on invariant patterns to make accurate predictions under distribution shifts. To this end, we propose two invariance loss from two levels for guiding the model to find and rely on invariant patterns, which are respectively inspired by the causal theory and invariant learning literature.} 
\subsubsection{Sample-Level Invariance Loss}
By causal theory~\cite{pearl2000models,glymour2016causal}, Eq.~\eqref{eq:objective2} can be transformed into 
\begin{equation}
\begin{aligned}
\min_{\theta_1,\theta_2} \mathbb{E}_{(y^{t},\mathcal{G}_v^{1:t}) \sim p_{tr}(\mathbf{y}^{t},\mathbf{G}_v^{1:t})} \mathcal{L}(f_{\theta_1}(\phi_{\theta_2}(\mathcal{G}^{1:t}_v)),y^{t}) + \\
\lambda \text{Var}_{s\in \mathcal{S} }(\mathbb{E}_{(y^{t},\mathcal{G}_v^{1:t}) \sim p_{tr}(\mathbf{y}^{t},\mathbf{G}_v^{1:t}|\text{do}(\mathbf{P}^t_V=s))}\mathcal{L}(f_{\theta_1}(\phi_{\theta_2}(\mathcal{G}^{1:t}_v)),y^{t})),
\end{aligned}
\label{eq:objective3}
\end{equation}
where `\texttt{do}' denotes do-calculas to intervene the original distribution~\cite{tian2006characterization,glymour2016causal}, $\mathcal{S}$ denotes the intervention set and $\lambda$ is a balancing hyperparameter. The idea can be informally described that as in Eq.~\eqref{eq:objective2}, variant patterns $\mathbf{P}^t_V$ have no influence on the label $\mathbf{y}^t$ given the invariant patterns  $\mathbf{P}^t_I$, then the prediction would not be varied if we intervene the variant patterns and keep invariant patterns untouched. \red{As this loss intervenes the distributions in the sample-level (\ie, nodes), and pursues the invariance of the invariant patterns for each sample, we name the variance term in Eq.~\eqref{eq:objective3} as sample-level invariance loss.}
\begin{remark} \label{remark}
Minimizing the variance term in Eq.~\eqref{eq:objective3} help the model to satisfy the constraint of $\mathbf{y}^t \perp \tilde{\mathbf{P}}^t_V(v) \mid \tilde{\mathbf{P}}^t_I(v)$ in Eq.~\eqref{eq:objective2}, \ie, $p(\mathbf{y}^t \mid \tilde{\mathbf{P}}^t_I(v),\tilde{\mathbf{P}}^t_V(v) ) = p(\mathbf{y}^t \mid \tilde{\mathbf{P}}^t_I(v))$.
\end{remark}

\subsubsection{Environment-Level Invariance Loss}
\red{ Invariant learning~\cite{arjovsky2019invariant,sagawa2019distributionally,krueger2021out} is a promising research direction with the goal of empowering the model with invariant predictive abilities under distribution shifts. Environments, commonly as a critical concept for the method assumption and design in the invariant learning literature, refer to where the observed instances are sampled from, which may have variant correlations with labels. In road networks, for example, two traffic jams in different places and times may happen simultaneously by chance or there can be causal relations, e.g., the road structure let one traffic jam block other roads and inevitably lead to another traffic jam. In this case, places and times may act as the environments which may have spurious correlations with labels and should not be exploited by the model under distribution shifts. Inspired by invariant learning, we propose to promote the invariance property of the invariant patterns by designing an environment-level invariance loss,} 
\begin{equation}
\text{Var}_{k\in \mathcal{K} }(\mathbb{E}_{(y^{t},\mathcal{G}_v^{1:t}) \sim p_{tr}(\mathbf{y}^{t},\mathbf{G}_v^{1:t}|k)}\mathcal{L}(f_{\theta_1}(\phi_{\theta_2}(\mathcal{G}^{1:t}_v)),y^{t})),
\end{equation}
\red{where $k$ denotes the $k$-th environment from the environment set $\mathcal{K}$, and $p_{tr}(\mathbf{y}^t,\mathbf{G}_v^{1:t}|k)$ denotes the data distribution of the $k$-th environment. Intuitively, minimizing the environment-level invariance loss encourages the model to make stable predictions regardless of the environments. Together with the sample-level invariance loss and environment-level invariance loss, we can help the model discover the invariant and variant patterns, and rely on invariant patterns to make predictions. We will describe how to implement these insights in an end-to-end manner in the following sections.}

\subsection{Disentangled Dynamic Graph Attention Networks}
\label{sec:ddgat}
\subsubsection{Dynamic Neighborhood}
To simultaneously consider the spatio-temporal information, we define the dynamic neighborhood as $\mathcal{N}^{t}(u)=\{v: (u,v) \in \mathcal{E}^t\}$, which includes all nodes that have interactions with node $u$ at time $t$. \red{ For node $u$ at time $t_1$, the dynamic neighborhoods $\mathcal{N}^{t}(u), t\le t_1$ describe the historical structural information of $u^t$, which enables different views of historical structural information based on the current time, e.g., $u^{t_2}$ and $u^{t_3}$ may aggregate different messages from $\mathcal{N}^{t_1}(u)$ for $t_1\le t_2\le t_3$. For example, the interest of the same user may have evolved through time, and the messages, even from the same neighborhood, adopted by the user to conduct transactions also vary. The model should be designed to be aware of these evolving patterns in the dynamic neighborhood. Note that the defined dynamic neighborhood includes only 1-order spatial neighbors at time $t$ for the brevity of notations, while the concept of n-order neighbors can be extended by considering the neighbors which can be reached by n-hop paths. Following classical message passing networks, we take into consideration the information of the n-order neighborhood by stacking multiple layers for message passing and aggregation. } 

\subsubsection{Disentangled Spatio-temporal Graph Attention Layer}
To capture spatio-temporal patterns for each node, we propose a spatio-temporal graph attention to enable each node to attend to its dynamic neighborhood simultaneously. 
For a node $u$ at time stamp $t$ and its neighbors $v \in \mathcal{N}^{t'}(u), \forall t'\leq t$, we calculate the Query-Key-Value vectors as
\begin{equation}
\begin{aligned}
\mathbf{q}_u^t&=\mathbf{W}_q\Bigl(\mathbf{h}_u^t || \text{TE}(t)\Bigr),\\
\mathbf{k}_v^{t'}&=\mathbf{W}_k\Bigl(\mathbf{h}_v^{t'} || \text{TE}(t')\Bigr),\\
\mathbf{v}_v^{t'}&=\mathbf{W}_v\Bigl(\mathbf{h}_v^{t'} || \text{TE}(t')\Bigr),\\
\end{aligned}
\end{equation}
where $\mathbf{h}^t_u$ denotes the representation of node $u$ at the time stamp $t$, $\mathbf{q}$, $\mathbf{k}$, $\mathbf{v}$ represents the query, key and value vector, respectively, and we omit the bias term for brevity. For simplicity of notations, the vectors in this paper are represented as row vectors. TE($t$) denotes the temporal encoding techniques to obtain embeddings of time $t$ so that the time of link occurrence can be considered inherently ~\cite{xu2020inductive,rossi2020temporal}. Then, we can calculate the attention scores among nodes in the dynamic neighborhood to obtain the structural masks,
\begin{equation}
\begin{aligned}
\mathbf{m}_I&=\text{Softmax}(\frac{\mathbf{q}\cdot \mathbf{k}^T}{\sqrt{d}}),\\
\mathbf{m}_V&=\text{Softmax}(-\frac{\mathbf{q}\cdot \mathbf{k}^T}{\sqrt{d}}),\\
\end{aligned}
\end{equation}
where $d$ denotes feature dimension, $\mathbf{m}_I$ and $\mathbf{m}_V$ represent the masks of invariant and variant structural patterns. In this way, dynamic neighbors with higher attention scores in invariant patterns will have lower attention scores in variant ones, which means the invariant and variant patterns have a negative correlation. To capture invariant featural pattern, we adopt a learnable featural mask $\mathbf{m}_f= \text{Softmax}(\mathbf{w}_f)$ to 
select features from the messages of dynamic neighbors. Then the messages of the dynamic neighborhood can be summarized with respective masks,
\begin{equation}
\begin{aligned}
\mathbf{z}^t_I(u)&=\text{Agg}_I(\mathbf{m}_{I},\mathbf{v}\odot \mathbf{m}_f), \\
\mathbf{z}^t_V(u)&=\text{Agg}_V(\mathbf{m}_{V},\mathbf{v}),
\end{aligned}
\end{equation}
where Agg$(\cdot)$ denotes aggregating and summarizing messages from the dynamic neighborhood. To further disentangle the invariant and variant patterns, we design different aggregation functions $\text{Agg}_I(\cdot)$ and $\text{Agg}_V(\cdot)$ to summarize specific messages from masked dynamic neighborhood respectively. Then the pattern summarizations are added up as hidden embeddings to be fed into subsequent layers,
\begin{equation}
    \mathbf{h}^t_u \leftarrow \mathbf{z}^t_I(u)+\mathbf{z}^t_V(u).
\end{equation}
\subsubsection{Overall Architecture} 
The overall architecture is a stacking of spatio-temporal graph attention layers. Like classic graph message-passing networks, this enables each node to access high-order dynamic neighborhood indirectly, where $\mathbf{z}^t_I(u)$ and $\mathbf{z}^t_V(u)$ at $l$-th layer can be a summarization of invariant and variant patterns in $l$-order dynamic neighborhood. In practice, the attention can be easily extended to multi-head attention~\cite{vaswani2017attention} to stable the training process and model multi-faceted graph evolution~\cite{sankar2020dysat}.
\subsection{Spatio-Temporal Intervention Mechanism}
\label{sec:intervener}
\subsubsection{Direct Intervention}
One way of intervening the distribution of the variant pattern as Eq.~\eqref{eq:objective3} is directly generating and altering the variant patterns.
However, this is infeasible in practice due to the following reasons: 
First, since it has to intervene the dynamic neighborhood and features node-wisely, the computational complexity is unbearable. Second, generating variant patterns including time-varying structures and features is another intractable problem. 

\subsubsection{Approximate Intervention} 
To tackle the problems mentioned above, we propose to approximate the patterns $\mathbf{P}^t$ with summarized patterns $\mathbf{z}^t$ found in Sec.~\ref{sec:ddgat}. As $\mathbf{z}^t_I(u)$ and $\mathbf{z}^t_V(u)$ act as summarizations of invariant and variant spatio-temporal patterns for node $u$ at time $t$, we approximate the intervention process by sampling and replacing the variant pattern summarizations instead of altering original structures and features with generated ones. 
To do spatio-temporal intervention, we collect variant patterns of all nodes at all time, from which we sample one variant pattern to replace the variant patterns of other nodes across time. For example, we can use the variant pattern of node $v$ at time $t_2$ to replace the variant pattern of node $u$ at time $t_1$ as 
\begin{equation}
    \mathbf{z}^{t_1}_I(u),\mathbf{z}^{t_1}_V(u)\leftarrow \mathbf{z}^{t_1}_I(u),\mathbf{z}^{t_2}_V(v).
    \label{eq:intervene_node}
\end{equation}
As the invariant pattern summarization is kept the same, the label should not be changed.
Thanks to the disentangled spatio-temporal graph attention, we get variant patterns across neighborhoods and time, which can act as natural intervention samples inside data so that the complexity of the generation problem can also be avoided. 
By doing Eq.~\eqref{eq:intervene_node} multiple times, we can obtain multiple intervened data distributions for the subsequent optimization. 

\subsection{Spatio-Temporal Environment Inference}
\label{sec:env_inference}
\red{It is challenging to obtain environment labels on dynamic graphs, since the environments on dynamic graphs are complex that include spatio-temporal information and may also vary by periods or communities. For these reasons, environment labels are not available on dynamic graphs in practice. To tackle this problem, we introduce the spatio-temporal environment inference module in this section.}

\red{Recall that in Sec.~\ref{sec:ddgat}, we obtain the summarized invariant and variant spatio-temporal patterns $\mathbf{z}^t_I$ and $\mathbf{z}^t_V$, which can be further exploited to infer the environment labels $k(u^t)$ for each node $u$ at time $t$.  Since the invariant patterns capture the invariant relationships between predictive ego-graph trajectories and labels, the variant patterns in turn capture variant correlations under different distributions, which could be helpful for discriminating spatio-temporal environments. Inspired by ~\cite{liu2021heterogeneous, li2022gil}, we utilize the variant patterns to infer the latent environments. Specifically, to infer the node environment labels $\mathbf{K} \in \mathcal{K}^{N\times T}$, we adopt an off-the-shelf clustering algorithm K-means in this paper, while other more sophisticated clustering methods can be easily incorporated,}
\begin{equation}
\mathbf{K} = \text{K-means}([\mathbf{Z}^1_V,\mathbf{Z}^2_V,\dots,\mathbf{Z}^T_V]),
\label{eq:kmeans}
\end{equation}
\red{where $k(u^t) \in \mathcal{K}$ denote the corresponding environment label for each node $u$ at time $t$, $\mathcal{K}$=\{0,1,\dots,$K$\} denotes the set of $K$ environments, and $K$ is a hyperparameter that reflects the assumption of the number of the environments. Using $\mathbf{K}$, we can partition the nodes at different time on dynamic graphs into multiple training environments. Note that the spatio-temporal environment inference module is unsupervised without any ground-truth environment labels, which is more practical on real-world dynamic graphs.}

\subsection{Optimization with Invariance Loss}
\subsubsection{Sample-Level Invariance Loss}
Based on the multiple intervened data distributions with different variant patterns, we can next optimize the model to focus on invariant patterns to make predictions.
Here, we introduce invariance loss to instantiate Eq.~\eqref{eq:objective3}. 
Let $\mathbf{z}_I$ and $\mathbf{z}_V$ be the summarized invariant and variant patterns, we calculate the task loss by only using the invariant patterns 
\begin{equation}
\mathcal{L}=\ell( f(\mathbf{z}_I),\mathbf{y}),
\label{eq:invariant}
\end{equation}
where $f(\cdot)$ is the predictor. The task loss let the model utilize the invariant patterns to make predictions. Then we calculate the mixed loss as 
\begin{equation}
\mathcal{L}_m=\ell( g(\mathbf{z}_V,\mathbf{z}_I),\mathbf{y}),
\label{eq:mix}
\end{equation}
where another predictor $g(\cdot)$ makes predictions using both invariant patterns $\mathbf{z}_V$ and variant patterns $\mathbf{z}_I$. The mixed loss measures the model's prediction ability when variant patterns are also exposed to the model.
Then the invariance loss is calculated by 
\begin{equation}
\mathcal{L}_{do}=\text{Var}_{s_i \in \mathcal{S}}(\mathcal{L}_m|\text{do}(\mathbf{P}^t_V=s_i)),
\label{eq:do}
\end{equation}
where `do' denotes the intervention mechanism as mentioned in Section ~\ref{sec:intervener}. The invariance loss measures the variance of the model's prediction ability under multiple intervened distributions. 

\subsubsection{Environment-Level Invariance Loss}
\red{After obtaining the environment labels by the spatio-temporal environment inference module in Sec. \ref{sec:env_inference}, we have the samples from different environments and the loss of the $k$-th environment is calculated by}
\begin{equation}
\mathcal{L}^{k} = \ell(f(\{\mathbf{z}^t_I(u):k(u^t)=k\},\mathbf{y}),
\label{eq:loss_env}
\end{equation}
\red{and the environment-level invariance loss can be calculated by}
\begin{equation}
\label{eq:loss_env_all}
\mathcal{L}_{env} = \text{Var}(\{\mathcal{L}^k\}_{k=1}^K).
\end{equation}

\red{In this way, minimizing the variance term encourages the invariance of the model predictions among different environments, which potentially reduces the effects of spurious correlations that may be caused by the spatio-temporal environments under distribution shifts. }

\subsubsection{Overall Training Objective}
The final training objective is
\begin{equation}
\label{eq:loss}
\min_{\theta} \mathcal{L} +  \lambda_{do} \mathcal{L}_{do} + \lambda_{e} \mathcal{L}_{env},
\end{equation}
\red{where the task loss $\mathcal{L}$ is minimized to exploit invariant patterns, while the sample-level invariance loss $\mathcal{L}_{do}$ and environment-level invariance loss $\mathcal{L}_{env}$ help the model to discover invariant and variant patterns, and $\lambda_{do}$ and $\lambda_{e}$ are hyperparameters to balance between two objectives.}
After training, we only adopt invariant patterns
to make predictions in the inference stage. The overall algorithm is summarized in Algorithm \ref{algo:pipeline}.

\begin{algorithm}
\caption{Training pipeline for \modelpnosp} 
\label{algo:pipeline}
\begin{algorithmic}[1]
\REQUIRE 
Training epochs $L$, number of intervention samples $S$, \red{number of environments $K$, hyperparameters $\lambda_{do}$ and $\lambda_{e}$.}
\FOR{$l = 1, \dots, L$}
    \STATE Obtain $\mathbf{z}^t_V,\mathbf{z}^t_I$ for each node and time as described in Section \ref{sec:ddgat}
    \STATE Calculate task loss and mixed loss as Eq. \eqref{eq:invariant} and Eq. \eqref{eq:mix}
    \STATE Sample $S$ variant patterns from collections of $\mathbf{z}^t_V$, to construct intervention set $\mathcal{S}$
    \FOR{$s$ in $\mathcal{S}$}
        \STATE Replace the nodes' variant pattern summarizations with $s$ as Section \ref{sec:intervener}
        \STATE Calculate mixed loss as Eq.  \eqref{eq:mix}
    \ENDFOR
    \STATE Calculate the sample-level invariance loss as Eq. \eqref{eq:do}
    \STATE \red{Infer the environment labels as Eq. \eqref{eq:kmeans}}
    \FOR{$k$ = 1, \dots, $K$}
        \STATE \red{Calculate the $k$-th environment loss as Eq. \eqref{eq:loss_env}}
    \ENDFOR
    \STATE \red{Calculate the environment-level invariance loss as Eq. \eqref{eq:loss_env_all}}
    \STATE Update the model according to Eq. \eqref{eq:loss}
\ENDFOR
\end{algorithmic}
\end{algorithm}

\subsection{Discussions}
\subsubsection{Complexity Analysis}
We analyze the computational complexity of \modelp as follows. 

Denote $|V|$ and $|E|$ as the total number of nodes and edges in the graph, respectively, and $d$ as the dimensionality of the hidden representation. The spatio-temporal aggregation has a time complexity of $O(|E|d+|V|d^2)$. The disentangled component adds a constant multiplier $2$, which does not affect the time complexity of aggregation. Denote $|E_p|$ as the number of edges to predict and $|S|$ as the size of the intervention set. Denote $K$ as the number of environments, $T$ as the number of iterations for the K-means algorithm.  Our intervention mechanism has a time complexity of $O(|E_p||S|d)$ and the environment inference module has a time complexity of $O(K|V|Td)$ in training. Moreover, these modules do not put extra time complexity in inference, since they are only adopted in the training state. 

Therefore, the overall time complexity of \modelp is $O(|E|d+|V|d^2 + |E_p||S|d + K|V|Td )$. Notice that $|S|$ is a hyper-parameter and is usually set as a small constant. In summary, \modelp has a linear time complexity with respect to the number of nodes and edges, which is on par with the existing dynamic GNNs.

\subsubsection{Background of Assumption \ref{assum}}
It is widely adopted in out-of-distribution generalization literature~\cite{gagnon2022woods,arjovsky2019invariant,rosenfeld2020risks,wu2022discovering,chang2020invariant,ahuja2020invariant,mitrovic2020representation} about the assumption that the relationship between labels and some parts of features is invariant across data distributions, and these subsets of features with such properties are called invariant features. In this paper, we use invariant patterns $\mathbf{P}_I$ to denote the invariant structures and features. 

From the causal perspective, we can formulate the data-generating process in dynamic graphs with a structural causal model (SCM)~\cite{pearl2000models,glymour2016causal}, $\mathbf{P}_V \rightarrow \mathbf{G} \leftarrow \mathbf{P}_I \rightarrow \mathbf{y}$ and $\mathbf{P}_V\leftarrow \mathbf{P}_I$, where the arrow between variables denotes casual relationship, and the subscript $v$ and superscript $t$ are omitted for brevity. $\mathbf{P}_V \rightarrow \mathbf{G} \leftarrow \mathbf{P}_I$ denotes that variant and invariant patterns construct the ego-graph trajectories observed in the data, while $\mathbf{P}_I \rightarrow \mathbf{y}$ denotes that invariant patterns determine the ground truth label $\mathbf{y}$, no matter how the variant patterns change inside data across different distributions. 

Sometimes, the correlations between variant patterns and labels may be built by some exogenous factors like periods and communities. In some distributions, $\mathbf{P}_V \leftarrow \mathbf{P}_I$  would open a backdoor path~\cite{glymour2016causal} $\mathbf{P}_V \leftarrow \mathbf{P}_I \rightarrow \mathbf{y}$ so that variant patterns $\mathbf{P}_V$ and labels $\mathbf{y}$ are correlated statistically, and this correlation is also called spurious correlation. If the model highly relies on the relationship between variant patterns and labels, it will fail under distribution shift, since such relationship varies across distributions. Hence, we propose to help the model focus on invariant patterns to make predictions and thus handle distribution shift.

\subsubsection{Connections in Remark \ref{remark}}
\red{To eliminate the spurious correlation between variant patterns and labels, one way is to block the backdoor path by using do-calculus to intervene the variant patterns. By applying do-calculus on one variable, all in-coming arrows(causal relationship) to it will be removed~\cite{glymour2016causal} and the intervened distributions will be created. In our case, the operator do($\mathbf{P}_V$) will cut the causal relationship from invariant patterns to variant patterns, \ie, disabling $\mathbf{P}_V \leftarrow \mathbf{P}_I$ and then blocking the backdoor path $\mathbf{P}_V \leftarrow \mathbf{P}_I \rightarrow \mathbf{y}$. Hence, the model can learn the direct causal effects from invariant patterns to labels in the intervened distributions $p(\mathbf{y},\mathbf{G}|\text{do}(\mathbf{P}_V))$, and the risks should be the same across these intervened distributions. Therefore we can minimize the variance of empirical risks under different intervened distributions to help the model focus on the relationship between invariant patterns and labels. On the other hand, if we have the optimal predictor $f_{\theta_1}^*$ and pattern finder $\phi_{\theta_2}^*$ according to Eq.(3), then the variance term in Eq.(4) is minimized as the variant patterns will not affect the predictions of $f_{\theta_1}^* \circ \phi_{\theta_2}^*$ across different intervened distributions.}

\red{In this paper, we refer \modelp as our method Disentangled Intervention-based Dynamic Graph Attention Networks with Invariance Promotion, and \model as a special case where $\lambda_{e} = 0$.}

\section{Experiments}\label{sec:exp}
In this section, we conduct extensive experiments to verify that our framework can handle spatio-temporal distribution shifts by discovering and utilizing invariant patterns. 

\subsection{Baselines}
We adopt several representative GNNs and Out-of-Distribution (OOD) generalization methods as our baselines. 
The first group of these methods is static GNNs, including:
\begin{itemize}
    \item \textbf{GAE}~\cite{kipf2016variational} is a representative static graph neural network with a stack of graph convolutions to capture the information of structures and attributes on graphs.
    \item \textbf{VGAE}~\cite{kipf2016variational} further introduces variational variables into GAE to obtain more robust and generalized graph representations.
\end{itemize}
The second group of these methods includes the following dynamic GNNs:
\begin{itemize}
    \item \textbf{GCRN}~\cite{seo2018structured} is a representative dynamic GNN that first adopts a GCN\cite{kipf2016variational} to obtain node embeddings and then a GRU~\cite{Cho2014LearningPR} to model the network evolution.
    \item \textbf{EvolveGCN}~\cite{pareja2020evolvegcn} adopts an LSTM~\cite{hochreiter1997long} or GRU~\cite{Cho2014LearningPR} to flexibly evolve the GCN~\cite{kipf2016variational} parameters instead of directly learning the temporal node embeddings, which is applicable to frequent change of the node set on dynamic graphs.
    \item \textbf{DySAT}~\cite{sankar2020dysat} aggregates neighborhood information at each graph snapshot using structural attention and models network dynamics with temporal self-attention so that the weights can be adaptively assigned for the messages from different neighbors in the aggregation.
\end{itemize}
And the third group of these methods consists of OOD generalization methods:
\begin{itemize}
    \item \textbf{IRM}~\cite{arjovsky2019invariant} aims at learning an invariant predictor which minimizes the empirical risks for all training domains to achieve out-of-distribution generalization.
    \item \textbf{GroupDRO}~\cite{sagawa2019distributionally} puts more weight on training domains with larger errors when minimizing empirical risk to minimize worst-group risks across training domains.
    \item \textbf{VREx}~\cite{krueger2021out} reduces differences in risk across training domains to reduce the model’s sensitivity to distributional shifts.
\end{itemize}

\red{These representative OOD generalization methods aim at improving the robustness and generalization ability of models against distribution shift, which requires explicit environment labels to calculate the loss. For fair comparisons, we randomly split the samples into different domains, as the field information is unknown to all methods. Since they are general OOD generalization methods and are not specifically designed for dynamic graphs, we adopt the best-performed DyGNN on the training datasets as their backbones.}

\begin{table*}[]
\centering
\caption{Summarization of dataset statistics. Evolving features denote whether the node features vary through time. Unseen nodes denote whether the test nodes are partially or fully unseen in the past.}
\label{tab:data}
\begin{tabular}{llllll}
\toprule
\textbf{Dataset}              & \textbf{COLLAB}  & \textbf{Yelp}    & \textbf{Synthetic} & \textbf{OGBN-Arxiv} & \textbf{Aminer}  \\ \midrule
\textbf{\# Timestamps}        & 16      & 24      & 16        & 20         & 17      \\
\textbf{\# Nodes}             & 23,035  & 13,095  & 23,035    & 168,195    & 43,141  \\
\textbf{\# Links}             & 151,790 & 65,375  & 151,790   & 3,127,274  & 851,527 \\
\textbf{Temporal Granularity} & Year    & Month   & Year      & Year       & Year    \\
\textbf{Feature Dimension}    & 32      & 32      & 64        & 128        & 128     \\
\textbf{Evolving Features}    & No      & No      & Yes       & No         & No      \\
\textbf{Unseen Nodes}         & Partial & Partial & Partial   & Full       & Full   \\ 
\textbf{Classification Tasks}         & Link & Link & Link   & Node       & Node   \\ \bottomrule
\end{tabular}
\end{table*}

\subsection{Real-world Link Prediction Datasets}
\subsubsection{Experimental Settings}
\red{We use two real-world dynamic graph datasets, including COLLAB and Yelp}. We adopt the challenging inductive future link prediction task, where the model exploits past graphs to make link prediction in the next time step. Each dataset can be split into several partial dynamic graphs based on its field information. For brevity, we use `w/ DS' and `w/o DS' to represent test data with and without distribution shift respectively.  
To measure models' performance under spatio-temporal distribution shift, we choose one field as `w/ DS' and the left others are further split into training, validation and test data (`w/o DS') chronologically. 
Note that the `w/o DS' is a merged dynamic graph without field information and `w/ DS' is unseen during training, which is more practical and challenging in real-world scenarios. Here we briefly introduce the real-world datasets as follows:
\begin{itemize}[leftmargin=0.5cm]
    \item \textbf{COLLAB}~\cite{Tang:12KDDCross}\footnote{https://www.aminer.cn/collaboration} is an academic collaboration dataset with papers that were published during 1990-2006. Node and edge represent author and coauthorship respectively. Based on the field of co-authored publication, each edge has the field information including "Data Mining", "Database", "Medical Informatics", "Theory" and "Visualization". The time granularity is year, including 16 time slices in total. We use "Data Mining" as `w/ DS' and the left as `w/o DS'. \red{We use word2vec~\cite{mikolov2013efficient} to extract 32-dimensional features from paper abstracts and average to obtain author features. We use 10,1,5 chronological graph slices for training, validation and testing respectively. The dataset includes 23,035 nodes and 151,790 links in total.} 
    \item \textbf{Yelp}~\cite{sankar2020dysat}\footnote{https://www.yelp.com/dataset} is a business review dataset, containing customer reviews on the business. Node and edge represent customer/business and review behavior respectively. We consider interactions in five categories of business including "Pizza", "American (New) Food", "Coffee \& Tea ",  "Sushi Bars" and "Fast Food" from January 2019 to December 2020. The time granularity is month, including 24 time slices in total. We use "Pizza" as `w/ DS' and the left as `w/o DS'. \red{We use word2vec~\cite{mikolov2013efficient} to extract 32-dimensional features from reviews and averages to obtain user and business features. We select users and items with interactions of more than 10. We use 15, 1, 8 chronological graph slices for training, validation and test respectively. The dataset includes 13,095 nodes and 65,375 links in total.}
\end{itemize}

\begin{table}[]
\centering
\caption{Results (AUC\%) of different methods on real-world link prediction datasets. The best results are in bold and the second-best results are underlined. `w/o DS' and `w/ DS' denote test data with and without distribution shift. }
\adjustbox{max width=\textwidth}{
\begin{tabular}{ccccc}
\toprule
\textbf{\textbf{Model}  $\backslash $ \textbf{Dataset} } & \multicolumn{2}{c}{\textbf{COLLAB}}       & \multicolumn{2}{c}{\textbf{Yelp}}         \\ \midrule
Test Data               & w/o DS              & w/ DS               & w/o DS              & w/ DS               \\ \midrule
GAE                     & \ms{77.15}{0.50}    & \ms{74.04}{0.75}    & \ms{70.67}{1.11}    & \ms{64.45}{5.02}    \\
VGAE                    & \ms{86.47}{0.04}    & \ms{74.95}{1.25}    & \ms{76.54}{0.50}    & \ms{65.33}{1.43}    \\
GCRN                    & \ms{82.78}{0.54}    & \ms{69.72}{0.45}    & \ms{68.59}{1.05}    & \ms{54.68}{7.59}    \\
EGCN                    & \ms{86.62}{0.95}    & \ms{76.15}{0.91}    & \ms{78.21}{0.03}    & \ms{53.82}{2.06}    \\
DySAT                   & \mstwo{88.77}{0.23} & \mstwo{76.59}{0.20} & \ms{78.87}{0.57}    & \ms{66.09}{1.42}    \\
IRM                     & \ms{87.96}{0.90}    & \ms{75.42}{0.87}    & \ms{66.49}{10.78}   & \ms{56.02}{16.08}   \\
VREx                    & \ms{88.31}{0.32}    & \ms{76.24}{0.77}    & \mstwo{79.04}{0.16} & \ms{66.41}{1.87}    \\
GroupDRO                & \ms{88.76}{0.12}    & \ms{76.33}{0.29}    & \msone{79.38}{0.42} & \mstwo{66.97}{0.61} \\ \midrule
\modelnosp              & \ms{91.97}{0.05} & \ms{81.87}{0.40} & \ms{78.22}{0.40}    & \ms{75.92}{0.90} \\
\modelpnosp             & \msone{92.17}{0.40} & \msone{82.40}{0.70} & \ms{78.17}{0.76} & \msone{76.90}{1.87} \\ \bottomrule
\end{tabular}
}
\label{tab:real}
\end{table}

\subsubsection{Experimental Results}
Based on the results on real-world link prediction datasets in Table~\ref{tab:real}, we have the following observations:
\begin{itemize}[leftmargin=0.5cm]
    \item Baselines fail dramatically under distribution shift: 1) Although DyGNN baselines perform well on test data without distribution shift, their performance drops greatly under distribution shift. In particular, the performance of DySAT, which is the best-performed DyGNN in `w/o DS', drops by nearly 12\%, 12\% and 5\% in `w/ DS'. In Yelp, GCRN and EGCN even underperform static GNNs, GAE and VGAE. This phenomenon shows that the existing DyGNNs may exploit variant patterns and thus fail to handle distribution shift. 2) Moreover, as generalization baselines are not specially designed to consider spatio-temporal distribution shift in dynamic graphs, they only have limited improvements in Yelp. In particular, they rely on ground-truth environment labels to achieve OOD generalization, which are unavailable for real dynamic graphs. The inferior performance indicates that they cannot generalize well without accurate environment labels, which verifies that lacking environmental labels is also a key challenge for handling distribution shifts of dynamic graphs. 
    
    \item Our method can better handle distribution shift than the baselines, especially in stronger distribution shift.
    \modelp improves significantly over all baselines in `w/ DS' for all datasets. 
    Note that Yelp has stronger temporal distribution shift since COVID-19 happens in the midway, strongly affecting consumers' behavior in business, while \modelp outperforms the most competitive baseline GroupDRO by 9\% in `w/ DS'. In comparison to similar field information in Yelp (all restaurants),
    COLLAB has stronger spatial distribution shift since the fields are more different to each other,
    while \modelp outperforms the most competitive baseline DySAT by 5\% in `w/ DS'. 
\end{itemize}
\subsection{Real-world Node Classification Datasets}
\subsubsection{Experimental Settings}
\red{We use 2 real-world dynamic graph datasets, including OGBN-Arxiv~\cite{hu2020open} and Aminer~\cite{Tang:08KDD,sinha2015overview}. The two datasets are both citation networks, where nodes represent papers, and edges from $u$ to $v$ with timestamp $t$ denote the paper $u$ published at year $t$ cites the paper $v$. The node classification task on dynamic graphs is challenging since the nodes come in the future, e.g., new papers are published in the future, so that the model should exploit the spatio-temporal information to classify the nodes.  Following ~\cite{wu2022handling}, we also use the inductive learning settings, \ie, the test nodes are strictly unseen during training, which is more practical and challenging in real-world dynamic graphs. Here, we briefly introduce the real-world datasets as follows. }
\begin{itemize}
    \item \red{\textbf{OGBN-Arxiv}~\cite{hu2020open} is a citation network between all Computer Science (CS) arXiv papers indexed by MAG~\cite{wang2020microsoft}. Each paper has a 128-dimensional feature vector obtained by averaging the embeddings of words in its title and abstract, where the embeddings of individual words are computed by running the skip-gram model~\cite{mikolov2013distributed} over the MAG corpus. The task is to predict the 40 subject areas of arXiv CS papers, e.g., cs.AI, cs.LG, and cs.OS. We train on papers published between 2001 - 2011, validate on those published in 2012-2014, and test on those published since 2015. With the volume of scientific publications doubling every 12 years over the past century, spatio-temporal distribution shifts naturally exist on these dynamic graphs. The dataset has 168,195 nodes and 3,127,274 links in total.}

    \item \red{\textbf{Aminer}~\cite{Tang:08KDD,sinha2015overview} is a citation network extracted from DBLP, ACM, MAG, and other sources. We use word2vec~\cite{mikolov2013efficient} to extract 128-dimensional features from paper abstracts and average to obtain paper features. We select the top 20 venues, and the task is to predict the venues of the papers. Similar to the OGBN-Arxiv dataset, we train on papers published between 2001 - 2011, validate on those published in 2012-2014, and test on those published since 2015. As the test nodes are not seen during training, the model is tested to exploit the invariant spatio-temporal patterns and make stable predictions under distribution shifts. The dataset has 43,141 nodes and 851,527 links in total.}
\end{itemize}

\begin{table*}[]
\centering
\caption{Results (ACC\%) of different methods on real-world node classification datasets. The best results are in bold and the second-best results are underlined.}

\begin{tabular}{ccccccc}
\toprule
\textbf{Model}  $\backslash $ \textbf{Dataset} & \multicolumn{3}{c}{\textbf{OGBN-Arxiv}}                         & \multicolumn{3}{c}{\textbf{Aminer}}                             \\ \midrule
Split                                          & 2015-2016           & 2017-2018           & 2019-2020           & 2015                & 2016                & 2017                \\ \midrule
GRCN                                           & \ms{46.77}{2.03}    & \ms{45.89}{3.41}    & \ms{46.61}{3.29}    & \ms{47.96}{1.12}    & \mstwo{51.33}{0.62} & \mstwo{42.93}{0.71} \\
EGCN                                           & \ms{48.70}{2.12}    & \ms{47.31}{3.45}    & \ms{46.93}{5.17}    & \ms{44.14}{1.12}    & \ms{46.28}{1.84}    & \ms{37.71}{1.84}    \\
DySAT                                          & \ms{48.83}{1.07}    & \ms{47.24}{1.24}    & \ms{46.87}{1.37}    & \ms{48.41}{0.81}    & \ms{49.76}{0.96}    & \ms{42.39}{0.62}    \\
IRM                                            & \mstwo{49.57}{1.02} & \mstwo{48.28}{1.51} & \ms{46.76}{3.52}    & \ms{48.44}{0.13}    & \ms{50.18}{0.73}    & \ms{42.40}{0.27}    \\
VREx                                           & \ms{48.21}{2.44}    & \ms{46.09}{4.13}    & \ms{46.60}{5.02}    & \ms{48.70}{0.73}    & \ms{49.24}{0.27}    & \ms{42.59}{0.37}    \\
GroupDRO                                       & \ms{49.51}{2.32}    & \ms{47.44}{4.06}    & \mstwo{47.10}{4.39} & \mstwo{48.73}{0.61} & \ms{49.74}{0.26}    & \ms{42.80}{0.36}    \\ \midrule
\model                                           & \ms{51.46}{1.25}    & \ms{49.98}{2.04}    & \ms{50.91}{2.88}    & \ms{50.34}{0.81}    & \ms{51.43}{0.27}    & \ms{44.69}{0.06}    \\
\modelp                                          & \msone{51.53}{1.22} & \msone{50.44}{1.83} & \msone{51.87}{2.01} & \msone{51.12}{0.33} & \msone{52.35}{0.82} & \msone{45.09}{0.23} \\ \bottomrule
\end{tabular}
\label{tab:real_node}
\end{table*}

\subsubsection{Experimental Results}
\red{Based on the results on real-world node classification datasets in Table \ref{tab:real_node}, we have the following observations:}
\begin{itemize}
    \item \red{Most baselines have significant performance drops as time goes. On OGBN-Arxiv, for example, EGCN gradually drops from 48.70\% to 46.93\% from 2015 to 2020. This phenomenon may result from the spatio-temporal distribution shifts on dynamic graphs as time goes, e.g., there has been a significant increase in the quantity of academic papers being published, and topics as well as the citation patterns might be different from the past. Moreover, general out-of-distribution baselines have performance improvement over the DyGNN baselines, while the improvements are far from satisfactory since they are not specially designed for handling the complex spatio-temporal distribution shifts on dynamic graphs.}
    \item \red{Our method significantly alleviates the performance drop as time goes. On OGBN-Arxiv, for example, \modelp has a performance improvement of 2\%, 2\%, 4\% from 2015 to 2020 in comparisons with the best baselines, which verifies that our method can capture the invariant and variant spatio-temporal patterns inside data and exploit the invariant patterns to make predictions under distribution shifts. Moreover, our method has less variance in most cases, which may be due to that the sample-level and environment-level invariance loss can reduce the effects of the spurious correlations to obtain better performance under distribution shifts.}
\end{itemize}
\subsection{Synthetic Datasets}
\subsubsection{Experimental Settings}
To evaluate the model's generalization ability under spatio-temporal distribution shift, following ~\cite{wu2022handling}, we introduce manually designed shifts in dataset COLLAB with all fields merged. Denote original features and structures as $\mathbf{X}_1^t \in \mathbb{R}^{N \times d}$ and $\mathbf{A}^t \in \{0,1\}^{N\times N}$. For each time $t$, we uniformly sample $p(t)|\mathcal{E}^{t+1}|$ positive links and $(1-p(t))|\mathcal{E}^{t+1}|$ negative links in $\mathbf{A}^{t+1}$. Then they are factorized into variant features $\mathbf{X}_2^t \in \mathbb{R}^{N \times d}$ with a property of structural preservation. Two portions of features are concatenated as $\mathbf{X}^t=[\mathbf{X}_1^t, \mathbf{X}_2^t]$ as input node features for training and inference. The sampling probability $p(t)=\text{clip}(\overline p + \sigma cos(t),0,1)$ refers to the intensity of shifts, where the variant features $\mathbf{X}_2^t$ constructed with higher $p(t)$ will have stronger correlations with future link $\mathbf{A}^{t+1}$. We set $\overline p_{test} = 0.1,\sigma_{test} = 0,\sigma_{train} = 0.05$ and vary $\overline p_{train}$ in from 0.4 to 0.8 for evaluation. Since the correlations between $\mathbf{X}_2^t$ and label $\mathbf{A}^{t+1}$ vary through time and neighborhood, patterns include $\mathbf{X}_2^t$ are variant under distribution shifts. As static GNNs can not support time-varying features, we omit their results. 

\red{Here, we detail the construction of variant features $\mathbf{X}_2^t$. We use the same features as $\mathbf{X}_1^t$ and structures as $\mathbf{A}^t$ in COLLAB, and introduce features $\mathbf{X}_2^t$ with variable correlation with supervision signals. $\mathbf{X}_2^t$ are obtained by training the embeddings $\mathbf{X}_2 \in \mathbb{R}^{N\times d}$ with reconstruction loss $\ell(\mathbf{X}_2\mathbf{X}_2^{T},\tilde{\mathbf{A}}^{t+1})$, where $\tilde{\mathbf{A}}^{t+1}$ refers to the sampled links, and $\ell$ refers to cross-entropy loss function. The embeddings $\mathbf{X}_2^t$ are trained with Adam optimizer, learning rate 1e-1, weight decay 1e-5 and earlystop patience 50. In this way, we empirically find that the inner product predictor can achieve results of over 99\% AUC by using $\mathbf{X}^t_2$ to predict the sampled links $\tilde{\mathbf{A}}^{t+1}$, so that the generated features can have strong correlations with the sampled links. By controlling the $p$ mentioned in Section 4.2, we can control the correlations of $\mathbf{X}^t$ and labels $\mathbf{A}^{t+1}$ to vary in training and test stage.}

\begin{table*}[]
\centering
\caption{Results (AUC\%) of different methods on the synthetic dataset. The best results are in bold and the second-best results are underlined. Larger $\overline{p}$ denotes higher distribution shift level.}

\begin{tabular}{ccccccc}
\toprule
\textbf{Model}  $\backslash \overline{p}$ & \multicolumn{2}{c}{\textbf{0.4}}                   & \multicolumn{2}{c}{\textbf{0.6}}                   & \multicolumn{2}{c}{\textbf{0.8}}                   \\ \midrule
Split                                     & Train               & Test                & Train               & Test                & Train               & Test                \\ \midrule
GCRN                                      & \ms{69.60}{1.14}    & \mstwo{72.57}{0.72} & \ms{74.71}{0.17}    & \mstwo{72.29}{0.47} & \ms{75.69}{0.07}    & \mstwo{67.26}{0.22} \\
EGCN                                      & \ms{78.82}{1.40}    & \ms{69.00}{0.53}    & \ms{79.47}{1.68}    & \ms{62.70}{1.14}    & \ms{81.07}{4.10}    & \ms{60.13}{0.89}    \\
DySAT                                     & \ms{84.71}{0.80}    & \ms{70.24}{1.26}    & \ms{89.77}{0.32}    & \ms{64.01}{0.19}    & \ms{94.02}{1.29}    & \ms{62.19}{0.39}    \\
IRM                                       & \mstwo{85.20}{0.07} & \ms{69.40}{0.09}    & \ms{89.48}{0.22}    & \ms{63.97}{0.37}    & \msone{95.02}{0.09} & \ms{62.66}{0.33}    \\
VREx                                      & \ms{84.77}{0.84}    & \ms{70.44}{1.08}    & \ms{89.81}{0.21}    & \ms{63.99}{0.21}    & \ms{94.06}{1.30}    & \ms{62.21}{0.40}    \\
GroupDRO                                  & \ms{84.78}{0.85}    & \ms{70.30}{1.23}    & \mstwo{89.90}{0.11} & \ms{64.05}{0.21}    & \ms{94.08}{1.33}    & \ms{62.13}{0.35}    \\ \midrule
\model                                    & \ms{87.92}{0.92}    & \ms{85.20}{0.84}    & \ms{91.22}{0.59}    & \ms{82.89}{0.23}    & \ms{92.72}{2.16}    & \ms{72.59}{3.31}    \\
\modelp                                   & \msone{88.50}{0.46} & \msone{85.27}{0.06} & \msone{92.27}{1.02} & \msone{83.00}{1.08} & \mstwo{94.23}{0.23} & \msone{74.87}{1.59} \\ \bottomrule
\end{tabular}

\label{tab:synthetic}
\end{table*}

\subsubsection{Experimental Results}
Based on the results on the synthetic dataset in Table. ~\ref{tab:synthetic}, we have the following observations:
\begin{itemize}[leftmargin=0.5cm]
    \item Our method can better handle distribution shift than the baselines. Although the baselines achieve high performance when training, their performance drops drastically in the test stage, which shows that the existing DyGNNs fail to handle distribution shifts. In terms of test results, \modelp consistently outperforms DyGNN baselines by a significantly large margin. In particular, \modelp surpasses the best-performed baseline by nearly 13\%/10\%/5\% in test results for different shift levels.  For the general OOD baselines, they reduce the variance in some cases while their improvements are not significant. Instead, \modelp is specially designed for dynamic graphs and can exploit the invariant spatio-temporal patterns to handle distribution shift. 
    
    \item Our method can exploit invariant patterns to consistently alleviate harmful effects of variant patterns under different distribution shift levels. As shift level increases, almost all baselines increase in train results and decline in test results. This phenomenon shows that as the relationship between variant patterns and labels goes stronger, the existing DyGNNs become more dependent on the variant patterns when training, causing their failure in the test stage. Instead, the rise in train results and drop in test results of \modelp are significantly lower than baselines, which demonstrates that \modelp can exploit invariant patterns and alleviate the harmful effects of variant patterns under distribution shift.
\end{itemize}

\subsection{Ablation Studies}
In this section, we conduct ablation studies to verify the effectiveness of the proposed spatio-temporal environment inference, spatio-temporal intervention mechanism and disentangled graph attention in \modelpnosp. 

\begin{figure}
    \centering
    \includegraphics[width=0.7\textwidth]{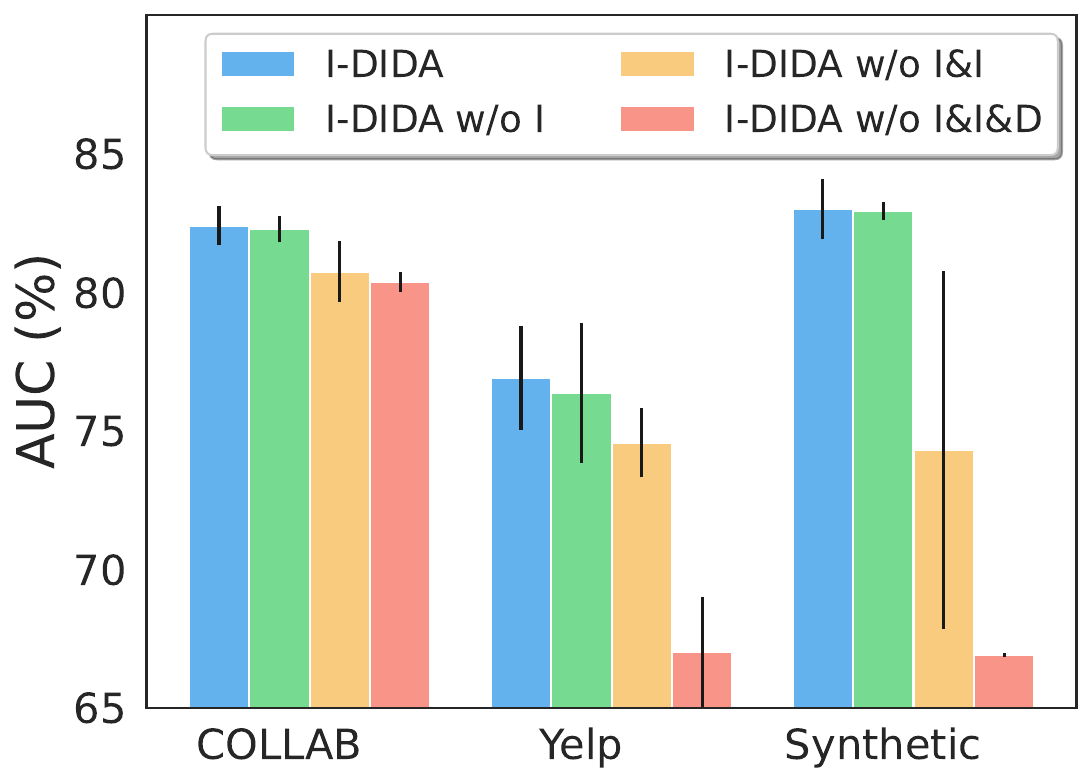}    
    \caption{\red{Ablation studies on the environment inference, intervention mechanism and disentangled attention, where 'w/o I' removes the spatio-temporal environment inference module, 'w/o I\&I' further removes the spatio-temporal intervention mechanism and 'w/o I\&I\&D' further removes disentangled attention. (Best viewed in color)}}
    \label{fig:ablation}
\end{figure}

\subsubsection{Spatio-Temporal Environment Inference}
\red{We remove the environment inference module mentioned in Sec.~\ref{sec:env_inference}. From Figure~\ref{fig:ablation}, we can see that without the spatio-temporal environment inference module, the model has a performance drop especially in the Yelp dataset, which verifies that our environment-level invariance loss helps the model to promote the invariance properties of the invariant patterns. } 
\subsubsection{Spatio-Temporal Intervention Mechanism}
We remove the intervention mechanism mentioned in Sec.~\ref{sec:intervener}. From Figure~\ref{fig:ablation}, we can see that without spatio-temporal intervention, the model's performance drop significantly especially in the synthetic dataset, which verifies that our intervention mechanism helps the model to focus on invariant patterns to make predictions. 

\subsubsection{Disentangled Dynamic Graph Attention}
We further remove the disentangled attention mentioned in Sec~\ref{sec:ddgat}. From Figure~\ref{fig:ablation}, we can see that disentangled attention is a critical component in the model design, especially in Yelp dataset. Moreover, without disentangled module, the model is unable to obtain variant and invariant patterns for the subsequent intervention.

\subsection{Additional Experiments}

\begin{figure}
    \centering
    \subfloat[COLLAB]{\includegraphics[width=0.48\textwidth]{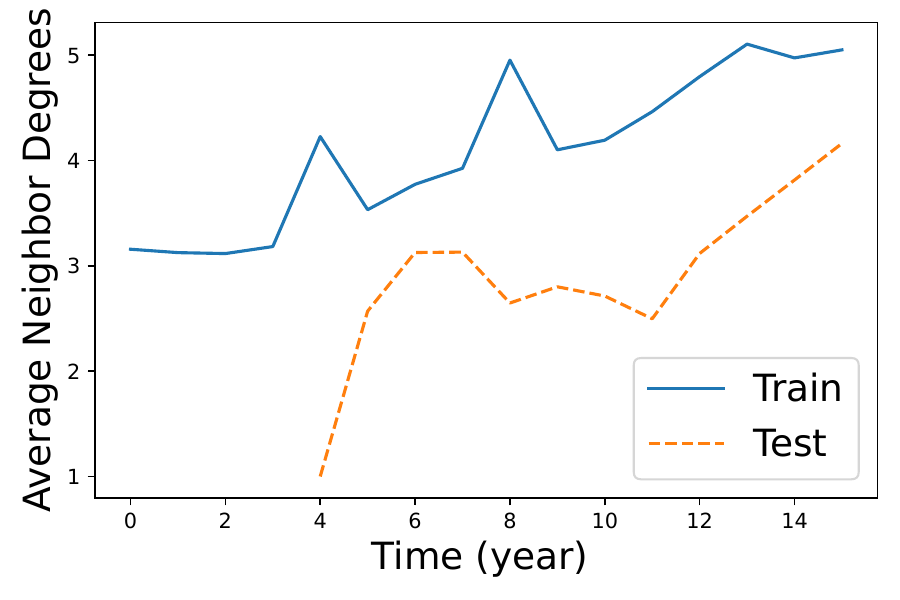}}
    \subfloat[Yelp]{\includegraphics[width=0.482\textwidth]{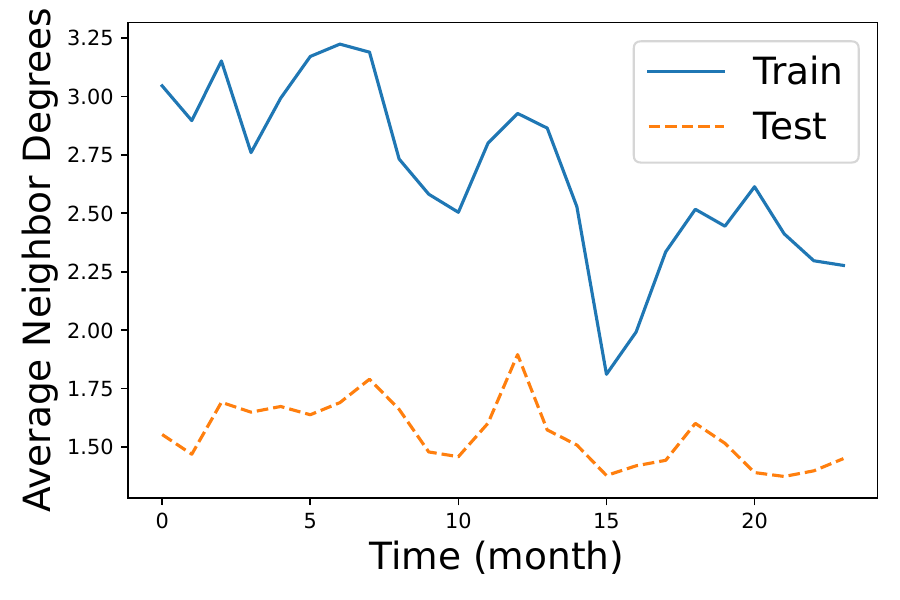}}
    
    \subfloat[OGBN-Arxiv]{\includegraphics[width=0.48\textwidth]{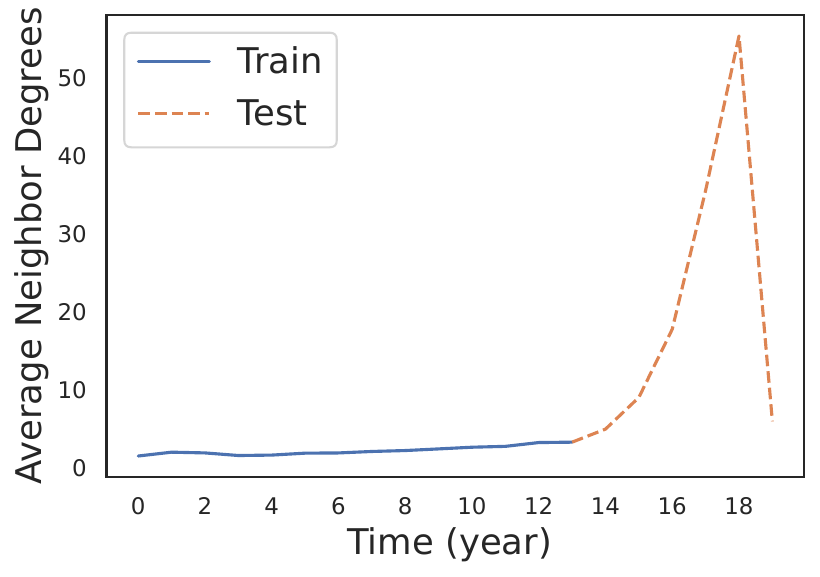}}
    \subfloat[Aminer]{\includegraphics[width=0.483\textwidth]{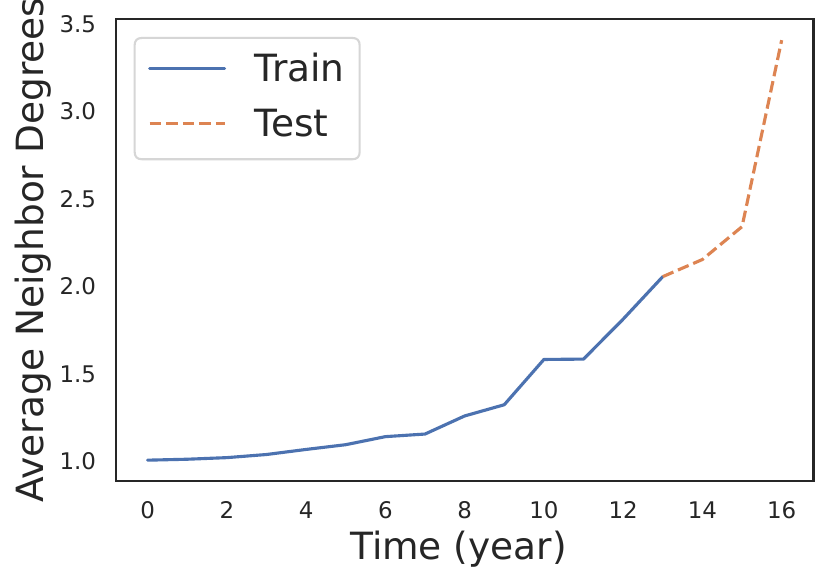}}
    \caption{Average neighbor degrees in the graph slice as time goes.}
    \label{fig:degs}
\end{figure}

\begin{figure*}
    \centering
    \subfloat[COLLAB]{\includegraphics[width=0.48\textwidth]{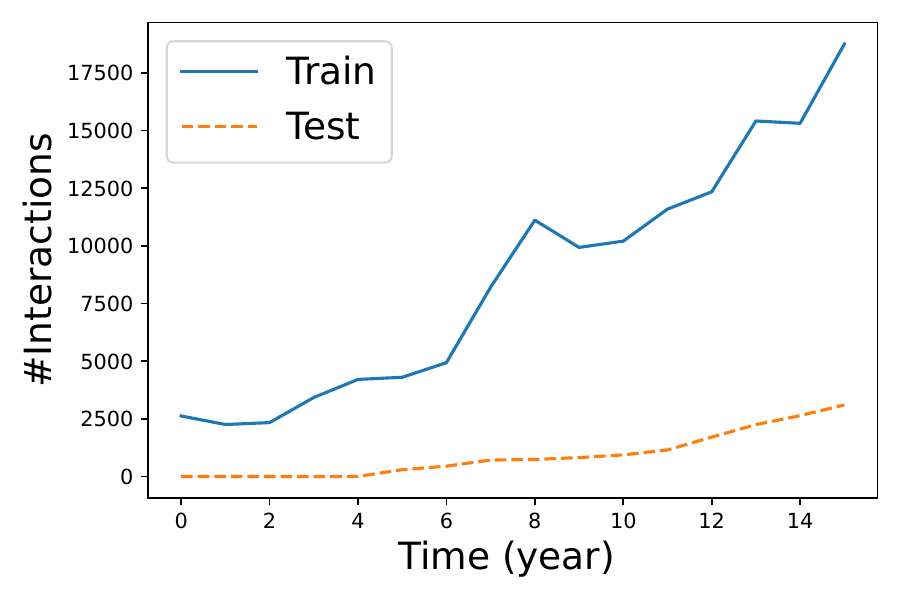}}
    \subfloat[Yelp]{\includegraphics[width=0.48\textwidth]{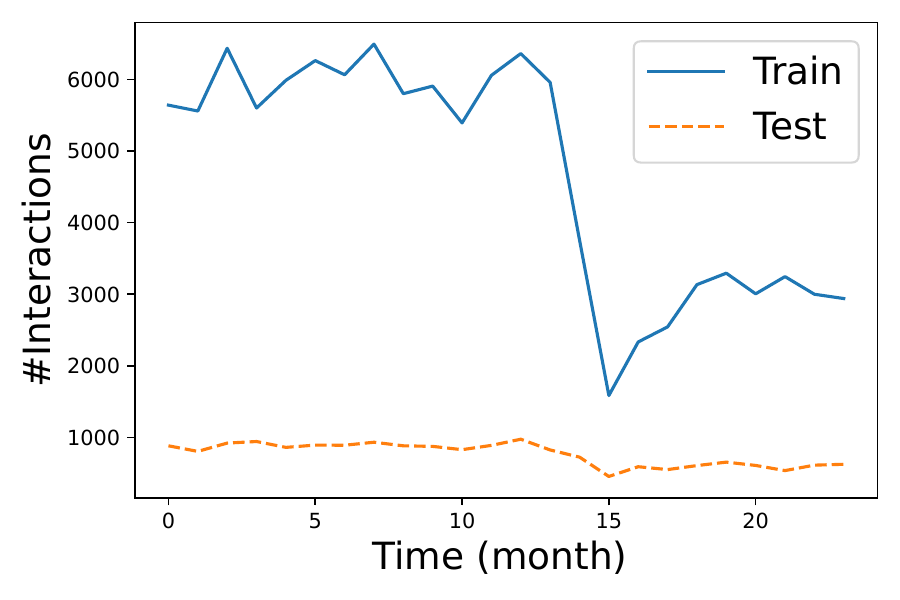}}
    
    \subfloat[OGBN-Arxiv]{\includegraphics[width=0.46\textwidth]{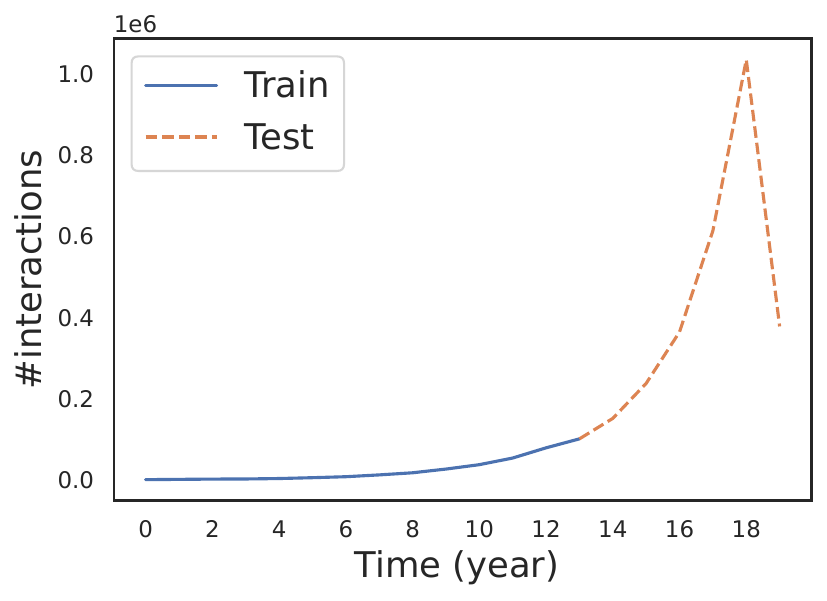}}
    \subfloat[Aminer]{\includegraphics[width=0.48\textwidth]{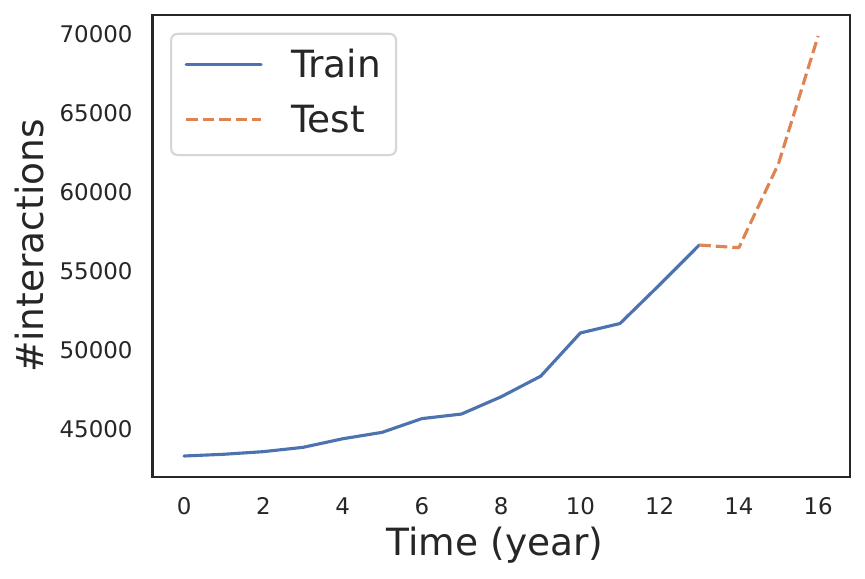}}
    \caption{Number of links in the graph slice as time goes.}
    \label{fig:links}
\end{figure*}

\subsubsection{Distribution Shifts in Real-world Datasets}
\red{We illustrate the distribution shifts in the real-world datasets with two statistics, number of links and average neighbor degrees~\cite{barrat2004architecture}. Figure~\ref{fig:degs} shows that the average neighbor degrees are lower in test data compared to training data. Lower average neighbor degree indicates that the nodes have less affinity to connect with high-degree neighbors. Moreover, in COLLAB, the test data has less history than training data, \ie, the graph trajectory is not always complete in training and test data distribution. This phenomenon of incomplete history is common in real-world scenarios, e.g. not all the users join the social platforms at the same time. Figure~\ref{fig:links} shows that the number of links and its trend also differ in training and test data. In COLLAB, \#links of test data has a slower rising trend than training data. In Yelp, \#links of training and test data both have a drop during time 13-15 and rise again thereafter, due to the outbreak of COVID-19, which strongly affected the 
consumers' behavior. Similarly, Figure~\ref{fig:degs} and Figure~\ref{fig:links} show that the number of links and the average neighbor degrees have a drastic increase in the test split on the Aminer and OGBN-Arxiv datasets, leading that the recent patterns on dynamic graphs might be significantly different from the past.}

\begin{figure}[]
\centering
\subfloat[]{\includegraphics[width=0.35\textwidth]{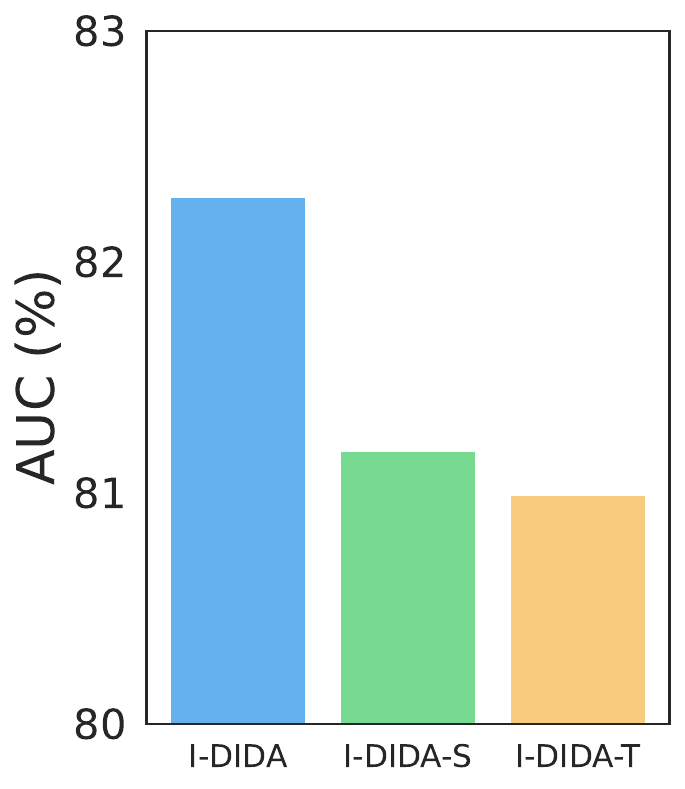}
\label{fig:st}
}
\hspace{10pt}
\subfloat[]{\includegraphics[width=0.36\textwidth]{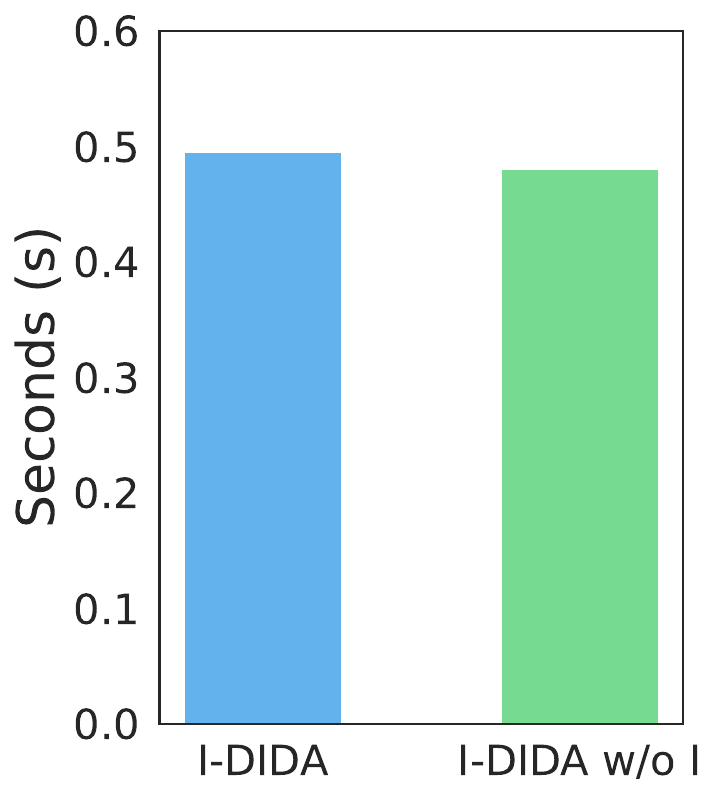}
\label{fig:iv_time}
}

\caption{(a) Comparison of different intervention mechansim on COLLAB dataset, where \red{I-DIDA-S only uses spatial intervention and I-DIDA-T only uses temporal intervention}. (b) Comparison in terms of training time for each epoch on COLLAB dataset, where 'w/o I' means removing intervention mechanism in \modelpnosp. (Best viewed in color)}
\end{figure}

\subsubsection{Spatial or Temporal Intervention}
\red{We compare two other versions of \modelp, where I-DIDA-S only uses spatial intervention and I-DIDA-T only uses temporal intervention. For I-DIDA-S, we put the constraint that the variant patterns used to intervene must come from the same timestamp in Eq.(9) so that the variant patterns across time are forbidden for intervention. Similarly, we put the constraint that the variant patterns used to intervene must come from the same node in Eq.(9) for I-DIDA-T. Figure~\ref{fig:st} shows that \modelp improves significantly over the other two ablated versions, which verifies that it is important to take into consideration both the spatial and temporal aspects of distribution shifts.}

\subsubsection{Efficiency of Intervention}
\red{For \modelp and \modelp without intervention mechanism, we compare their training time for each epoch on COLLAB dataset. As shown in Figure~\ref{fig:iv_time}, the intervention mechanism adds few costs in training time (lower than 5\%). Moreover, as \modelp does not use the intervention mechanism in the test stage, it does not add extra computational costs in the inference time.}

\begin{figure*}
    \centering
    \subfloat[COLLAB]{\includegraphics[width=0.33\textwidth]{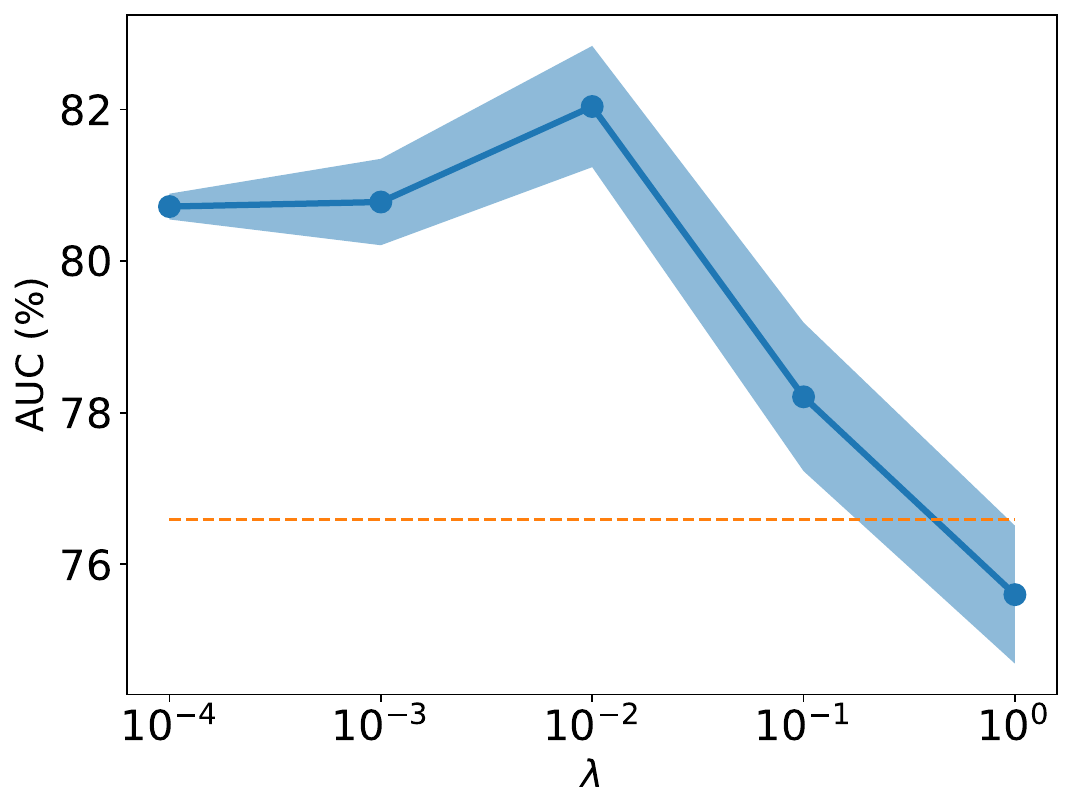}}
    \subfloat[Yelp]{\includegraphics[width=0.33\textwidth]{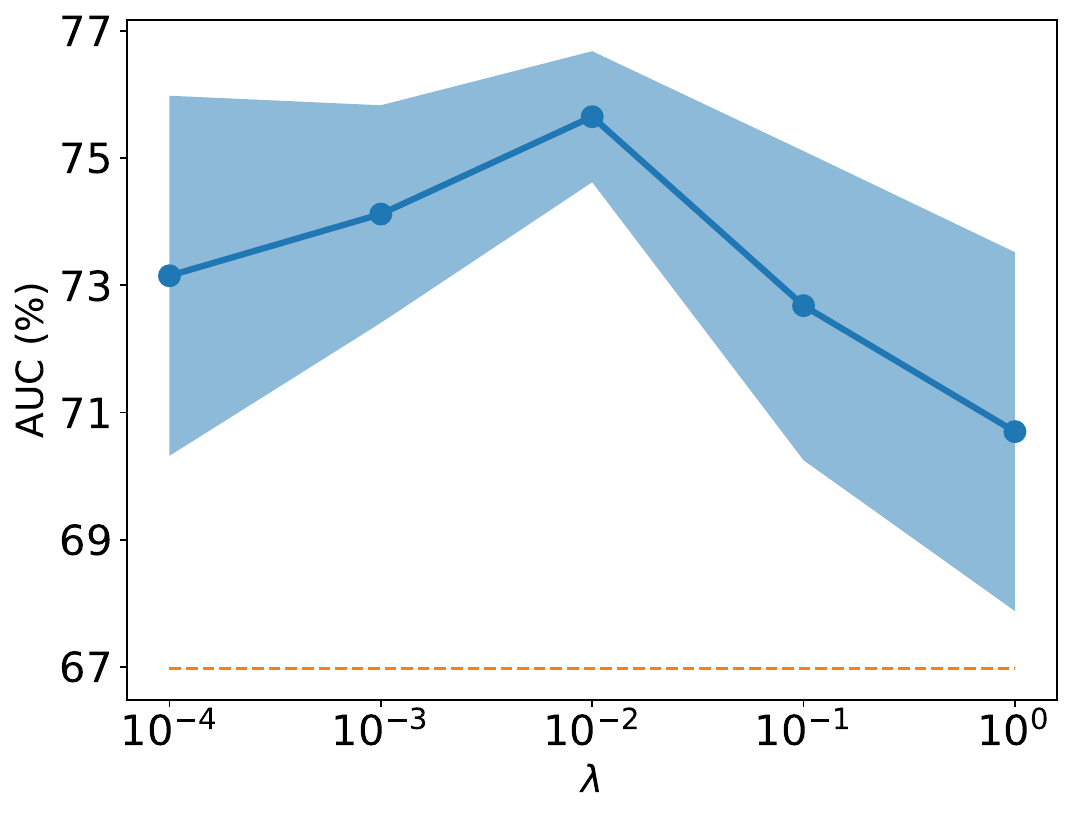}}
    \subfloat[Synthetic]{\includegraphics[width=0.33\textwidth]{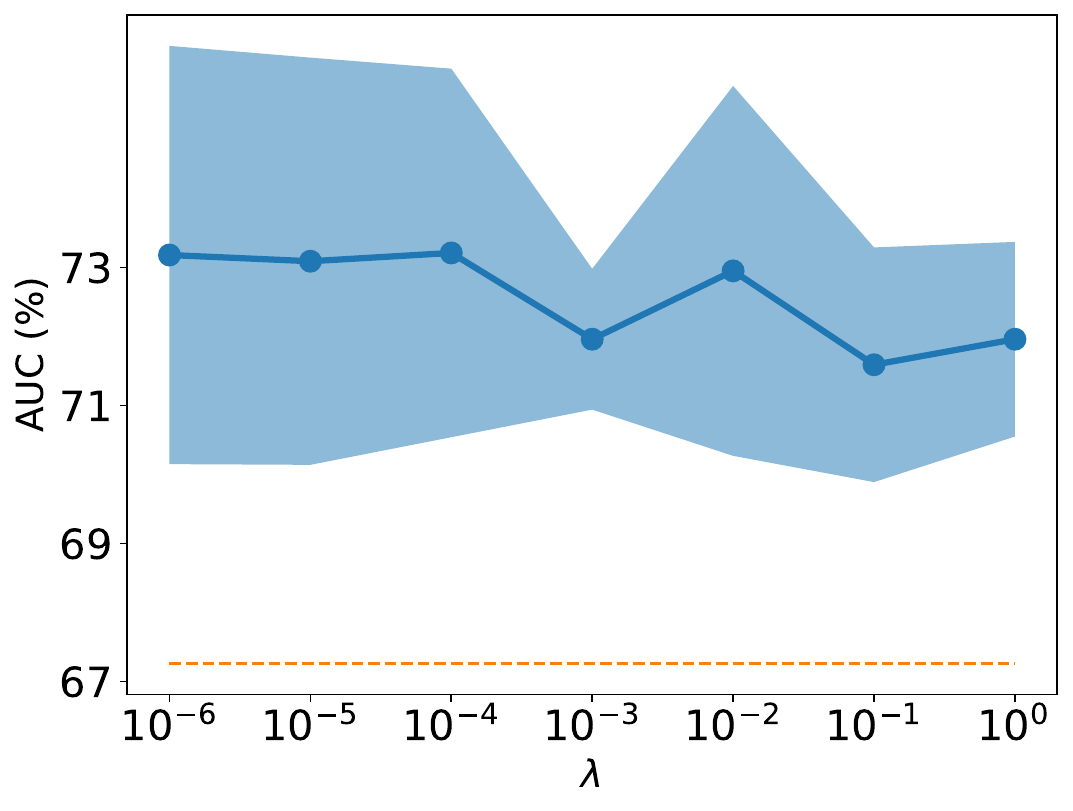}}
    \caption{\red{Sensitivity of hyperparameter $\lambda_{do}$ on different datasets.} The area shows the average AUC and standard deviations in the test stage. The dashed line represents the average AUC of the best-performed baseline.}
    \label{fig:sens}
\end{figure*}

\begin{figure*}
    \centering
    \subfloat[2015-2016]{\includegraphics[width=0.33\textwidth]{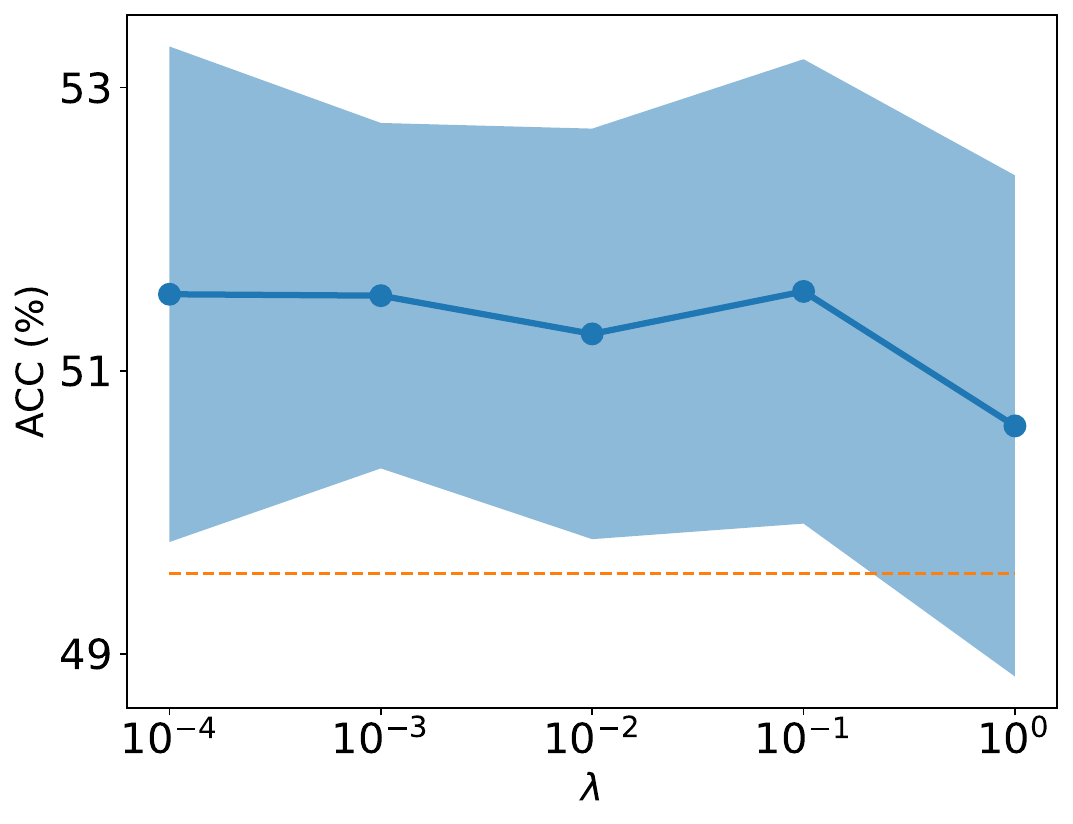}}
    \subfloat[2017-2018]{\includegraphics[width=0.33\textwidth]{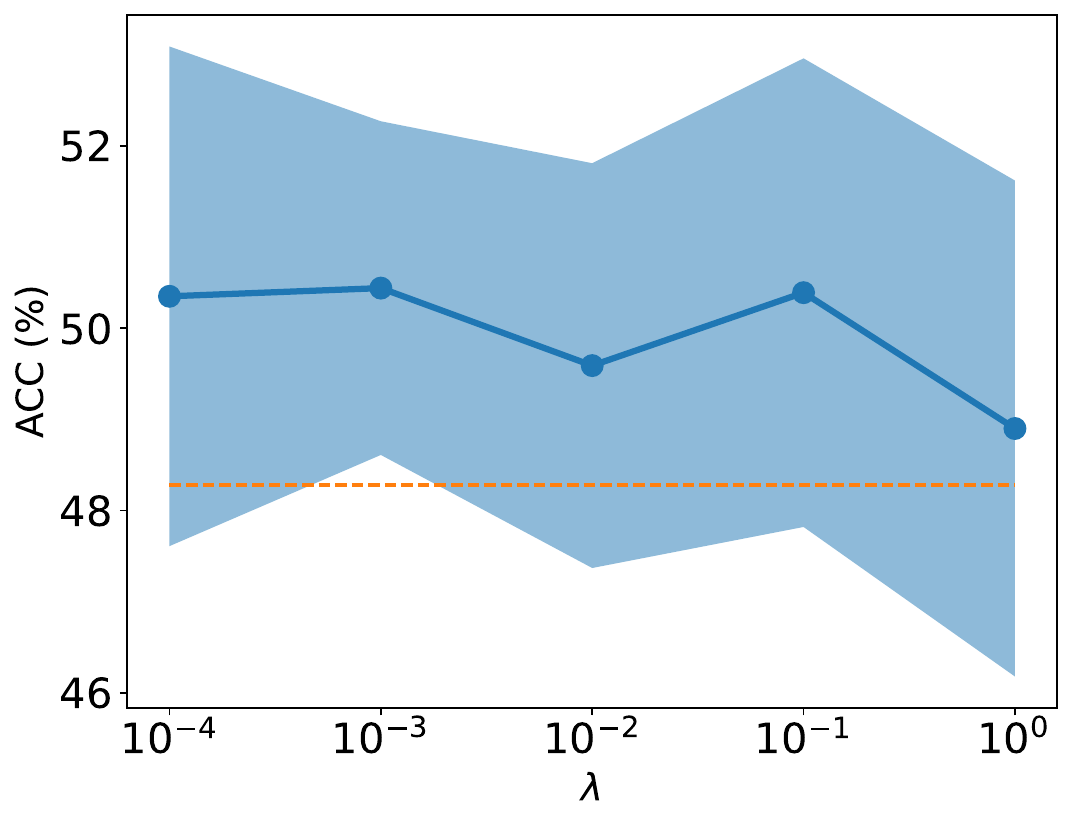}}
    \subfloat[2019-2020]{\includegraphics[width=0.33\textwidth]{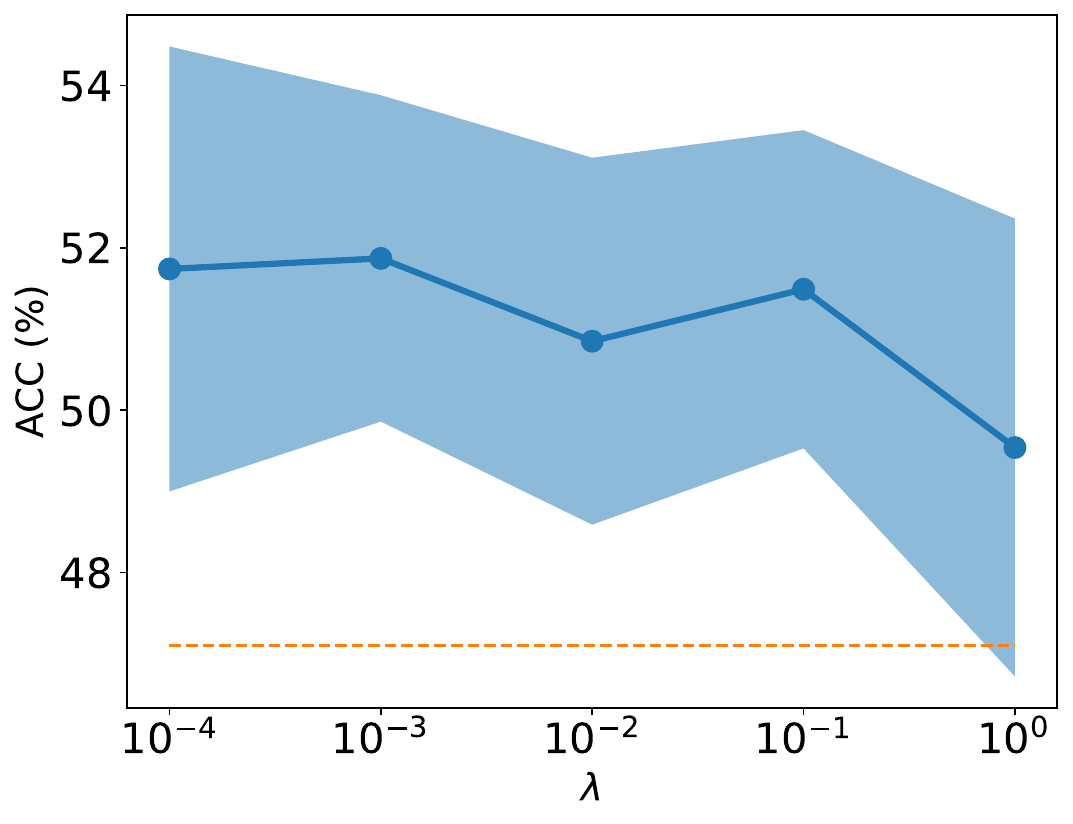}}
    \caption{Sensitivity of hyperparameter $\lambda_{e}$ on the OGBN-Arxiv dataset. The area shows the average accuracy and standard deviations in the test stage, which ranges from 2015 to 2020. The dashed line represents the average accuracy of the best-performed baseline.}
    \label{fig:sens-Arxiv}
\end{figure*}

\begin{figure*}
    \centering
    \subfloat[2015]{\includegraphics[width=0.33\textwidth]{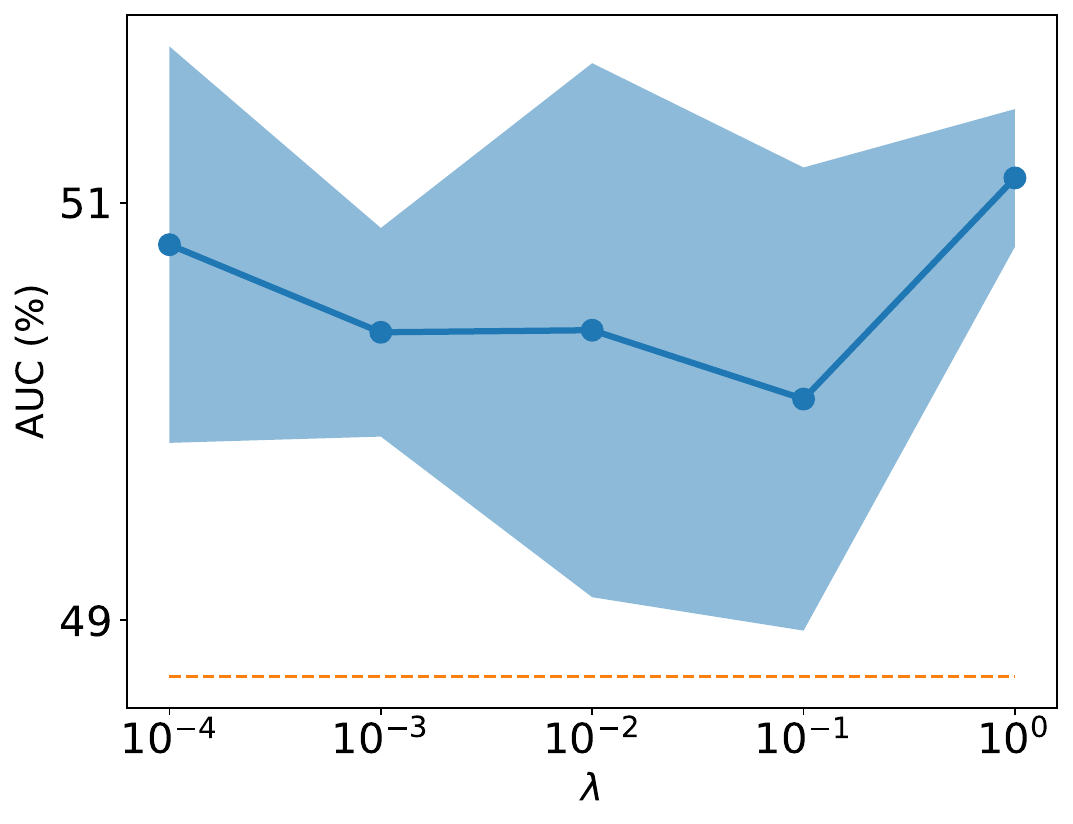}}
    \subfloat[2016]{\includegraphics[width=0.33\textwidth]{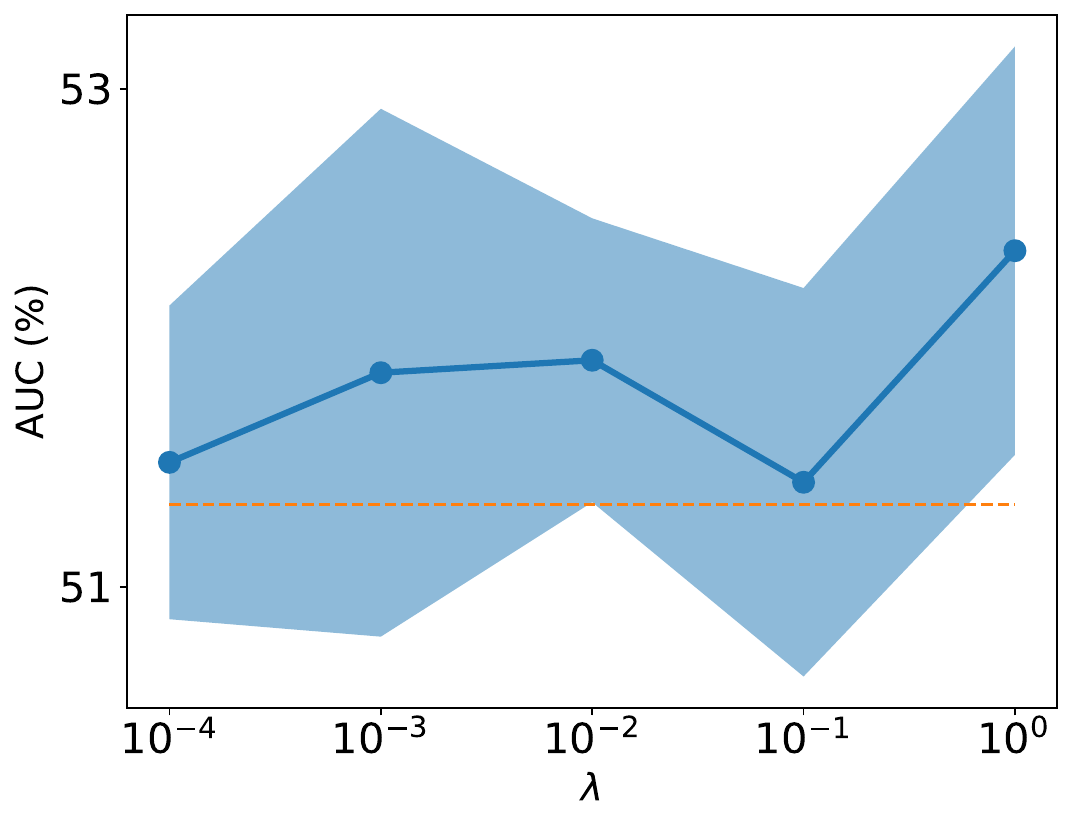}}
    \subfloat[2017]{\includegraphics[width=0.33\textwidth]{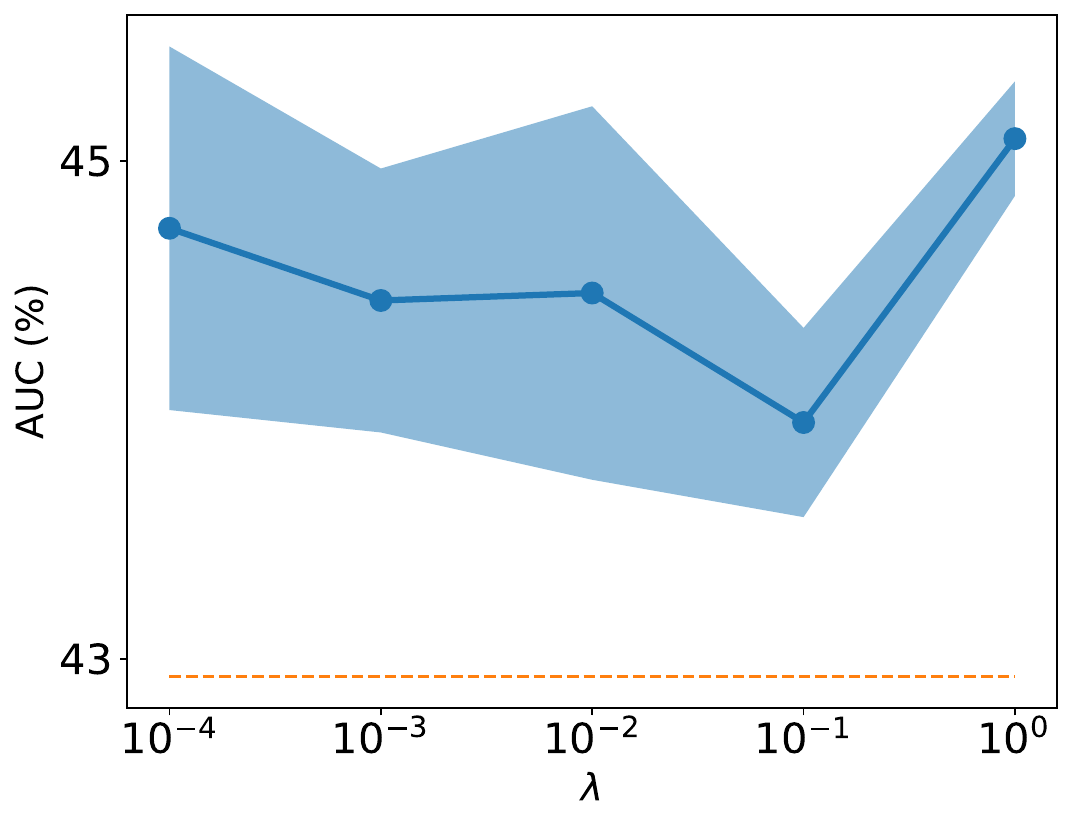}}
    \caption{Sensitivity of hyperparameter $\lambda_{e}$ on the Aminer dataset. The area shows the average accuracy and standard deviations in the test stage, which ranges from 2015 to 2017. The dashed line represents the average accuracy of the best-performed baseline.}
    \label{fig:sens-Aminer}
\end{figure*}

\subsubsection{Hyperparameter Sensitivity}
\red{We analyze the sensitivity of hyperparameter $\lambda_{do}$ in \modelp for each dataset. From Figure \ref{fig:sens}, we can see that as $\lambda_{do}$ is too small or too large, the model's performance drops in most datasets. It shows that $\lambda_{do}$ acts as a balance between how \modelp exploits the patterns and satisfies the invariance constraint. From Figure \ref{fig:sens-Aminer} and Figure \ref{fig:sens-Arxiv}, the model significantly outperforms the best-performed baseline with a large range of hyperparameters $\lambda_e$. It shows that the environment-level invariance loss promotes the invariance properties of the invariant patterns, and similarly, the hyperparameter $\lambda_e$ controls the balance between the empirical risk minimization and the invariance constraint.}

\subsection{Implementation Details}
\subsubsection{Hyperparameters}
\red{For all methods, we adopt the Adam optimizer~\cite{kingma2014adam} with a learning rate $0.01$, weight decay 5e-7 and set the patience of early stopping on the validation set as 50.  The hidden dimension is set to 16 for link prediction tasks and 32 for node classification tasks.
The number of layers is set to 2. Other hyper-parameters are selected using the validation datasets. For \modelnosp, we set the number of intervention samples as 1000 for link prediction tasks, and 100 for node classification tasks, and set $\lambda_{do}$ as 1e-2,1e-2,1e-1,1e-4,1e-4 for COLLAB, Yelp, Synthetic, Arxiv and Aminer dataset respectively. For \modelpnosp, we adopt cosine distance for all datasets, and coefficient $\lambda_{e}$ as 1e-2,1e-2,1e-1,1e-4,1 for COLLAB, Yelp, Synthetic, Arxiv and Aminer dataset respectively. }
\subsubsection{Evaluation Details}
\red{For link prediction tasks, we randomly sample negative samples from nodes that do not have links, and the negative samples for validation and testing set are kept the same for all comparing methods. The number of negative samples is the same as the positive ones. We use Area under the ROC Curve (AUC) as the evaluation metric. We use the inner product of the two learned node representations to predict links and use cross-entropy as the loss function $\ell$. We randomly run the experiments three times, and report the average results and standard deviations. For node classification tasks, we adopt cross-entropy as the loss function $\ell$ and use Accuracy (ACC) as the evaluation metric.}
\subsubsection{Model Details}
\red{Before stacking of disentangled spatio-temporal graph attention Layers, we use a fully-connected layer $\text{FC}(\cdot)$ to transform the features into hidden embeddings.} 
\begin{equation}
    \text{FC}(\mathbf{x})=\mathbf{W}\mathbf{x} + \mathbf{b}.
\end{equation}
\red{We implement the aggregation function for the invariant and variant patterns as }
\begin{equation}
\centering
\begin{aligned}
\tilde{\mathbf{z}}^t_I(u) &= \sum_i \mathbf{m}_{I,i} (\mathbf{v}_i\odot \mathbf{m}_f),
\\ 
\mathbf{z}^t_I(u) &=\text{FFN}(\tilde{\mathbf{z}}^t_I(u)+ \mathbf{h}_u^t),
\end{aligned}
\end{equation}

\begin{equation}
\centering
\begin{aligned}
\tilde{\mathbf{z}}^t_V(u) &= \sum_i \mathbf{m}_{V,i}  \mathbf{v}_i,
\\
\mathbf{z}^t_V(u) &=\text{FFN}(\tilde{\mathbf{z}}^t_V(u)),
\end{aligned}
\end{equation}
\red{where the FFN includes a layer normalization~\cite{ba2016layer}, multi-layer perceptron and skip connection, }
\begin{equation}
    \text{FFN}(\mathbf{x})=\alpha \cdot \text{MLP(LayerNorm}(\mathbf{x}))+(1-\alpha) \cdot \mathbf{x},
\end{equation}
\red{where $\alpha$ is a learnable parameter. For link prediction tasks, we implement the predictor $f(\cdot)$ in Eq.(10) as inner product of hidden embeddings, \ie,} 
\begin{equation}
f(\mathbf{z}_I^t(u),\mathbf{z}_I^t(v))=\mathbf{z}_I^t(u) \cdot (\mathbf{z}_I^t(v))^T,
\end{equation}
\red{
which conforms to classic link prediction settings. To implement the predictor $g(\cdot)$ in Eq.(11), we adopt the biased training technique following~\cite{cadene2019rubi}, \ie,}
\begin{equation}
\centering
\begin{aligned}
&g(\mathbf{z}^t_V(u),\mathbf{z}^t_I(u),\mathbf{z}^t_V(v),\mathbf{z}^t_I(v))\\
=&f(\mathbf{z}_I^t(u),\mathbf{z}_I^t(v))\cdot \sigma(f(\mathbf{z}_V^t(u),\mathbf{z}_V^t(v))),
\end{aligned}
\end{equation}
For node classification tasks, we implement the predictor $f(\cdot)$ in Eq.(10) as a linear classifer, \ie, 
\begin{equation}
f(\mathbf{z}_I^t(u)) = \mathbf{W}\mathbf{z}_I^t(u) + \mathbf{b}.
\end{equation}
Following ~\cite{wu2022discovering}, we use an additional shortcut loss to train the linear classifier of the variant patterns for the node $u$, \ie, 
\begin{equation}
\mathcal{L}_{s} = \ell(f(\mathbf{z}_V^t(u)),\mathbf{y}_u)
\end{equation}
Note that this loss is just used for training the classifier, and does not update other neural networks, e.g., the disentangled dynamic graph attention. Similarly, we implement the predictor $g(\cdot)$ in Eq.(11) as 
\begin{equation}
g(\mathbf{z}^t_V(u),\mathbf{z}^t_I(u))=f(\mathbf{z}_I^t(u))\cdot \sigma(f(\mathbf{z}_V^t(u)).
\end{equation}

\subsubsection{Configurations}

\red{We implement our method with PyTorch, and conduct the experiments on all datasets with:}
\begin{itemize}[leftmargin=0.5cm]
    \item \red{Operating System: Ubuntu 18.04.1 LTS}
    \item \red{CPU: Intel(R) Xeon(R) Gold 6240R CPU @ 2.40GHz}
    \item \red{GPU: NVIDIA GeForce RTX 3090 with 24 GB of memory}
    \item \red{Software: Python 3.8.13, Cuda 11.3, PyTorch~\cite{paszke2019pytorch} 1.11.0, PyTorch Geometric~\cite{Fey/Lenssen/2019} 2.0.3.}
\end{itemize} 

\section{Related Work}\label{sec:related}

In this section, we review the related works of dynamic graph neural networks, out-of-distribution generalization, and disentangled representation learning.

\subsection{Dynamic Graph Neural Networks}
To tackle the complex structural and temporal information in dynamic graphs, considerable research attention has been devoted to dynamic graph neural networks (DyGNNs)~\cite{skarding2021foundations,zhu2022learnable}. 

A classic of DyGNNs first adopt a graph neural networks~\cite{wu2020comprehensive,wang2019heterogeneous,wang2017community,zhou2020graph} to aggregate structural information for the graph at each time, followed by a sequence model like RNN~\cite{yang2021discrete,sun2021hyperbolic,hajiramezanali2019variational,seo2018structured} or temporal self-attention~\cite{sankar2020dysat} to process temporal information. \red{GCRN~\cite{seo2018structured} models the structural information for each graph snapshot at different timestamps with graph convolution networks~\cite{kipf2016semi} and adopt GRU~\cite{Cho2014LearningPR} to model the graph evolution along the temporal dimension. DyGGNN~\cite{taheri2019learning} adopts gated graph neural networks to learn the graph topology at each time step and LSTM~\cite{hochreiter1997long} to propagate the temporal information among the time steps. Variational inference is further introduced to model the node dynamics in the latent space~\cite{hajiramezanali2019variational}. DySAT~\cite{sankar2020dysat} aggregates neighborhood information at each snapshot similar to graph attention networks~\cite{velivckovicgraph} and aggregates temporal information with temporal self-attention. By introducing the attention mechanism, the model can draw context from all past graphs to adaptively assign weights for messages from different time and neighbors. Some works ~\cite{sun2021hyperbolic,yang2021discrete,bai2023hgwavenet} learn the embeddings of dynamic graphs in hyperbolic space to exploit the hyperbolic geometry's advantages of the exponential capacity and hierarchical awareness.} 

Another classic of DyGNNs first introduce time-encoding techniques to represent each temporal link as a function of time, followed by a spatial module like GNN or memory module~\cite{wang2021inductive,cong2021dynamic,xu2020inductive,rossi2020temporal} to process structural information. \red{For example, TGAT~\cite{xu2020inductive} proposes a functional time encoding technique based on the classical Bochner's theorem from harmonic analysis, which enables the learned node embeddings to be inherently represented as a function of time.} To obtain more fine-grained continuous node embeddings in dynamic graphs, some work further leverages neural interaction processes ~\cite{chang2020continuous} and ordinary differential equation~\cite{huang2021coupled}.  \red{EvolveGCN~\cite{pareja2020evolvegcn} models the network evolution from a different perspective, which learns to evolve the parameters of graph convolutional networks instead of the node embeddings by RNNs. In this way, the model does not require the knowledge of a node in the full time span, and is applicable to the frequent change of the node set.}

DyGNNs have been widely applied in real-world applications, including dynamic anomaly detection~\cite{cai2021structural}, event forecasting~\cite{deng2020dynamic}, dynamic recommendation~\cite{you2019hierarchical}, social character prediction~\cite{wang2021tedic}, user modeling~\cite{li2019fates}, temporal knowledge graph completion~\cite{wu2020temp}, entity linking ~\cite{wu2020dynamic}, {\it etc}. For example, DGEL~\cite{tang2023dynamic} proposes a dynamic graph evolution learning framework for generating satisfying recommendations in dynamic environments, including three efficient real-time update learning methods for nodes from the perspectives of inherent interaction potential, time-decay neighbor augmentation and symbiotic local structure learning. DynShare~\cite{zhao2023time} proposes a dynamic share recommendation model that is able to recommend a friend who would like to share a particular item at a certain timestamp for social-oriented e-commerce platforms. PTGCN~\cite{huang2023position} models the patterns between user-item interactions in sequential recommendation by defining a position-enhanced and time-aware graph convolution operation, demonstrating great potential for online session-based recommendation scenarios.  

In this paper, we consider DyGNNs under spatio-temporal distribution shift, which remains unexplored in 
dynamic graph neural networks literature.

\subsection{Out-of-Distribution Generalization}
Most existing machine learning methods assume that the testing and training data are independent and identically distributed, which is not guaranteed to hold in many real-world scenarios~\cite{shen2021towards}. In particular, there might be uncontrollable distribution shifts between training and testing data distribution, which may lead to a sharp drop in model performance. 

To solve this problem, Out-of-Distribution (OOD) generalization problem has recently become a central research topic in various areas ~\cite{shen2021towards,yao2022improving,wang2023tutorial}. 
\red{As a representative work tackling OOD generalization problems,  IRM~\cite{arjovsky2019invariant} aims at learning an invariant predictor which minimizes the empirical risks for all training domains, so that the classifier and learned representations match for all environments and achieve out-of-distribution generalization. GroupDRO~\cite{sagawa2019distributionally} minimizes worst-group risks across training domains by putting more weight on training domains with larger errors when minimizing empirical risk. VREx~\cite{krueger2021out} reduces differences in risk across training domains to reduce the model's sensitivity to distribution shifts.} 

Recently, several works attempt to handle distribution shift on graphs~\cite{chen2022invariance,qin2022graph,li2022ood,zhang2022learning,fan2021generalizing,li2022gil,10.1145/3604427,li2022out,yang2023interpretable, chen2022learning}, where the distribution shift can exist on graph topologies, e.g., graph sizes and other structural properties. For example, some work~\cite{bevilacqua2021size} assumes independence between cause and mechanism, and constructs a structural causal model to learn the graph representations that can extrapolate among different size distributions for graph classification tasks. Some work~\cite{han2022g} interpolates the node features and graph structure in embedding space as data augmentation to improve the model's OOD generalization abilities. EERM~\cite{wu2022handling} proposes to utilize multiple context explorers that are adversarially trained to maximize the variance of risks from multiple virtual environments, so that the model can extrapolate from a single observed environment for node-level prediction. DIR~\cite{wu2022discovering} attempts to capture the causal rationales that are invariant under structural distribution shift and filter out the unstable spurious patterns. DR-GST~\cite{liu2022confidence} finds that high-confidence unlabeled nodes may introduce the distribution shift issue between the original labeled dataset and the augmented dataset in self-training, and proposes a framework to recover the distribution of the original labeled dataset. SR-GNN~\cite{zhu2021shift} adapts GNN models to tackle the distributional differences between biased training data and the graph's true inference distribution. GDN~\cite{gao2023alleviating} discovers the structural distribution shifts in graph anomaly detection, that is, the heterophily and homophily can change across training and testing data. They solve the problem by teasing out the anomaly features, on which they constrain to mitigate the effect of heterophilous neighbors and make them invariant. GOOD-D~\cite{liu2023good} studies the problem of unsupervised graph out-of-distribution detection and creates a comprehensive benchmark to make comparisons of several state-of-the-art methods.

Another classic of OOD methods most related to our works handle distribution shifts on time-series data~\cite{venkateswaran2021environment,lu2021diversify}. For example, some work~\cite{kim2021reversible} observes that statistical properties such as mean and variance often change over time in time series, and propose a reversible instance normalization method to remove and restore the statistical information for tackling the distribution shifts. AdaRNN~\cite{du2021adarnn} formulates the temporal covariate shift problem for time series forecasting and proposes to characterize the distribution information and reduce the distribution mismatch during the training of RNN-based prediction models. DROS~\cite{yang2023generic} proposes a distributionally robust optimization mechanism with a distribution adaption paradigm to capture the dynamics of data distribution and explore the possible distribution shifts for sequential recommendation. Wild-Time~\cite{yao2022wildtime} creates a benchmark of datasets that reflect the temporal distribution shifts arising in a variety of real-world time-series applications like patient prognosis, showing that current time-series and out-of-distribution methods still have limitations in tackling temporal distribution shifts. WOODS~\cite{gagnon2022woods} is another benchmark for out-of-distribution generalization methods in time series tasks, including videos, brain recordings, smart device sensory signals, {\it etc.} 

Current works consider either only structural distribution shift for static graphs or only temporal distribution shift for time-series data. However, spatio-temporal distribution shifts in dynamic graphs are more complex yet remain unexplored. To the best of our knowledge, this paper is the first study of spatio-temporal distribution shifts in dynamic graphs. 
\subsection{Disentangled Representation Learning}
Disentangled representation learning aims to characterize the multiple latent explanatory factors behind the observed data, where the factors are represented by different vectors ~\cite{bengio2013representation}. 
Besides its applications in computer vision ~\cite{hsieh2018learning, ma2018disentangled,chen2016infogan,denton2017unsupervised,tran2017disentangled}
and recommendation ~\cite{wang2022disentangled,chen2021curriculum,wang2021multimodal,ma2020disentangled,ma2019learning,li2021intention}, several disentangled GNNs have proposed to generalize disentangled representation learning in graph data recently. 
\red{DisenGCN~\cite{ma2019disentangled} learns disentangled node representations by proposing a neighborhood routing mechanism in the graph convolution networks to identify the factors that may cause the links from the nodes to their neighbors. IPGDN~\cite{liu2020independence} further encourages the graph latent factors to be as independent as possible by minimizing the dependence among representations with a kernel-based measure. FactorGCN~\cite{yang2020factorizable} decomposes the input graph into several interpretable factor graphs, and each of the factor graphs is fed into a different GCN so that different aspects of the graph can be modeled into factorized graph representations. DGCL~\cite{li2021disentangled} and IDGCL~\cite{li2022disentangled} aim to learn disentangled graph-level representations with self-supervision to reduce the potential negative effects of the bias brought by supervised signals. However, most of these methods are designed for static graphs and may not disentangle the factors with the consideration of the structural and temporal information on graphs.} GRACES~\cite{qin2022graph} designs a self-supervised disentangled graph encoder to characterize the invariant factors hidden in diverse graph structures, and thus facilitates the subsequent graph neural architecture search. Some other works factorize deep generative models based on node, edge, static, dynamic factors~\cite{zhang2021disentangled} or spatial, temporal, graph factors~\cite{du2022disentangled} to achieve interpretable dynamic graph generation. DisenCTR~\cite{wang2022disenctr} proposes a disentangled graph representation module to extract diverse user interests and exploit the fluidity of user interests and model the temporal effect of historical behaviors using a mixture of Hawkes process. 
\red{In this paper, we borrow the idea of disentangled representation learning, and disentangle the spatio-temporal patterns on dynamic graphs into invariant and variant parts for the subsequent invariant learning to enhance the model's generalization ability under distribution shifts.} 

\section{Conclusion}\label{sec:conclusion}
\red{In this paper, we propose Disentangled Intervention-based Dynamic Graph Attention Networks with Invariance Promotion (\modelpnosp) to handle spatio-temporal distribution shift in dynamic graphs.} First, we propose a disentangled dynamic graph attention network to capture invariant and variant spatio-temporal patterns. Then, based on the causal inference literature, we design a spatio-temporal intervention mechanism to create multiple intervened distributions and propose an invariance regularization term to help the model focus on invariant patterns under distribution shifts. \red{Moreover, based on the invariant learning literature, we design a spatio-temporal environment inference to infer the latent environments of the nodes at different time on dynamic graphs, and propose an environment-level invariance loss to promote the invariance properties of the captured invariant patterns.} \red{Extensive experiments on one synthetic dataset and several real-world datasets demonstrate the superiority of our proposed method against state-of-the-art baselines to handle spatio-temporal distribution shift.} 

\section*{Acknowledgment}
This work was supported in part by the National Key Research and Development Program of China No. 2020AAA0106300, National Natural Science Foundation of China (No. 62250008, 62222209, 62102222, 62206149), China National Postdoctoral Program for Innovative Talents No. BX20220185 and China Postdoctoral Science Foundation No. 2022M711813. All opinions, findings, conclusions and recommendations in this paper are those of the authors and do not necessarily reflect the views of the funding agencies.

\bibliographystyle{ACM-Reference-Format}
\bibliography{main}


\begin{thebibliography}{146}


\ifx \showCODEN    \undefined \def \showCODEN     #1{\unskip}     \fi
\ifx \showDOI      \undefined \def \showDOI       #1{#1}\fi
\ifx \showISBNx    \undefined \def \showISBNx     #1{\unskip}     \fi
\ifx \showISBNxiii \undefined \def \showISBNxiii  #1{\unskip}     \fi
\ifx \showISSN     \undefined \def \showISSN      #1{\unskip}     \fi
\ifx \showLCCN     \undefined \def \showLCCN      #1{\unskip}     \fi
\ifx \shownote     \undefined \def \shownote      #1{#1}          \fi
\ifx \showarticletitle \undefined \def \showarticletitle #1{#1}   \fi
\ifx \showURL      \undefined \def \showURL       {\relax}        \fi
\providecommand\bibfield[2]{#2}
\providecommand\bibinfo[2]{#2}
\providecommand\natexlab[1]{#1}
\providecommand\showeprint[2][]{arXiv:#2}

\bibitem[\protect\citeauthoryear{Ahuja, Shanmugam, Varshney, and Dhurandhar}{Ahuja et~al\mbox{.}}{2020}]%
        {ahuja2020invariant}
\bibfield{author}{\bibinfo{person}{Kartik Ahuja}, \bibinfo{person}{Karthikeyan Shanmugam}, \bibinfo{person}{Kush Varshney}, {and} \bibinfo{person}{Amit Dhurandhar}.} \bibinfo{year}{2020}\natexlab{}.
\newblock \showarticletitle{Invariant risk minimization games}. In \bibinfo{booktitle}{\emph{International Conference on Machine Learning}}. PMLR, \bibinfo{pages}{145--155}.
\newblock


\bibitem[\protect\citeauthoryear{Arjovsky, Bottou, Gulrajani, and Lopez-Paz}{Arjovsky et~al\mbox{.}}{2019}]%
        {arjovsky2019invariant}
\bibfield{author}{\bibinfo{person}{Martin Arjovsky}, \bibinfo{person}{L{\'e}on Bottou}, \bibinfo{person}{Ishaan Gulrajani}, {and} \bibinfo{person}{David Lopez-Paz}.} \bibinfo{year}{2019}\natexlab{}.
\newblock \showarticletitle{Invariant risk minimization}.
\newblock \bibinfo{journal}{\emph{arXiv preprint}} (\bibinfo{year}{2019}).
\newblock


\bibitem[\protect\citeauthoryear{Ba, Kiros, and Hinton}{Ba et~al\mbox{.}}{2016}]%
        {ba2016layer}
\bibfield{author}{\bibinfo{person}{Jimmy~Lei Ba}, \bibinfo{person}{Jamie~Ryan Kiros}, {and} \bibinfo{person}{Geoffrey~E Hinton}.} \bibinfo{year}{2016}\natexlab{}.
\newblock \showarticletitle{Layer normalization}.
\newblock \bibinfo{journal}{\emph{arXiv preprint arXiv:1607.06450}} (\bibinfo{year}{2016}).
\newblock


\bibitem[\protect\citeauthoryear{Bai, Nie, Zhang, Zhao, and Yuan}{Bai et~al\mbox{.}}{2023}]%
        {bai2023hgwavenet}
\bibfield{author}{\bibinfo{person}{Qijie Bai}, \bibinfo{person}{Changli Nie}, \bibinfo{person}{Haiwei Zhang}, \bibinfo{person}{Dongming Zhao}, {and} \bibinfo{person}{Xiaojie Yuan}.} \bibinfo{year}{2023}\natexlab{}.
\newblock \showarticletitle{HGWaveNet: A Hyperbolic Graph Neural Network for Temporal Link Prediction}. In \bibinfo{booktitle}{\emph{Proceedings of the ACM Web Conference 2023}}. \bibinfo{pages}{523--532}.
\newblock


\bibitem[\protect\citeauthoryear{Barrat, Barthelemy, Pastor-Satorras, and Vespignani}{Barrat et~al\mbox{.}}{2004}]%
        {barrat2004architecture}
\bibfield{author}{\bibinfo{person}{Alain Barrat}, \bibinfo{person}{Marc Barthelemy}, \bibinfo{person}{Romualdo Pastor-Satorras}, {and} \bibinfo{person}{Alessandro Vespignani}.} \bibinfo{year}{2004}\natexlab{}.
\newblock \showarticletitle{The architecture of complex weighted networks}.
\newblock \bibinfo{journal}{\emph{Proceedings of the national academy of sciences}} \bibinfo{volume}{101}, \bibinfo{number}{11} (\bibinfo{year}{2004}), \bibinfo{pages}{3747--3752}.
\newblock


\bibitem[\protect\citeauthoryear{Bengio, Courville, and Vincent}{Bengio et~al\mbox{.}}{2013}]%
        {bengio2013representation}
\bibfield{author}{\bibinfo{person}{Yoshua Bengio}, \bibinfo{person}{Aaron Courville}, {and} \bibinfo{person}{Pascal Vincent}.} \bibinfo{year}{2013}\natexlab{}.
\newblock \showarticletitle{Representation learning: A review and new perspectives}.
\newblock \bibinfo{journal}{\emph{IEEE transactions on pattern analysis and machine intelligence}} \bibinfo{volume}{35}, \bibinfo{number}{8} (\bibinfo{year}{2013}), \bibinfo{pages}{1798--1828}.
\newblock


\bibitem[\protect\citeauthoryear{Benson, Gleich, and Leskovec}{Benson et~al\mbox{.}}{2016}]%
        {benson2016higher}
\bibfield{author}{\bibinfo{person}{Austin~R Benson}, \bibinfo{person}{David~F Gleich}, {and} \bibinfo{person}{Jure Leskovec}.} \bibinfo{year}{2016}\natexlab{}.
\newblock \showarticletitle{Higher-order organization of complex networks}.
\newblock \bibinfo{journal}{\emph{Science}} \bibinfo{volume}{353}, \bibinfo{number}{6295} (\bibinfo{year}{2016}), \bibinfo{pages}{163--166}.
\newblock


\bibitem[\protect\citeauthoryear{Berger-Wolf and Saia}{Berger-Wolf and Saia}{2006}]%
        {berger2006framework}
\bibfield{author}{\bibinfo{person}{Tanya~Y Berger-Wolf} {and} \bibinfo{person}{Jared Saia}.} \bibinfo{year}{2006}\natexlab{}.
\newblock \showarticletitle{A framework for analysis of dynamic social networks}. In \bibinfo{booktitle}{\emph{Proceedings of the 12th ACM SIGKDD international conference on Knowledge discovery and data mining}}. \bibinfo{pages}{523--528}.
\newblock


\bibitem[\protect\citeauthoryear{Berk}{Berk}{1983}]%
        {berk1983introduction}
\bibfield{author}{\bibinfo{person}{Richard~A Berk}.} \bibinfo{year}{1983}\natexlab{}.
\newblock \showarticletitle{An introduction to sample selection bias in sociological data}.
\newblock \bibinfo{journal}{\emph{American sociological review}} (\bibinfo{year}{1983}), \bibinfo{pages}{386--398}.
\newblock


\bibitem[\protect\citeauthoryear{Bevilacqua, Zhou, and Ribeiro}{Bevilacqua et~al\mbox{.}}{2021}]%
        {bevilacqua2021size}
\bibfield{author}{\bibinfo{person}{Beatrice Bevilacqua}, \bibinfo{person}{Yangze Zhou}, {and} \bibinfo{person}{Bruno Ribeiro}.} \bibinfo{year}{2021}\natexlab{}.
\newblock \showarticletitle{Size-invariant graph representations for graph classification extrapolations}. In \bibinfo{booktitle}{\emph{International Conference on Machine Learning}}. \bibinfo{pages}{837--851}.
\newblock


\bibitem[\protect\citeauthoryear{Bi, Xu, Sun, Xu, Shen, and Cheng}{Bi et~al\mbox{.}}{2023}]%
        {bi2023predicting}
\bibfield{author}{\bibinfo{person}{Wendong Bi}, \bibinfo{person}{Bingbing Xu}, \bibinfo{person}{Xiaoqian Sun}, \bibinfo{person}{Li Xu}, \bibinfo{person}{Huawei Shen}, {and} \bibinfo{person}{Xueqi Cheng}.} \bibinfo{year}{2023}\natexlab{}.
\newblock \showarticletitle{Predicting the silent majority on graphs: Knowledge transferable graph neural network}. In \bibinfo{booktitle}{\emph{Proceedings of the ACM Web Conference 2023}}. \bibinfo{pages}{274--285}.
\newblock


\bibitem[\protect\citeauthoryear{Brown, Goetzmann, Ibbotson, and Ross}{Brown et~al\mbox{.}}{1992}]%
        {brown1992survivorship}
\bibfield{author}{\bibinfo{person}{Stephen~J Brown}, \bibinfo{person}{William Goetzmann}, \bibinfo{person}{Roger~G Ibbotson}, {and} \bibinfo{person}{Stephen~A Ross}.} \bibinfo{year}{1992}\natexlab{}.
\newblock \showarticletitle{Survivorship bias in performance studies}.
\newblock \bibinfo{journal}{\emph{The Review of Financial Studies}} \bibinfo{volume}{5}, \bibinfo{number}{4} (\bibinfo{year}{1992}), \bibinfo{pages}{553--580}.
\newblock


\bibitem[\protect\citeauthoryear{Cadene, Dancette, Cord, Parikh, et~al\mbox{.}}{Cadene et~al\mbox{.}}{2019}]%
        {cadene2019rubi}
\bibfield{author}{\bibinfo{person}{Remi Cadene}, \bibinfo{person}{Corentin Dancette}, \bibinfo{person}{Matthieu Cord}, \bibinfo{person}{Devi Parikh}, {et~al\mbox{.}}} \bibinfo{year}{2019}\natexlab{}.
\newblock \showarticletitle{Rubi: Reducing unimodal biases for visual question answering}.
\newblock \bibinfo{journal}{\emph{Advances in neural information processing systems}} (\bibinfo{year}{2019}).
\newblock


\bibitem[\protect\citeauthoryear{Cai, Qian, Fang, Hu, and Xu}{Cai et~al\mbox{.}}{2023}]%
        {cai2023user}
\bibfield{author}{\bibinfo{person}{Desheng Cai}, \bibinfo{person}{Shengsheng Qian}, \bibinfo{person}{Quan Fang}, \bibinfo{person}{Jun Hu}, {and} \bibinfo{person}{Changsheng Xu}.} \bibinfo{year}{2023}\natexlab{}.
\newblock \showarticletitle{User cold-start recommendation via inductive heterogeneous graph neural network}.
\newblock \bibinfo{journal}{\emph{ACM Transactions on Information Systems (TOIS)}} \bibinfo{volume}{41}, \bibinfo{number}{3} (\bibinfo{year}{2023}), \bibinfo{pages}{1--27}.
\newblock


\bibitem[\protect\citeauthoryear{Cai, Chen, Luo, Gui, Ni, Li, and Chen}{Cai et~al\mbox{.}}{2021}]%
        {cai2021structural}
\bibfield{author}{\bibinfo{person}{Lei Cai}, \bibinfo{person}{Zhengzhang Chen}, \bibinfo{person}{Chen Luo}, \bibinfo{person}{Jiaping Gui}, \bibinfo{person}{Jingchao Ni}, \bibinfo{person}{Ding Li}, {and} \bibinfo{person}{Haifeng Chen}.} \bibinfo{year}{2021}\natexlab{}.
\newblock \showarticletitle{Structural temporal graph neural networks for anomaly detection in dynamic graphs}. In \bibinfo{booktitle}{\emph{Proceedings of the 30th ACM international conference on Information \& Knowledge Management}}. \bibinfo{pages}{3747--3756}.
\newblock


\bibitem[\protect\citeauthoryear{Chang, Zhang, Yu, and Jaakkola}{Chang et~al\mbox{.}}{2020b}]%
        {chang2020invariant}
\bibfield{author}{\bibinfo{person}{Shiyu Chang}, \bibinfo{person}{Yang Zhang}, \bibinfo{person}{Mo Yu}, {and} \bibinfo{person}{Tommi Jaakkola}.} \bibinfo{year}{2020}\natexlab{b}.
\newblock \showarticletitle{Invariant rationalization}. In \bibinfo{booktitle}{\emph{International Conference on Machine Learning}}. PMLR, \bibinfo{pages}{1448--1458}.
\newblock


\bibitem[\protect\citeauthoryear{Chang, Liu, Wen, Li, Fang, Song, and Qi}{Chang et~al\mbox{.}}{2020a}]%
        {chang2020continuous}
\bibfield{author}{\bibinfo{person}{Xiaofu Chang}, \bibinfo{person}{Xuqin Liu}, \bibinfo{person}{Jianfeng Wen}, \bibinfo{person}{Shuang Li}, \bibinfo{person}{Yanming Fang}, \bibinfo{person}{Le Song}, {and} \bibinfo{person}{Yuan Qi}.} \bibinfo{year}{2020}\natexlab{a}.
\newblock \showarticletitle{Continuous-time dynamic graph learning via neural interaction processes}. In \bibinfo{booktitle}{\emph{Proceedings of the 29th ACM International Conference on Information \& Knowledge Management}}. \bibinfo{pages}{145--154}.
\newblock


\bibitem[\protect\citeauthoryear{Chen, Ye, Wang, and Gao}{Chen et~al\mbox{.}}{2022a}]%
        {chen2022learning}
\bibfield{author}{\bibinfo{person}{Cen Chen}, \bibinfo{person}{Tiandi Ye}, \bibinfo{person}{Li Wang}, {and} \bibinfo{person}{Ming Gao}.} \bibinfo{year}{2022}\natexlab{a}.
\newblock \showarticletitle{Learning to generalize in heterogeneous federated networks}. In \bibinfo{booktitle}{\emph{Proceedings of the 31st ACM International Conference on Information \& Knowledge Management}}. \bibinfo{pages}{159--168}.
\newblock


\bibitem[\protect\citeauthoryear{Chen, Chen, Wang, Xie, Wang, Xia, and Zhu}{Chen et~al\mbox{.}}{2021}]%
        {chen2021curriculum}
\bibfield{author}{\bibinfo{person}{Hong Chen}, \bibinfo{person}{Yudong Chen}, \bibinfo{person}{Xin Wang}, \bibinfo{person}{Ruobing Xie}, \bibinfo{person}{Rui Wang}, \bibinfo{person}{Feng Xia}, {and} \bibinfo{person}{Wenwu Zhu}.} \bibinfo{year}{2021}\natexlab{}.
\newblock \showarticletitle{Curriculum Disentangled Recommendation with Noisy Multi-feedback}.
\newblock \bibinfo{journal}{\emph{Advances in Neural Information Processing Systems}}  \bibinfo{volume}{34} (\bibinfo{year}{2021}), \bibinfo{pages}{26924--26936}.
\newblock


\bibitem[\protect\citeauthoryear{Chen, Duan, Houthooft, Schulman, Sutskever, and Abbeel}{Chen et~al\mbox{.}}{2016}]%
        {chen2016infogan}
\bibfield{author}{\bibinfo{person}{Xi Chen}, \bibinfo{person}{Yan Duan}, \bibinfo{person}{Rein Houthooft}, \bibinfo{person}{John Schulman}, \bibinfo{person}{Ilya Sutskever}, {and} \bibinfo{person}{Pieter Abbeel}.} \bibinfo{year}{2016}\natexlab{}.
\newblock \showarticletitle{Infogan: Interpretable representation learning by information maximizing generative adversarial nets}.
\newblock \bibinfo{journal}{\emph{Advances in neural information processing systems}}  \bibinfo{volume}{29} (\bibinfo{year}{2016}).
\newblock


\bibitem[\protect\citeauthoryear{Chen, Xiong, Zhang, Xia, Yin, and Huang}{Chen et~al\mbox{.}}{2020}]%
        {chen2020neural}
\bibfield{author}{\bibinfo{person}{Xu Chen}, \bibinfo{person}{Kun Xiong}, \bibinfo{person}{Yongfeng Zhang}, \bibinfo{person}{Long Xia}, \bibinfo{person}{Dawei Yin}, {and} \bibinfo{person}{Jimmy~Xiangji Huang}.} \bibinfo{year}{2020}\natexlab{}.
\newblock \showarticletitle{Neural feature-aware recommendation with signed hypergraph convolutional network}.
\newblock \bibinfo{journal}{\emph{ACM Transactions on Information Systems (TOIS)}} \bibinfo{volume}{39}, \bibinfo{number}{1} (\bibinfo{year}{2020}), \bibinfo{pages}{1--22}.
\newblock


\bibitem[\protect\citeauthoryear{Chen, Zhang, Yang, Ma, Xie, Liu, Han, and Cheng}{Chen et~al\mbox{.}}{2022b}]%
        {chen2022invariance}
\bibfield{author}{\bibinfo{person}{Yongqiang Chen}, \bibinfo{person}{Yonggang Zhang}, \bibinfo{person}{Han Yang}, \bibinfo{person}{Kaili Ma}, \bibinfo{person}{Binghui Xie}, \bibinfo{person}{Tongliang Liu}, \bibinfo{person}{Bo Han}, {and} \bibinfo{person}{James Cheng}.} \bibinfo{year}{2022}\natexlab{b}.
\newblock \showarticletitle{Invariance Principle Meets Out-of-Distribution Generalization on Graphs}.
\newblock \bibinfo{journal}{\emph{arXiv preprint}} (\bibinfo{year}{2022}).
\newblock


\bibitem[\protect\citeauthoryear{Cho, van Merrienboer, Çaglar G{\"u}lçehre, Bahdanau, Bougares, Schwenk, and Bengio}{Cho et~al\mbox{.}}{2014}]%
        {Cho2014LearningPR}
\bibfield{author}{\bibinfo{person}{Kyunghyun Cho}, \bibinfo{person}{Bart van Merrienboer}, \bibinfo{person}{Çaglar G{\"u}lçehre}, \bibinfo{person}{Dzmitry Bahdanau}, \bibinfo{person}{Fethi Bougares}, \bibinfo{person}{Holger Schwenk}, {and} \bibinfo{person}{Yoshua Bengio}.} \bibinfo{year}{2014}\natexlab{}.
\newblock \showarticletitle{Learning Phrase Representations using RNN Encoder–Decoder for Statistical Machine Translation}. In \bibinfo{booktitle}{\emph{EMNLP}}.
\newblock


\bibitem[\protect\citeauthoryear{Coleman}{Coleman}{1994}]%
        {coleman1994foundations}
\bibfield{author}{\bibinfo{person}{James~S Coleman}.} \bibinfo{year}{1994}\natexlab{}.
\newblock \bibinfo{booktitle}{\emph{Foundations of social theory}}.
\newblock \bibinfo{publisher}{Harvard university press}.
\newblock


\bibitem[\protect\citeauthoryear{Cong, Wu, Tian, Gu, Xia, Mahdavi, and Chen}{Cong et~al\mbox{.}}{2021}]%
        {cong2021dynamic}
\bibfield{author}{\bibinfo{person}{Weilin Cong}, \bibinfo{person}{Yanhong Wu}, \bibinfo{person}{Yuandong Tian}, \bibinfo{person}{Mengting Gu}, \bibinfo{person}{Yinglong Xia}, \bibinfo{person}{Mehrdad Mahdavi}, {and} \bibinfo{person}{Chun-cheng~Jason Chen}.} \bibinfo{year}{2021}\natexlab{}.
\newblock \showarticletitle{Dynamic Graph Representation Learning via Graph Transformer Networks}.
\newblock \bibinfo{journal}{\emph{arXiv preprint arXiv:2111.10447}} (\bibinfo{year}{2021}).
\newblock


\bibitem[\protect\citeauthoryear{Deng, Rangwala, and Ning}{Deng et~al\mbox{.}}{2020}]%
        {deng2020dynamic}
\bibfield{author}{\bibinfo{person}{Songgaojun Deng}, \bibinfo{person}{Huzefa Rangwala}, {and} \bibinfo{person}{Yue Ning}.} \bibinfo{year}{2020}\natexlab{}.
\newblock \showarticletitle{Dynamic knowledge graph based multi-event forecasting}. In \bibinfo{booktitle}{\emph{Proceedings of the 26th ACM SIGKDD International Conference on Knowledge Discovery \& Data Mining}}. \bibinfo{pages}{1585--1595}.
\newblock


\bibitem[\protect\citeauthoryear{Denton et~al\mbox{.}}{Denton et~al\mbox{.}}{2017}]%
        {denton2017unsupervised}
\bibfield{author}{\bibinfo{person}{Emily~L Denton} {et~al\mbox{.}}} \bibinfo{year}{2017}\natexlab{}.
\newblock \showarticletitle{Unsupervised learning of disentangled representations from video}.
\newblock \bibinfo{journal}{\emph{Advances in neural information processing systems}}  \bibinfo{volume}{30} (\bibinfo{year}{2017}).
\newblock


\bibitem[\protect\citeauthoryear{Ding, Kong, Chen, Kirchenbauer, Goldblum, Wipf, Huang, and Goldstein}{Ding et~al\mbox{.}}{2021}]%
        {ding2021closer}
\bibfield{author}{\bibinfo{person}{Mucong Ding}, \bibinfo{person}{Kezhi Kong}, \bibinfo{person}{Jiuhai Chen}, \bibinfo{person}{John Kirchenbauer}, \bibinfo{person}{Micah Goldblum}, \bibinfo{person}{David Wipf}, \bibinfo{person}{Furong Huang}, {and} \bibinfo{person}{Tom Goldstein}.} \bibinfo{year}{2021}\natexlab{}.
\newblock \showarticletitle{A Closer Look at Distribution Shifts and Out-of-Distribution Generalization on Graphs}.
\newblock  (\bibinfo{year}{2021}).
\newblock


\bibitem[\protect\citeauthoryear{Du, Guo, Cao, Ye, and Zhao}{Du et~al\mbox{.}}{2022}]%
        {du2022disentangled}
\bibfield{author}{\bibinfo{person}{Yuanqi Du}, \bibinfo{person}{Xiaojie Guo}, \bibinfo{person}{Hengning Cao}, \bibinfo{person}{Yanfang Ye}, {and} \bibinfo{person}{Liang Zhao}.} \bibinfo{year}{2022}\natexlab{}.
\newblock \showarticletitle{Disentangled Spatiotemporal Graph Generative Models}. In \bibinfo{booktitle}{\emph{Thirty-Sixth {AAAI} Conference on Artificial Intelligence}}. \bibinfo{publisher}{{AAAI} Press}, \bibinfo{pages}{6541--6549}.
\newblock


\bibitem[\protect\citeauthoryear{Du, Wang, Feng, Pan, Qin, Xu, and Wang}{Du et~al\mbox{.}}{2021}]%
        {du2021adarnn}
\bibfield{author}{\bibinfo{person}{Yuntao Du}, \bibinfo{person}{Jindong Wang}, \bibinfo{person}{Wenjie Feng}, \bibinfo{person}{Sinno Pan}, \bibinfo{person}{Tao Qin}, \bibinfo{person}{Renjun Xu}, {and} \bibinfo{person}{Chongjun Wang}.} \bibinfo{year}{2021}\natexlab{}.
\newblock \showarticletitle{Adarnn: Adaptive learning and forecasting of time series}. In \bibinfo{booktitle}{\emph{Proceedings of the 30th ACM International Conference on Information \& Knowledge Management}}. \bibinfo{pages}{402--411}.
\newblock


\bibitem[\protect\citeauthoryear{Fan, Wang, Shi, Cui, and Wang}{Fan et~al\mbox{.}}{2021}]%
        {fan2021generalizing}
\bibfield{author}{\bibinfo{person}{Shaohua Fan}, \bibinfo{person}{Xiao Wang}, \bibinfo{person}{Chuan Shi}, \bibinfo{person}{Peng Cui}, {and} \bibinfo{person}{Bai Wang}.} \bibinfo{year}{2021}\natexlab{}.
\newblock \showarticletitle{Generalizing Graph Neural Networks on Out-Of-Distribution Graphs}.
\newblock \bibinfo{journal}{\emph{arXiv preprint arXiv:2111.10657}} (\bibinfo{year}{2021}).
\newblock


\bibitem[\protect\citeauthoryear{Fey and Lenssen}{Fey and Lenssen}{2019}]%
        {Fey/Lenssen/2019}
\bibfield{author}{\bibinfo{person}{Matthias Fey} {and} \bibinfo{person}{Jan~E. Lenssen}.} \bibinfo{year}{2019}\natexlab{}.
\newblock \showarticletitle{Fast Graph Representation Learning with {PyTorch Geometric}}. In \bibinfo{booktitle}{\emph{ICLR Workshop on Representation Learning on Graphs and Manifolds}}.
\newblock


\bibitem[\protect\citeauthoryear{Gagnon-Audet, Ahuja, Darvishi-Bayazi, Dumas, and Rish}{Gagnon-Audet et~al\mbox{.}}{2022}]%
        {gagnon2022woods}
\bibfield{author}{\bibinfo{person}{Jean-Christophe Gagnon-Audet}, \bibinfo{person}{Kartik Ahuja}, \bibinfo{person}{Mohammad-Javad Darvishi-Bayazi}, \bibinfo{person}{Guillaume Dumas}, {and} \bibinfo{person}{Irina Rish}.} \bibinfo{year}{2022}\natexlab{}.
\newblock \showarticletitle{WOODS: Benchmarks for Out-of-Distribution Generalization in Time Series Tasks}.
\newblock \bibinfo{journal}{\emph{arXiv preprint arXiv:2203.09978}} (\bibinfo{year}{2022}).
\newblock


\bibitem[\protect\citeauthoryear{Gao, Wang, He, Liu, Feng, and Zhang}{Gao et~al\mbox{.}}{2023}]%
        {gao2023alleviating}
\bibfield{author}{\bibinfo{person}{Yuan Gao}, \bibinfo{person}{Xiang Wang}, \bibinfo{person}{Xiangnan He}, \bibinfo{person}{Zhenguang Liu}, \bibinfo{person}{Huamin Feng}, {and} \bibinfo{person}{Yongdong Zhang}.} \bibinfo{year}{2023}\natexlab{}.
\newblock \showarticletitle{Alleviating structural distribution shift in graph anomaly detection}. In \bibinfo{booktitle}{\emph{Proceedings of the Sixteenth ACM International Conference on Web Search and Data Mining}}. \bibinfo{pages}{357--365}.
\newblock


\bibitem[\protect\citeauthoryear{Glymour, Pearl, and Jewell}{Glymour et~al\mbox{.}}{2016}]%
        {glymour2016causal}
\bibfield{author}{\bibinfo{person}{Madelyn Glymour}, \bibinfo{person}{Judea Pearl}, {and} \bibinfo{person}{Nicholas~P Jewell}.} \bibinfo{year}{2016}\natexlab{}.
\newblock \bibinfo{booktitle}{\emph{Causal inference in statistics: A primer}}.
\newblock \bibinfo{publisher}{John Wiley \& Sons}.
\newblock


\bibitem[\protect\citeauthoryear{Greene, Doyle, and Cunningham}{Greene et~al\mbox{.}}{2010}]%
        {greene2010tracking}
\bibfield{author}{\bibinfo{person}{Derek Greene}, \bibinfo{person}{Donal Doyle}, {and} \bibinfo{person}{Padraig Cunningham}.} \bibinfo{year}{2010}\natexlab{}.
\newblock \showarticletitle{Tracking the evolution of communities in dynamic social networks}. In \bibinfo{booktitle}{\emph{2010 international conference on advances in social networks analysis and mining}}. IEEE, \bibinfo{pages}{176--183}.
\newblock


\bibitem[\protect\citeauthoryear{Hajiramezanali, Hasanzadeh, Narayanan, Duffield, Zhou, and Qian}{Hajiramezanali et~al\mbox{.}}{2019}]%
        {hajiramezanali2019variational}
\bibfield{author}{\bibinfo{person}{Ehsan Hajiramezanali}, \bibinfo{person}{Arman Hasanzadeh}, \bibinfo{person}{Krishna Narayanan}, \bibinfo{person}{Nick Duffield}, \bibinfo{person}{Mingyuan Zhou}, {and} \bibinfo{person}{Xiaoning Qian}.} \bibinfo{year}{2019}\natexlab{}.
\newblock \showarticletitle{Variational graph recurrent neural networks}.
\newblock \bibinfo{journal}{\emph{Advances in neural information processing systems}}  \bibinfo{volume}{32} (\bibinfo{year}{2019}).
\newblock


\bibitem[\protect\citeauthoryear{Han, Jiang, Liu, and Hu}{Han et~al\mbox{.}}{2022}]%
        {han2022g}
\bibfield{author}{\bibinfo{person}{Xiaotian Han}, \bibinfo{person}{Zhimeng Jiang}, \bibinfo{person}{Ninghao Liu}, {and} \bibinfo{person}{Xia Hu}.} \bibinfo{year}{2022}\natexlab{}.
\newblock \showarticletitle{G-mixup: Graph data augmentation for graph classification}. In \bibinfo{booktitle}{\emph{International Conference on Machine Learning}}. \bibinfo{pages}{8230--8248}.
\newblock


\bibitem[\protect\citeauthoryear{Hochreiter and Schmidhuber}{Hochreiter and Schmidhuber}{1997}]%
        {hochreiter1997long}
\bibfield{author}{\bibinfo{person}{Sepp Hochreiter} {and} \bibinfo{person}{J{\"u}rgen Schmidhuber}.} \bibinfo{year}{1997}\natexlab{}.
\newblock \showarticletitle{Long short-term memory}.
\newblock \bibinfo{journal}{\emph{Neural computation}} \bibinfo{volume}{9}, \bibinfo{number}{8} (\bibinfo{year}{1997}), \bibinfo{pages}{1735--1780}.
\newblock


\bibitem[\protect\citeauthoryear{Hsieh, Liu, Huang, Fei-Fei, and Niebles}{Hsieh et~al\mbox{.}}{2018}]%
        {hsieh2018learning}
\bibfield{author}{\bibinfo{person}{Jun-Ting Hsieh}, \bibinfo{person}{Bingbin Liu}, \bibinfo{person}{De-An Huang}, \bibinfo{person}{Li~F Fei-Fei}, {and} \bibinfo{person}{Juan~Carlos Niebles}.} \bibinfo{year}{2018}\natexlab{}.
\newblock \showarticletitle{Learning to decompose and disentangle representations for video prediction}.
\newblock \bibinfo{journal}{\emph{Advances in neural information processing systems}}  \bibinfo{volume}{31} (\bibinfo{year}{2018}).
\newblock


\bibitem[\protect\citeauthoryear{Hu, Fey, Zitnik, Dong, Ren, Liu, Catasta, and Leskovec}{Hu et~al\mbox{.}}{2020}]%
        {hu2020open}
\bibfield{author}{\bibinfo{person}{Weihua Hu}, \bibinfo{person}{Matthias Fey}, \bibinfo{person}{Marinka Zitnik}, \bibinfo{person}{Yuxiao Dong}, \bibinfo{person}{Hongyu Ren}, \bibinfo{person}{Bowen Liu}, \bibinfo{person}{Michele Catasta}, {and} \bibinfo{person}{Jure Leskovec}.} \bibinfo{year}{2020}\natexlab{}.
\newblock \showarticletitle{Open graph benchmark: Datasets for machine learning on graphs}.
\newblock \bibinfo{journal}{\emph{Advances in neural information processing systems}}  \bibinfo{volume}{33} (\bibinfo{year}{2020}), \bibinfo{pages}{22118--22133}.
\newblock


\bibitem[\protect\citeauthoryear{Huang, Fang, Wang, Miao, and Jin}{Huang et~al\mbox{.}}{2020}]%
        {huang2020motif}
\bibfield{author}{\bibinfo{person}{Hong Huang}, \bibinfo{person}{Zixuan Fang}, \bibinfo{person}{Xiao Wang}, \bibinfo{person}{Youshan Miao}, {and} \bibinfo{person}{Hai Jin}.} \bibinfo{year}{2020}\natexlab{}.
\newblock \showarticletitle{Motif-Preserving Temporal Network Embedding.}. In \bibinfo{booktitle}{\emph{IJCAI}}. \bibinfo{pages}{1237--1243}.
\newblock


\bibitem[\protect\citeauthoryear{Huang, Tang, Liu, Luo, and Fu}{Huang et~al\mbox{.}}{2015}]%
        {huang2015triadic}
\bibfield{author}{\bibinfo{person}{Hong Huang}, \bibinfo{person}{Jie Tang}, \bibinfo{person}{Lu Liu}, \bibinfo{person}{JarDer Luo}, {and} \bibinfo{person}{Xiaoming Fu}.} \bibinfo{year}{2015}\natexlab{}.
\newblock \showarticletitle{Triadic closure pattern analysis and prediction in social networks}.
\newblock \bibinfo{journal}{\emph{IEEE Transactions on Knowledge and Data Engineering}} \bibinfo{volume}{27}, \bibinfo{number}{12} (\bibinfo{year}{2015}), \bibinfo{pages}{3374--3389}.
\newblock


\bibitem[\protect\citeauthoryear{Huang and Zitnik}{Huang and Zitnik}{2020}]%
        {huang2020graph}
\bibfield{author}{\bibinfo{person}{Kexin Huang} {and} \bibinfo{person}{Marinka Zitnik}.} \bibinfo{year}{2020}\natexlab{}.
\newblock \showarticletitle{Graph meta learning via local subgraphs}.
\newblock \bibinfo{journal}{\emph{Advances in Neural Information Processing Systems}}  \bibinfo{volume}{33} (\bibinfo{year}{2020}), \bibinfo{pages}{5862--5874}.
\newblock


\bibitem[\protect\citeauthoryear{Huang, Ma, Liu, Danny~Du, Wang, and Li}{Huang et~al\mbox{.}}{2023}]%
        {huang2023position}
\bibfield{author}{\bibinfo{person}{Liwei Huang}, \bibinfo{person}{Yutao Ma}, \bibinfo{person}{Yanbo Liu}, \bibinfo{person}{Bohong Danny~Du}, \bibinfo{person}{Shuliang Wang}, {and} \bibinfo{person}{Deyi Li}.} \bibinfo{year}{2023}\natexlab{}.
\newblock \showarticletitle{Position-enhanced and time-aware graph convolutional network for sequential recommendations}.
\newblock \bibinfo{journal}{\emph{ACM Transactions on Information Systems (TOIS)}} \bibinfo{volume}{41}, \bibinfo{number}{1} (\bibinfo{year}{2023}), \bibinfo{pages}{1--32}.
\newblock


\bibitem[\protect\citeauthoryear{Huang, Sun, and Wang}{Huang et~al\mbox{.}}{2021}]%
        {huang2021coupled}
\bibfield{author}{\bibinfo{person}{Zijie Huang}, \bibinfo{person}{Yizhou Sun}, {and} \bibinfo{person}{Wei Wang}.} \bibinfo{year}{2021}\natexlab{}.
\newblock \showarticletitle{Coupled Graph ODE for Learning Interacting System Dynamics.}. In \bibinfo{booktitle}{\emph{KDD}}. \bibinfo{pages}{705--715}.
\newblock


\bibitem[\protect\citeauthoryear{Jin, Wu, Ou, and Yu}{Jin et~al\mbox{.}}{2021}]%
        {jin2021community}
\bibfield{author}{\bibinfo{person}{Tian Jin}, \bibinfo{person}{Qiong Wu}, \bibinfo{person}{Xuan Ou}, {and} \bibinfo{person}{Jianjun Yu}.} \bibinfo{year}{2021}\natexlab{}.
\newblock \showarticletitle{Community detection and co-author recommendation in co-author networks}.
\newblock \bibinfo{journal}{\emph{International Journal of Machine Learning and Cybernetics}} \bibinfo{volume}{12}, \bibinfo{number}{2} (\bibinfo{year}{2021}), \bibinfo{pages}{597--609}.
\newblock


\bibitem[\protect\citeauthoryear{Kim, Kim, Tae, Park, Choi, and Choo}{Kim et~al\mbox{.}}{2021}]%
        {kim2021reversible}
\bibfield{author}{\bibinfo{person}{Taesung Kim}, \bibinfo{person}{Jinhee Kim}, \bibinfo{person}{Yunwon Tae}, \bibinfo{person}{Cheonbok Park}, \bibinfo{person}{Jang-Ho Choi}, {and} \bibinfo{person}{Jaegul Choo}.} \bibinfo{year}{2021}\natexlab{}.
\newblock \showarticletitle{Reversible Instance Normalization for Accurate Time-Series Forecasting against Distribution Shift}. In \bibinfo{booktitle}{\emph{International Conference on Learning Representations}}.
\newblock


\bibitem[\protect\citeauthoryear{Kingma and Ba}{Kingma and Ba}{2015}]%
        {kingma2014adam}
\bibfield{author}{\bibinfo{person}{Diederik~P. Kingma} {and} \bibinfo{person}{Jimmy Ba}.} \bibinfo{year}{2015}\natexlab{}.
\newblock \showarticletitle{Adam: {A} Method for Stochastic Optimization}. In \bibinfo{booktitle}{\emph{3rd International Conference on Learning Representations}}, \bibfield{editor}{\bibinfo{person}{Yoshua Bengio} {and} \bibinfo{person}{Yann LeCun}} (Eds.).
\newblock


\bibitem[\protect\citeauthoryear{Kipf and Welling}{Kipf and Welling}{2016}]%
        {kipf2016variational}
\bibfield{author}{\bibinfo{person}{Thomas~N Kipf} {and} \bibinfo{person}{Max Welling}.} \bibinfo{year}{2016}\natexlab{}.
\newblock \showarticletitle{Variational graph auto-encoders}.
\newblock \bibinfo{journal}{\emph{arXiv preprint arXiv:1611.07308}} (\bibinfo{year}{2016}).
\newblock


\bibitem[\protect\citeauthoryear{Kipf and Welling}{Kipf and Welling}{2017}]%
        {kipf2016semi}
\bibfield{author}{\bibinfo{person}{Thomas~N. Kipf} {and} \bibinfo{person}{Max Welling}.} \bibinfo{year}{2017}\natexlab{}.
\newblock \showarticletitle{Semi-Supervised Classification with Graph Convolutional Networks}. In \bibinfo{booktitle}{\emph{5th International Conference on Learning Representations}}. \bibinfo{publisher}{OpenReview.net}.
\newblock


\bibitem[\protect\citeauthoryear{Kovanen, Karsai, Kaski, Kert{\'e}sz, and Saram{\"a}ki}{Kovanen et~al\mbox{.}}{2011}]%
        {kovanen2011temporal}
\bibfield{author}{\bibinfo{person}{Lauri Kovanen}, \bibinfo{person}{M{\'a}rton Karsai}, \bibinfo{person}{Kimmo Kaski}, \bibinfo{person}{J{\'a}nos Kert{\'e}sz}, {and} \bibinfo{person}{Jari Saram{\"a}ki}.} \bibinfo{year}{2011}\natexlab{}.
\newblock \showarticletitle{Temporal motifs in time-dependent networks}.
\newblock \bibinfo{journal}{\emph{Journal of Statistical Mechanics: Theory and Experiment}} \bibinfo{volume}{2011}, \bibinfo{number}{11} (\bibinfo{year}{2011}), \bibinfo{pages}{P11005}.
\newblock


\bibitem[\protect\citeauthoryear{Kovanen, Kaski, Kert{\'e}sz, and Saram{\"a}ki}{Kovanen et~al\mbox{.}}{2013}]%
        {kovanen2013temporal}
\bibfield{author}{\bibinfo{person}{Lauri Kovanen}, \bibinfo{person}{Kimmo Kaski}, \bibinfo{person}{J{\'a}nos Kert{\'e}sz}, {and} \bibinfo{person}{Jari Saram{\"a}ki}.} \bibinfo{year}{2013}\natexlab{}.
\newblock \showarticletitle{Temporal motifs reveal homophily, gender-specific patterns, and group talk in call sequences}.
\newblock \bibinfo{journal}{\emph{Proceedings of the National Academy of Sciences}} \bibinfo{volume}{110}, \bibinfo{number}{45} (\bibinfo{year}{2013}), \bibinfo{pages}{18070--18075}.
\newblock


\bibitem[\protect\citeauthoryear{Krueger, Caballero, Jacobsen, Zhang, Binas, Zhang, Le~Priol, and Courville}{Krueger et~al\mbox{.}}{2021}]%
        {krueger2021out}
\bibfield{author}{\bibinfo{person}{David Krueger}, \bibinfo{person}{Ethan Caballero}, \bibinfo{person}{Joern-Henrik Jacobsen}, \bibinfo{person}{Amy Zhang}, \bibinfo{person}{Jonathan Binas}, \bibinfo{person}{Dinghuai Zhang}, \bibinfo{person}{Remi Le~Priol}, {and} \bibinfo{person}{Aaron Courville}.} \bibinfo{year}{2021}\natexlab{}.
\newblock \showarticletitle{Out-of-distribution generalization via risk extrapolation (rex)}. In \bibinfo{booktitle}{\emph{International Conference on Machine Learning}}. \bibinfo{pages}{5815--5826}.
\newblock


\bibitem[\protect\citeauthoryear{Li, Cui, Zang, Zhang, Zhu, and Lin}{Li et~al\mbox{.}}{2019}]%
        {li2019fates}
\bibfield{author}{\bibinfo{person}{Haoyang Li}, \bibinfo{person}{Peng Cui}, \bibinfo{person}{Chengxi Zang}, \bibinfo{person}{Tianyang Zhang}, \bibinfo{person}{Wenwu Zhu}, {and} \bibinfo{person}{Yishi Lin}.} \bibinfo{year}{2019}\natexlab{}.
\newblock \showarticletitle{Fates of Microscopic Social Ecosystems: Keep Alive or Dead?}. In \bibinfo{booktitle}{\emph{Proceedings of the 25th ACM SIGKDD International Conference on Knowledge Discovery \& Data Mining}}. \bibinfo{pages}{668--676}.
\newblock


\bibitem[\protect\citeauthoryear{Li, Wang, Zhang, Ma, Cui, and Zhu}{Li et~al\mbox{.}}{2021a}]%
        {li2021intention}
\bibfield{author}{\bibinfo{person}{Haoyang Li}, \bibinfo{person}{Xin Wang}, \bibinfo{person}{Ziwei Zhang}, \bibinfo{person}{Jianxin Ma}, \bibinfo{person}{Peng Cui}, {and} \bibinfo{person}{Wenwu Zhu}.} \bibinfo{year}{2021}\natexlab{a}.
\newblock \showarticletitle{Intention-aware sequential recommendation with structured intent transition}.
\newblock \bibinfo{journal}{\emph{IEEE Transactions on Knowledge and Data Engineering}} (\bibinfo{year}{2021}).
\newblock


\bibitem[\protect\citeauthoryear{Li, Wang, Zhang, Yuan, Li, and Zhu}{Li et~al\mbox{.}}{2021b}]%
        {li2021disentangled}
\bibfield{author}{\bibinfo{person}{Haoyang Li}, \bibinfo{person}{Xin Wang}, \bibinfo{person}{Ziwei Zhang}, \bibinfo{person}{Zehuan Yuan}, \bibinfo{person}{Hang Li}, {and} \bibinfo{person}{Wenwu Zhu}.} \bibinfo{year}{2021}\natexlab{b}.
\newblock \showarticletitle{Disentangled contrastive learning on graphs}.
\newblock \bibinfo{journal}{\emph{Advances in Neural Information Processing Systems}}  \bibinfo{volume}{34} (\bibinfo{year}{2021}), \bibinfo{pages}{21872--21884}.
\newblock


\bibitem[\protect\citeauthoryear{Li, Wang, Zhang, and Zhu}{Li et~al\mbox{.}}{2022b}]%
        {li2022ood}
\bibfield{author}{\bibinfo{person}{Haoyang Li}, \bibinfo{person}{Xin Wang}, \bibinfo{person}{Ziwei Zhang}, {and} \bibinfo{person}{Wenwu Zhu}.} \bibinfo{year}{2022}\natexlab{b}.
\newblock \showarticletitle{Ood-gnn: Out-of-distribution generalized graph neural network}.
\newblock \bibinfo{journal}{\emph{IEEE Transactions on Knowledge and Data Engineering}} (\bibinfo{year}{2022}).
\newblock


\bibitem[\protect\citeauthoryear{Li, Wang, Zhang, and Zhu}{Li et~al\mbox{.}}{2022c}]%
        {li2022out}
\bibfield{author}{\bibinfo{person}{Haoyang Li}, \bibinfo{person}{Xin Wang}, \bibinfo{person}{Ziwei Zhang}, {and} \bibinfo{person}{Wenwu Zhu}.} \bibinfo{year}{2022}\natexlab{c}.
\newblock \showarticletitle{Out-Of-Distribution Generalization on Graphs: A Survey}.
\newblock \bibinfo{journal}{\emph{arXiv preprint}} (\bibinfo{year}{2022}).
\newblock


\bibitem[\protect\citeauthoryear{Li, Zhang, Wang, and Zhu}{Li et~al\mbox{.}}{2022d}]%
        {li2022disentangled}
\bibfield{author}{\bibinfo{person}{Haoyang Li}, \bibinfo{person}{Ziwei Zhang}, \bibinfo{person}{Xin Wang}, {and} \bibinfo{person}{Wenwu Zhu}.} \bibinfo{year}{2022}\natexlab{d}.
\newblock \showarticletitle{Disentangled Graph Contrastive Learning With Independence Promotion}.
\newblock \bibinfo{journal}{\emph{IEEE Transactions on Knowledge and Data Engineering}} (\bibinfo{year}{2022}).
\newblock


\bibitem[\protect\citeauthoryear{Li, Zhang, Wang, and Zhu}{Li et~al\mbox{.}}{2022e}]%
        {li2022gil}
\bibfield{author}{\bibinfo{person}{Haoyang Li}, \bibinfo{person}{Ziwei Zhang}, \bibinfo{person}{Xin Wang}, {and} \bibinfo{person}{Wenwu Zhu}.} \bibinfo{year}{2022}\natexlab{e}.
\newblock \showarticletitle{Learning Invariant Graph Representations for Out-of-Distribution Generalization}. In \bibinfo{booktitle}{\emph{Thirty-Sixth Conference on Neural Information Processing Systems}}.
\newblock


\bibitem[\protect\citeauthoryear{Li, Zhang, Wang, and Zhu}{Li et~al\mbox{.}}{2023b}]%
        {10.1145/3604427}
\bibfield{author}{\bibinfo{person}{Haoyang Li}, \bibinfo{person}{Ziwei Zhang}, \bibinfo{person}{Xin Wang}, {and} \bibinfo{person}{Wenwu Zhu}.} \bibinfo{year}{2023}\natexlab{b}.
\newblock \showarticletitle{Invariant Node Representation Learning under Distribution Shifts with Multiple Latent Environments}.
\newblock \bibinfo{journal}{\emph{ACM Transactions on Information Systems (TOIS)}} (\bibinfo{date}{jun} \bibinfo{year}{2023}).
\newblock
\showISSN{1046-8188}
\urldef\tempurl%
\url{https://doi.org/10.1145/3604427}
\showDOI{\tempurl}
\newblock
\shownote{Just Accepted}.


\bibitem[\protect\citeauthoryear{Li, Huang, and Zitnik}{Li et~al\mbox{.}}{2022a}]%
        {li2022graph}
\bibfield{author}{\bibinfo{person}{Michelle~M Li}, \bibinfo{person}{Kexin Huang}, {and} \bibinfo{person}{Marinka Zitnik}.} \bibinfo{year}{2022}\natexlab{a}.
\newblock \showarticletitle{Graph representation learning in biomedicine and healthcare}.
\newblock \bibinfo{journal}{\emph{Nature Biomedical Engineering}} \bibinfo{volume}{6}, \bibinfo{number}{12} (\bibinfo{year}{2022}), \bibinfo{pages}{1353--1369}.
\newblock


\bibitem[\protect\citeauthoryear{Li, Hou, and Li}{Li et~al\mbox{.}}{2023a}]%
        {li2023preference}
\bibfield{author}{\bibinfo{person}{Yakun Li}, \bibinfo{person}{Lei Hou}, {and} \bibinfo{person}{Juanzi Li}.} \bibinfo{year}{2023}\natexlab{a}.
\newblock \showarticletitle{Preference-aware Graph Attention Networks for Cross-Domain Recommendations with Collaborative Knowledge Graph}.
\newblock \bibinfo{journal}{\emph{ACM Transactions on Information Systems (TOIS)}} \bibinfo{volume}{41}, \bibinfo{number}{3} (\bibinfo{year}{2023}), \bibinfo{pages}{1--26}.
\newblock


\bibitem[\protect\citeauthoryear{Liu, Hu, Wang, Shi, Zhang, and Zhou}{Liu et~al\mbox{.}}{2022}]%
        {liu2022confidence}
\bibfield{author}{\bibinfo{person}{Hongrui Liu}, \bibinfo{person}{Binbin Hu}, \bibinfo{person}{Xiao Wang}, \bibinfo{person}{Chuan Shi}, \bibinfo{person}{Zhiqiang Zhang}, {and} \bibinfo{person}{Jun Zhou}.} \bibinfo{year}{2022}\natexlab{}.
\newblock \showarticletitle{Confidence may cheat: Self-training on graph neural networks under distribution shift}. In \bibinfo{booktitle}{\emph{Proceedings of the ACM Web Conference 2022}}. \bibinfo{pages}{1248--1258}.
\newblock


\bibitem[\protect\citeauthoryear{Liu, Hu, Cui, Li, and Shen}{Liu et~al\mbox{.}}{2021}]%
        {liu2021heterogeneous}
\bibfield{author}{\bibinfo{person}{Jiashuo Liu}, \bibinfo{person}{Zheyuan Hu}, \bibinfo{person}{Peng Cui}, \bibinfo{person}{Bo Li}, {and} \bibinfo{person}{Zheyan Shen}.} \bibinfo{year}{2021}\natexlab{}.
\newblock \showarticletitle{Heterogeneous risk minimization}. In \bibinfo{booktitle}{\emph{International Conference on Machine Learning}}. PMLR, \bibinfo{pages}{6804--6814}.
\newblock


\bibitem[\protect\citeauthoryear{Liu, Ding, Liu, and Pan}{Liu et~al\mbox{.}}{2023}]%
        {liu2023good}
\bibfield{author}{\bibinfo{person}{Yixin Liu}, \bibinfo{person}{Kaize Ding}, \bibinfo{person}{Huan Liu}, {and} \bibinfo{person}{Shirui Pan}.} \bibinfo{year}{2023}\natexlab{}.
\newblock \showarticletitle{Good-d: On unsupervised graph out-of-distribution detection}. In \bibinfo{booktitle}{\emph{Proceedings of the Sixteenth ACM International Conference on Web Search and Data Mining}}. \bibinfo{pages}{339--347}.
\newblock


\bibitem[\protect\citeauthoryear{Liu, Wang, Wu, and Xiao}{Liu et~al\mbox{.}}{2020}]%
        {liu2020independence}
\bibfield{author}{\bibinfo{person}{Yanbei Liu}, \bibinfo{person}{Xiao Wang}, \bibinfo{person}{Shu Wu}, {and} \bibinfo{person}{Zhitao Xiao}.} \bibinfo{year}{2020}\natexlab{}.
\newblock \showarticletitle{Independence promoted graph disentangled networks}. In \bibinfo{booktitle}{\emph{Proceedings of the AAAI Conference on Artificial Intelligence}}, Vol.~\bibinfo{volume}{34}. \bibinfo{pages}{4916--4923}.
\newblock


\bibitem[\protect\citeauthoryear{Lu, Wang, Chen, and Sun}{Lu et~al\mbox{.}}{2021}]%
        {lu2021diversify}
\bibfield{author}{\bibinfo{person}{Wang Lu}, \bibinfo{person}{Jindong Wang}, \bibinfo{person}{Yiqiang Chen}, {and} \bibinfo{person}{Xinwei Sun}.} \bibinfo{year}{2021}\natexlab{}.
\newblock \showarticletitle{DIVERSIFY to Generalize: Learning Generalized Representations for Time Series Classification}.
\newblock \bibinfo{journal}{\emph{arXiv preprint}} (\bibinfo{year}{2021}).
\newblock


\bibitem[\protect\citeauthoryear{Ma, Cui, Kuang, Wang, and Zhu}{Ma et~al\mbox{.}}{2019a}]%
        {ma2019disentangled}
\bibfield{author}{\bibinfo{person}{Jianxin Ma}, \bibinfo{person}{Peng Cui}, \bibinfo{person}{Kun Kuang}, \bibinfo{person}{Xin Wang}, {and} \bibinfo{person}{Wenwu Zhu}.} \bibinfo{year}{2019}\natexlab{a}.
\newblock \showarticletitle{Disentangled graph convolutional networks}. In \bibinfo{booktitle}{\emph{International conference on machine learning}}. PMLR, \bibinfo{pages}{4212--4221}.
\newblock


\bibitem[\protect\citeauthoryear{Ma, Zhou, Cui, Yang, and Zhu}{Ma et~al\mbox{.}}{2019b}]%
        {ma2019learning}
\bibfield{author}{\bibinfo{person}{Jianxin Ma}, \bibinfo{person}{Chang Zhou}, \bibinfo{person}{Peng Cui}, \bibinfo{person}{Hongxia Yang}, {and} \bibinfo{person}{Wenwu Zhu}.} \bibinfo{year}{2019}\natexlab{b}.
\newblock \showarticletitle{Learning disentangled representations for recommendation}.
\newblock \bibinfo{journal}{\emph{Advances in neural information processing systems}}  \bibinfo{volume}{32} (\bibinfo{year}{2019}).
\newblock


\bibitem[\protect\citeauthoryear{Ma, Zhou, Yang, Cui, Wang, and Zhu}{Ma et~al\mbox{.}}{2020}]%
        {ma2020disentangled}
\bibfield{author}{\bibinfo{person}{Jianxin Ma}, \bibinfo{person}{Chang Zhou}, \bibinfo{person}{Hongxia Yang}, \bibinfo{person}{Peng Cui}, \bibinfo{person}{Xin Wang}, {and} \bibinfo{person}{Wenwu Zhu}.} \bibinfo{year}{2020}\natexlab{}.
\newblock \showarticletitle{Disentangled self-supervision in sequential recommenders}. In \bibinfo{booktitle}{\emph{Proceedings of the 26th ACM SIGKDD International Conference on Knowledge Discovery \& Data Mining}}. \bibinfo{pages}{483--491}.
\newblock


\bibitem[\protect\citeauthoryear{Ma, Sun, Georgoulis, Van~Gool, Schiele, and Fritz}{Ma et~al\mbox{.}}{2018}]%
        {ma2018disentangled}
\bibfield{author}{\bibinfo{person}{Liqian Ma}, \bibinfo{person}{Qianru Sun}, \bibinfo{person}{Stamatios Georgoulis}, \bibinfo{person}{Luc Van~Gool}, \bibinfo{person}{Bernt Schiele}, {and} \bibinfo{person}{Mario Fritz}.} \bibinfo{year}{2018}\natexlab{}.
\newblock \showarticletitle{Disentangled person image generation}. In \bibinfo{booktitle}{\emph{Proceedings of the IEEE Conference on Computer Vision and Pattern Recognition}}. \bibinfo{pages}{99--108}.
\newblock


\bibitem[\protect\citeauthoryear{Ma, Huang, Lu, and Hu}{Ma et~al\mbox{.}}{2023}]%
        {ma2023kr}
\bibfield{author}{\bibinfo{person}{Ting Ma}, \bibinfo{person}{Longtao Huang}, \bibinfo{person}{Qianqian Lu}, {and} \bibinfo{person}{Songlin Hu}.} \bibinfo{year}{2023}\natexlab{}.
\newblock \showarticletitle{Kr-gcn: Knowledge-aware reasoning with graph convolution network for explainable recommendation}.
\newblock \bibinfo{journal}{\emph{ACM Transactions on Information Systems (TOIS)}} \bibinfo{volume}{41}, \bibinfo{number}{1} (\bibinfo{year}{2023}), \bibinfo{pages}{1--27}.
\newblock


\bibitem[\protect\citeauthoryear{Mikolov, Chen, Corrado, and Dean}{Mikolov et~al\mbox{.}}{2013a}]%
        {mikolov2013efficient}
\bibfield{author}{\bibinfo{person}{Tom{\'{a}}s Mikolov}, \bibinfo{person}{Kai Chen}, \bibinfo{person}{Greg Corrado}, {and} \bibinfo{person}{Jeffrey Dean}.} \bibinfo{year}{2013}\natexlab{a}.
\newblock \showarticletitle{Efficient Estimation of Word Representations in Vector Space}. In \bibinfo{booktitle}{\emph{1st International Conference on Learning Representations}}, \bibfield{editor}{\bibinfo{person}{Yoshua Bengio} {and} \bibinfo{person}{Yann LeCun}} (Eds.).
\newblock


\bibitem[\protect\citeauthoryear{Mikolov, Sutskever, Chen, Corrado, and Dean}{Mikolov et~al\mbox{.}}{2013b}]%
        {mikolov2013distributed}
\bibfield{author}{\bibinfo{person}{Tomas Mikolov}, \bibinfo{person}{Ilya Sutskever}, \bibinfo{person}{Kai Chen}, \bibinfo{person}{Greg~S Corrado}, {and} \bibinfo{person}{Jeff Dean}.} \bibinfo{year}{2013}\natexlab{b}.
\newblock \showarticletitle{Distributed representations of words and phrases and their compositionality}.
\newblock \bibinfo{journal}{\emph{Advances in neural information processing systems}}  \bibinfo{volume}{26} (\bibinfo{year}{2013}).
\newblock


\bibitem[\protect\citeauthoryear{Mitrovic, McWilliams, Walker, Buesing, and Blundell}{Mitrovic et~al\mbox{.}}{2021}]%
        {mitrovic2020representation}
\bibfield{author}{\bibinfo{person}{Jovana Mitrovic}, \bibinfo{person}{Brian McWilliams}, \bibinfo{person}{Jacob~C. Walker}, \bibinfo{person}{Lars~Holger Buesing}, {and} \bibinfo{person}{Charles Blundell}.} \bibinfo{year}{2021}\natexlab{}.
\newblock \showarticletitle{Representation Learning via Invariant Causal Mechanisms}. In \bibinfo{booktitle}{\emph{9th International Conference on Learning Representations}}. \bibinfo{publisher}{OpenReview.net}.
\newblock


\bibitem[\protect\citeauthoryear{Nascimento, Pimentel, Souza, Costa, Gon{\c{c}}alves, and Louzada}{Nascimento et~al\mbox{.}}{2021}]%
        {nascimento2021dynamic}
\bibfield{author}{\bibinfo{person}{Diego~C Nascimento}, \bibinfo{person}{Bruno~A Pimentel}, \bibinfo{person}{Renata~MCR Souza}, \bibinfo{person}{Lilia Costa}, \bibinfo{person}{Sandro Gon{\c{c}}alves}, {and} \bibinfo{person}{Francisco Louzada}.} \bibinfo{year}{2021}\natexlab{}.
\newblock \showarticletitle{Dynamic graph in a symbolic data framework: An account of the causal relation using COVID-19 reports and some reflections on the financial world}.
\newblock \bibinfo{journal}{\emph{Chaos, Solitons \& Fractals}}  \bibinfo{volume}{153} (\bibinfo{year}{2021}), \bibinfo{pages}{111440}.
\newblock


\bibitem[\protect\citeauthoryear{Paranjape, Benson, and Leskovec}{Paranjape et~al\mbox{.}}{2017}]%
        {paranjape2017motifs}
\bibfield{author}{\bibinfo{person}{Ashwin Paranjape}, \bibinfo{person}{Austin~R Benson}, {and} \bibinfo{person}{Jure Leskovec}.} \bibinfo{year}{2017}\natexlab{}.
\newblock \showarticletitle{Motifs in temporal networks}. In \bibinfo{booktitle}{\emph{Proceedings of the tenth ACM international conference on web search and data mining}}. \bibinfo{pages}{601--610}.
\newblock


\bibitem[\protect\citeauthoryear{Pareja, Domeniconi, Chen, Ma, Suzumura, Kanezashi, Kaler, Schardl, and Leiserson}{Pareja et~al\mbox{.}}{2020}]%
        {pareja2020evolvegcn}
\bibfield{author}{\bibinfo{person}{Aldo Pareja}, \bibinfo{person}{Giacomo Domeniconi}, \bibinfo{person}{Jie Chen}, \bibinfo{person}{Tengfei Ma}, \bibinfo{person}{Toyotaro Suzumura}, \bibinfo{person}{Hiroki Kanezashi}, \bibinfo{person}{Tim Kaler}, \bibinfo{person}{Tao Schardl}, {and} \bibinfo{person}{Charles Leiserson}.} \bibinfo{year}{2020}\natexlab{}.
\newblock \showarticletitle{Evolvegcn: Evolving graph convolutional networks for dynamic graphs}. In \bibinfo{booktitle}{\emph{Proceedings of the AAAI Conference on Artificial Intelligence}}, Vol.~\bibinfo{volume}{34}. \bibinfo{pages}{5363--5370}.
\newblock


\bibitem[\protect\citeauthoryear{Paszke, Gross, Massa, Lerer, Bradbury, Chanan, Killeen, Lin, Gimelshein, Antiga, et~al\mbox{.}}{Paszke et~al\mbox{.}}{2019}]%
        {paszke2019pytorch}
\bibfield{author}{\bibinfo{person}{Adam Paszke}, \bibinfo{person}{Sam Gross}, \bibinfo{person}{Francisco Massa}, \bibinfo{person}{Adam Lerer}, \bibinfo{person}{James Bradbury}, \bibinfo{person}{Gregory Chanan}, \bibinfo{person}{Trevor Killeen}, \bibinfo{person}{Zeming Lin}, \bibinfo{person}{Natalia Gimelshein}, \bibinfo{person}{Luca Antiga}, {et~al\mbox{.}}} \bibinfo{year}{2019}\natexlab{}.
\newblock \showarticletitle{Pytorch: An imperative style, high-performance deep learning library}.
\newblock \bibinfo{journal}{\emph{Advances in neural information processing systems}} (\bibinfo{year}{2019}).
\newblock


\bibitem[\protect\citeauthoryear{Pearl et~al\mbox{.}}{Pearl et~al\mbox{.}}{2000}]%
        {pearl2000models}
\bibfield{author}{\bibinfo{person}{Judea Pearl} {et~al\mbox{.}}} \bibinfo{year}{2000}\natexlab{}.
\newblock \showarticletitle{Models, reasoning and inference}.
\newblock \bibinfo{journal}{\emph{Cambridge, UK: CambridgeUniversityPress}}  \bibinfo{volume}{19} (\bibinfo{year}{2000}), \bibinfo{pages}{2}.
\newblock


\bibitem[\protect\citeauthoryear{Peng, Du, Liu, Liu, Ji, Wang, Zhang, and He}{Peng et~al\mbox{.}}{2021}]%
        {peng2021dynamic}
\bibfield{author}{\bibinfo{person}{Hao Peng}, \bibinfo{person}{Bowen Du}, \bibinfo{person}{Mingsheng Liu}, \bibinfo{person}{Mingzhe Liu}, \bibinfo{person}{Shumei Ji}, \bibinfo{person}{Senzhang Wang}, \bibinfo{person}{Xu Zhang}, {and} \bibinfo{person}{Lifang He}.} \bibinfo{year}{2021}\natexlab{}.
\newblock \showarticletitle{Dynamic graph convolutional network for long-term traffic flow prediction with reinforcement learning}.
\newblock \bibinfo{journal}{\emph{Information Sciences}}  \bibinfo{volume}{578} (\bibinfo{year}{2021}), \bibinfo{pages}{401--416}.
\newblock


\bibitem[\protect\citeauthoryear{Peng, Wang, Du, Bhuiyan, Ma, Liu, Wang, Yang, Du, Wang, et~al\mbox{.}}{Peng et~al\mbox{.}}{2020}]%
        {peng2020spatial}
\bibfield{author}{\bibinfo{person}{Hao Peng}, \bibinfo{person}{Hongfei Wang}, \bibinfo{person}{Bowen Du}, \bibinfo{person}{Md~Zakirul~Alam Bhuiyan}, \bibinfo{person}{Hongyuan Ma}, \bibinfo{person}{Jianwei Liu}, \bibinfo{person}{Lihong Wang}, \bibinfo{person}{Zeyu Yang}, \bibinfo{person}{Linfeng Du}, \bibinfo{person}{Senzhang Wang}, {et~al\mbox{.}}} \bibinfo{year}{2020}\natexlab{}.
\newblock \showarticletitle{Spatial temporal incidence dynamic graph neural networks for traffic flow forecasting}.
\newblock \bibinfo{journal}{\emph{Information Sciences}}  \bibinfo{volume}{521} (\bibinfo{year}{2020}), \bibinfo{pages}{277--290}.
\newblock


\bibitem[\protect\citeauthoryear{Qin, Wang, Zhang, Xie, and Zhu}{Qin et~al\mbox{.}}{2022}]%
        {qin2022graph}
\bibfield{author}{\bibinfo{person}{Yijian Qin}, \bibinfo{person}{Xin Wang}, \bibinfo{person}{Ziwei Zhang}, \bibinfo{person}{Pengtao Xie}, {and} \bibinfo{person}{Wenwu Zhu}.} \bibinfo{year}{2022}\natexlab{}.
\newblock \showarticletitle{Graph Neural Architecture Search Under Distribution Shifts}. In \bibinfo{booktitle}{\emph{International Conference on Machine Learning}}. \bibinfo{pages}{18083--18095}.
\newblock


\bibitem[\protect\citeauthoryear{Qiu, Hu, Wu, Liu, Du, and Jia}{Qiu et~al\mbox{.}}{2020}]%
        {qiu2020temporal}
\bibfield{author}{\bibinfo{person}{Zhenyu Qiu}, \bibinfo{person}{Wenbin Hu}, \bibinfo{person}{Jia Wu}, \bibinfo{person}{Weiwei Liu}, \bibinfo{person}{Bo Du}, {and} \bibinfo{person}{Xiaohua Jia}.} \bibinfo{year}{2020}\natexlab{}.
\newblock \showarticletitle{Temporal network embedding with high-order nonlinear information}. In \bibinfo{booktitle}{\emph{Proceedings of the AAAI Conference on Artificial Intelligence}}, Vol.~\bibinfo{volume}{34}. \bibinfo{pages}{5436--5443}.
\newblock


\bibitem[\protect\citeauthoryear{Rosenfeld, Ravikumar, and Risteski}{Rosenfeld et~al\mbox{.}}{2021}]%
        {rosenfeld2020risks}
\bibfield{author}{\bibinfo{person}{Elan Rosenfeld}, \bibinfo{person}{Pradeep~Kumar Ravikumar}, {and} \bibinfo{person}{Andrej Risteski}.} \bibinfo{year}{2021}\natexlab{}.
\newblock \showarticletitle{The Risks of Invariant Risk Minimization}. In \bibinfo{booktitle}{\emph{9th International Conference on Learning Representations}}. \bibinfo{publisher}{OpenReview.net}.
\newblock


\bibitem[\protect\citeauthoryear{Rossi, Chamberlain, Frasca, Eynard, Monti, and Bronstein}{Rossi et~al\mbox{.}}{2020}]%
        {rossi2020temporal}
\bibfield{author}{\bibinfo{person}{Emanuele Rossi}, \bibinfo{person}{Ben Chamberlain}, \bibinfo{person}{Fabrizio Frasca}, \bibinfo{person}{Davide Eynard}, \bibinfo{person}{Federico Monti}, {and} \bibinfo{person}{Michael Bronstein}.} \bibinfo{year}{2020}\natexlab{}.
\newblock \showarticletitle{Temporal graph networks for deep learning on dynamic graphs}.
\newblock \bibinfo{journal}{\emph{arXiv preprint arXiv:2006.10637}} (\bibinfo{year}{2020}).
\newblock


\bibitem[\protect\citeauthoryear{Sagawa, Koh, Hashimoto, and Liang}{Sagawa et~al\mbox{.}}{[n.\,d.]}]%
        {sagawa2019distributionally}
\bibfield{author}{\bibinfo{person}{Shiori Sagawa}, \bibinfo{person}{Pang~Wei Koh}, \bibinfo{person}{Tatsunori~B Hashimoto}, {and} \bibinfo{person}{Percy Liang}.} \bibinfo{year}{[n.\,d.]}\natexlab{}.
\newblock \showarticletitle{Distributionally Robust Neural Networks}. In \bibinfo{booktitle}{\emph{International Conference on Learning Representations}}.
\newblock


\bibitem[\protect\citeauthoryear{Sankar, Wu, Gou, Zhang, and Yang}{Sankar et~al\mbox{.}}{2020}]%
        {sankar2020dysat}
\bibfield{author}{\bibinfo{person}{Aravind Sankar}, \bibinfo{person}{Yanhong Wu}, \bibinfo{person}{Liang Gou}, \bibinfo{person}{Wei Zhang}, {and} \bibinfo{person}{Hao Yang}.} \bibinfo{year}{2020}\natexlab{}.
\newblock \showarticletitle{Dysat: Deep neural representation learning on dynamic graphs via self-attention networks}. In \bibinfo{booktitle}{\emph{Proceedings of the 13th International Conference on Web Search and Data Mining}}. \bibinfo{pages}{519--527}.
\newblock


\bibitem[\protect\citeauthoryear{Seo, Defferrard, Vandergheynst, and Bresson}{Seo et~al\mbox{.}}{2018}]%
        {seo2018structured}
\bibfield{author}{\bibinfo{person}{Youngjoo Seo}, \bibinfo{person}{Micha{\"e}l Defferrard}, \bibinfo{person}{Pierre Vandergheynst}, {and} \bibinfo{person}{Xavier Bresson}.} \bibinfo{year}{2018}\natexlab{}.
\newblock \showarticletitle{Structured sequence modeling with graph convolutional recurrent networks}. In \bibinfo{booktitle}{\emph{International Conference on Neural Information Processing}}. Springer, \bibinfo{pages}{362--373}.
\newblock


\bibitem[\protect\citeauthoryear{Shen, Liu, He, Zhang, Xu, Yu, and Cui}{Shen et~al\mbox{.}}{2021}]%
        {shen2021towards}
\bibfield{author}{\bibinfo{person}{Zheyan Shen}, \bibinfo{person}{Jiashuo Liu}, \bibinfo{person}{Yue He}, \bibinfo{person}{Xingxuan Zhang}, \bibinfo{person}{Renzhe Xu}, \bibinfo{person}{Han Yu}, {and} \bibinfo{person}{Peng Cui}.} \bibinfo{year}{2021}\natexlab{}.
\newblock \showarticletitle{Towards out-of-distribution generalization: A survey}.
\newblock \bibinfo{journal}{\emph{arXiv preprint arXiv:2108.13624}} (\bibinfo{year}{2021}).
\newblock


\bibitem[\protect\citeauthoryear{Simmel}{Simmel}{1950}]%
        {simmel1950sociology}
\bibfield{author}{\bibinfo{person}{Georg Simmel}.} \bibinfo{year}{1950}\natexlab{}.
\newblock \bibinfo{booktitle}{\emph{The sociology of georg simmel}}. Vol.~\bibinfo{volume}{92892}.
\newblock \bibinfo{publisher}{Simon and Schuster}.
\newblock


\bibitem[\protect\citeauthoryear{Sinha, Shen, Song, Ma, Eide, Hsu, and Wang}{Sinha et~al\mbox{.}}{2015}]%
        {sinha2015overview}
\bibfield{author}{\bibinfo{person}{Arnab Sinha}, \bibinfo{person}{Zhihong Shen}, \bibinfo{person}{Yang Song}, \bibinfo{person}{Hao Ma}, \bibinfo{person}{Darrin Eide}, \bibinfo{person}{Bo-june~Paul Hsu}, {and} \bibinfo{person}{Kuansan Wang}.} \bibinfo{year}{2015}\natexlab{}.
\newblock \showarticletitle{An overview of microsoft academic service (mas) and applications}. In \bibinfo{booktitle}{\emph{Proceedings of the 24th international conference on world wide web}}. ACM, \bibinfo{pages}{243--246}.
\newblock


\bibitem[\protect\citeauthoryear{Skarding, Gabrys, and Musial}{Skarding et~al\mbox{.}}{2021}]%
        {skarding2021foundations}
\bibfield{author}{\bibinfo{person}{Joakim Skarding}, \bibinfo{person}{Bogdan Gabrys}, {and} \bibinfo{person}{Katarzyna Musial}.} \bibinfo{year}{2021}\natexlab{}.
\newblock \showarticletitle{Foundations and Modeling of Dynamic Networks Using Dynamic Graph Neural Networks: A Survey}.
\newblock \bibinfo{journal}{\emph{IEEE Access}} (\bibinfo{year}{2021}), \bibinfo{pages}{79143--79168}.
\newblock


\bibitem[\protect\citeauthoryear{Sun, Zhang, Zhang, Wang, Peng, Su, and Yu}{Sun et~al\mbox{.}}{2021}]%
        {sun2021hyperbolic}
\bibfield{author}{\bibinfo{person}{Li Sun}, \bibinfo{person}{Zhongbao Zhang}, \bibinfo{person}{Jiawei Zhang}, \bibinfo{person}{Feiyang Wang}, \bibinfo{person}{Hao Peng}, \bibinfo{person}{Sen Su}, {and} \bibinfo{person}{Philip~S Yu}.} \bibinfo{year}{2021}\natexlab{}.
\newblock \showarticletitle{Hyperbolic variational graph neural network for modeling dynamic graphs}. In \bibinfo{booktitle}{\emph{Proceedings of the AAAI Conference on Artificial Intelligence}}, Vol.~\bibinfo{volume}{35}. \bibinfo{pages}{4375--4383}.
\newblock


\bibitem[\protect\citeauthoryear{Taheri, Gimpel, and Berger-Wolf}{Taheri et~al\mbox{.}}{2019}]%
        {taheri2019learning}
\bibfield{author}{\bibinfo{person}{Aynaz Taheri}, \bibinfo{person}{Kevin Gimpel}, {and} \bibinfo{person}{Tanya Berger-Wolf}.} \bibinfo{year}{2019}\natexlab{}.
\newblock \showarticletitle{Learning to represent the evolution of dynamic graphs with recurrent models}. In \bibinfo{booktitle}{\emph{Proceedings of the ACM Web Conference 2019}}. \bibinfo{pages}{301--307}.
\newblock


\bibitem[\protect\citeauthoryear{Tang, Wu, Xu, and Li}{Tang et~al\mbox{.}}{2023}]%
        {tang2023dynamic}
\bibfield{author}{\bibinfo{person}{Haoran Tang}, \bibinfo{person}{Shiqing Wu}, \bibinfo{person}{Guandong Xu}, {and} \bibinfo{person}{Qing Li}.} \bibinfo{year}{2023}\natexlab{}.
\newblock \showarticletitle{Dynamic Graph Evolution Learning for Recommendation}. In \bibinfo{booktitle}{\emph{Proceedings of the 46th International ACM SIGIR Conference on Research and Development in Information Retrieval}}. \bibinfo{pages}{1589--1598}.
\newblock


\bibitem[\protect\citeauthoryear{Tang, Wu, Sun, and Su}{Tang et~al\mbox{.}}{2012}]%
        {Tang:12KDDCross}
\bibfield{author}{\bibinfo{person}{Jie Tang}, \bibinfo{person}{Sen Wu}, \bibinfo{person}{Jimeng Sun}, {and} \bibinfo{person}{Hang Su}.} \bibinfo{year}{2012}\natexlab{}.
\newblock \showarticletitle{Cross-domain Collaboration Recommendation}. In \bibinfo{booktitle}{\emph{KDD'2012}}.
\newblock


\bibitem[\protect\citeauthoryear{Tang, Zhang, Yao, Li, Zhang, and Su}{Tang et~al\mbox{.}}{2008}]%
        {Tang:08KDD}
\bibfield{author}{\bibinfo{person}{Jie Tang}, \bibinfo{person}{Jing Zhang}, \bibinfo{person}{Limin Yao}, \bibinfo{person}{Juanzi Li}, \bibinfo{person}{Li Zhang}, {and} \bibinfo{person}{Zhong Su}.} \bibinfo{year}{2008}\natexlab{}.
\newblock \showarticletitle{ArnetMiner: Extraction and Mining of Academic Social Networks}. In \bibinfo{booktitle}{\emph{KDD'08}}. \bibinfo{pages}{990--998}.
\newblock


\bibitem[\protect\citeauthoryear{Tian, Kang, and Pearl}{Tian et~al\mbox{.}}{2006}]%
        {tian2006characterization}
\bibfield{author}{\bibinfo{person}{Jin Tian}, \bibinfo{person}{Changsung Kang}, {and} \bibinfo{person}{Judea Pearl}.} \bibinfo{year}{2006}\natexlab{}.
\newblock \bibinfo{booktitle}{\emph{A characterization of interventional distributions in semi-Markovian causal models}}.
\newblock \bibinfo{publisher}{eScholarship, University of California}.
\newblock


\bibitem[\protect\citeauthoryear{Tran, Yin, and Liu}{Tran et~al\mbox{.}}{2017}]%
        {tran2017disentangled}
\bibfield{author}{\bibinfo{person}{Luan Tran}, \bibinfo{person}{Xi Yin}, {and} \bibinfo{person}{Xiaoming Liu}.} \bibinfo{year}{2017}\natexlab{}.
\newblock \showarticletitle{Disentangled representation learning gan for pose-invariant face recognition}. In \bibinfo{booktitle}{\emph{Proceedings of the IEEE conference on computer vision and pattern recognition}}. \bibinfo{pages}{1415--1424}.
\newblock


\bibitem[\protect\citeauthoryear{Trivedi, Farajtabar, Biswal, and Zha}{Trivedi et~al\mbox{.}}{2019}]%
        {trivedi2019dyrep}
\bibfield{author}{\bibinfo{person}{Rakshit Trivedi}, \bibinfo{person}{Mehrdad Farajtabar}, \bibinfo{person}{Prasenjeet Biswal}, {and} \bibinfo{person}{Hongyuan Zha}.} \bibinfo{year}{2019}\natexlab{}.
\newblock \showarticletitle{Dyrep: Learning representations over dynamic graphs}. In \bibinfo{booktitle}{\emph{International conference on learning representations}}.
\newblock


\bibitem[\protect\citeauthoryear{Vaswani, Shazeer, Parmar, Uszkoreit, Jones, Gomez, Kaiser, and Polosukhin}{Vaswani et~al\mbox{.}}{2017}]%
        {vaswani2017attention}
\bibfield{author}{\bibinfo{person}{Ashish Vaswani}, \bibinfo{person}{Noam Shazeer}, \bibinfo{person}{Niki Parmar}, \bibinfo{person}{Jakob Uszkoreit}, \bibinfo{person}{Llion Jones}, \bibinfo{person}{Aidan~N Gomez}, \bibinfo{person}{{\L}ukasz Kaiser}, {and} \bibinfo{person}{Illia Polosukhin}.} \bibinfo{year}{2017}\natexlab{}.
\newblock \showarticletitle{Attention is all you need}.
\newblock \bibinfo{journal}{\emph{Advances in neural information processing systems}} (\bibinfo{year}{2017}).
\newblock


\bibitem[\protect\citeauthoryear{Veli{\v{c}}kovi{\'c}, Cucurull, Casanova, Romero, Li{\`o}, and Bengio}{Veli{\v{c}}kovi{\'c} et~al\mbox{.}}{[n.\,d.]}]%
        {velivckovicgraph}
\bibfield{author}{\bibinfo{person}{Petar Veli{\v{c}}kovi{\'c}}, \bibinfo{person}{Guillem Cucurull}, \bibinfo{person}{Arantxa Casanova}, \bibinfo{person}{Adriana Romero}, \bibinfo{person}{Pietro Li{\`o}}, {and} \bibinfo{person}{Yoshua Bengio}.} \bibinfo{year}{[n.\,d.]}\natexlab{}.
\newblock \showarticletitle{Graph Attention Networks}. In \bibinfo{booktitle}{\emph{International Conference on Learning Representations}}.
\newblock


\bibitem[\protect\citeauthoryear{Venkateswaran, Muthusamy, Isahagian, and Venkatasubramanian}{Venkateswaran et~al\mbox{.}}{2021}]%
        {venkateswaran2021environment}
\bibfield{author}{\bibinfo{person}{Praveen Venkateswaran}, \bibinfo{person}{Vinod Muthusamy}, \bibinfo{person}{Vatche Isahagian}, {and} \bibinfo{person}{Nalini Venkatasubramanian}.} \bibinfo{year}{2021}\natexlab{}.
\newblock \showarticletitle{Environment agnostic invariant risk minimization for classification of sequential datasets}. In \bibinfo{booktitle}{\emph{Proceedings of the 27th ACM SIGKDD Conference on Knowledge Discovery \& Data Mining}}. \bibinfo{pages}{1615--1624}.
\newblock


\bibitem[\protect\citeauthoryear{Wang and Leskovec}{Wang and Leskovec}{2021}]%
        {wang2021combining}
\bibfield{author}{\bibinfo{person}{Hongwei Wang} {and} \bibinfo{person}{Jure Leskovec}.} \bibinfo{year}{2021}\natexlab{}.
\newblock \showarticletitle{Combining graph convolutional neural networks and label propagation}.
\newblock \bibinfo{journal}{\emph{ACM Transactions on Information Systems (TOIS)}} \bibinfo{volume}{40}, \bibinfo{number}{4} (\bibinfo{year}{2021}), \bibinfo{pages}{1--27}.
\newblock


\bibitem[\protect\citeauthoryear{Wang, Li, Pan, and Xie}{Wang et~al\mbox{.}}{2023}]%
        {wang2023tutorial}
\bibfield{author}{\bibinfo{person}{Jindong Wang}, \bibinfo{person}{Haoliang Li}, \bibinfo{person}{Sinno Pan}, {and} \bibinfo{person}{Xing Xie}.} \bibinfo{year}{2023}\natexlab{}.
\newblock \showarticletitle{A Tutorial on Domain Generalization}. In \bibinfo{booktitle}{\emph{Proceedings of the Sixteenth ACM International Conference on Web Search and Data Mining}}. \bibinfo{pages}{1236--1239}.
\newblock


\bibitem[\protect\citeauthoryear{Wang, Shen, Huang, Wu, Dong, and Kanakia}{Wang et~al\mbox{.}}{2020}]%
        {wang2020microsoft}
\bibfield{author}{\bibinfo{person}{Kuansan Wang}, \bibinfo{person}{Zhihong Shen}, \bibinfo{person}{Chiyuan Huang}, \bibinfo{person}{Chieh-Han Wu}, \bibinfo{person}{Yuxiao Dong}, {and} \bibinfo{person}{Anshul Kanakia}.} \bibinfo{year}{2020}\natexlab{}.
\newblock \showarticletitle{Microsoft academic graph: When experts are not enough}.
\newblock \bibinfo{journal}{\emph{Quantitative Science Studies}} \bibinfo{volume}{1}, \bibinfo{number}{1} (\bibinfo{year}{2020}), \bibinfo{pages}{396--413}.
\newblock


\bibitem[\protect\citeauthoryear{Wang, Lin, Feng, He, Lin, and Chua}{Wang et~al\mbox{.}}{2022b}]%
        {wang2022causal}
\bibfield{author}{\bibinfo{person}{Wenjie Wang}, \bibinfo{person}{Xinyu Lin}, \bibinfo{person}{Fuli Feng}, \bibinfo{person}{Xiangnan He}, \bibinfo{person}{Min Lin}, {and} \bibinfo{person}{Tat-Seng Chua}.} \bibinfo{year}{2022}\natexlab{b}.
\newblock \showarticletitle{Causal Representation Learning for Out-of-Distribution Recommendation}. In \bibinfo{booktitle}{\emph{Proceedings of the ACM Web Conference 2022}}. \bibinfo{pages}{3562--3571}.
\newblock


\bibitem[\protect\citeauthoryear{Wang, Chen, Zhou, Ma, and Zhu}{Wang et~al\mbox{.}}{2022a}]%
        {wang2022disentangled}
\bibfield{author}{\bibinfo{person}{Xin Wang}, \bibinfo{person}{Hong Chen}, \bibinfo{person}{Yuwei Zhou}, \bibinfo{person}{Jianxin Ma}, {and} \bibinfo{person}{Wenwu Zhu}.} \bibinfo{year}{2022}\natexlab{a}.
\newblock \showarticletitle{Disentangled Representation Learning for Recommendation}.
\newblock \bibinfo{journal}{\emph{IEEE Transactions on Pattern Analysis and Machine Intelligence}} (\bibinfo{year}{2022}).
\newblock


\bibitem[\protect\citeauthoryear{Wang, Chen, and Zhu}{Wang et~al\mbox{.}}{2021b}]%
        {wang2021multimodal}
\bibfield{author}{\bibinfo{person}{Xin Wang}, \bibinfo{person}{Hong Chen}, {and} \bibinfo{person}{Wenwu Zhu}.} \bibinfo{year}{2021}\natexlab{b}.
\newblock \showarticletitle{Multimodal disentangled representation for recommendation}. In \bibinfo{booktitle}{\emph{2021 IEEE International Conference on Multimedia and Expo (ICME)}}. \bibinfo{pages}{1--6}.
\newblock


\bibitem[\protect\citeauthoryear{Wang, Cui, Wang, Pei, Zhu, and Yang}{Wang et~al\mbox{.}}{2017}]%
        {wang2017community}
\bibfield{author}{\bibinfo{person}{Xiao Wang}, \bibinfo{person}{Peng Cui}, \bibinfo{person}{Jing Wang}, \bibinfo{person}{Jian Pei}, \bibinfo{person}{Wenwu Zhu}, {and} \bibinfo{person}{Shiqiang Yang}.} \bibinfo{year}{2017}\natexlab{}.
\newblock \showarticletitle{Community preserving network embedding}. In \bibinfo{booktitle}{\emph{Proceedings of the AAAI conference on artificial intelligence}}, Vol.~\bibinfo{volume}{31}.
\newblock


\bibitem[\protect\citeauthoryear{Wang, Ji, Shi, Wang, Ye, Cui, and Yu}{Wang et~al\mbox{.}}{2019}]%
        {wang2019heterogeneous}
\bibfield{author}{\bibinfo{person}{Xiao Wang}, \bibinfo{person}{Houye Ji}, \bibinfo{person}{Chuan Shi}, \bibinfo{person}{Bai Wang}, \bibinfo{person}{Yanfang Ye}, \bibinfo{person}{Peng Cui}, {and} \bibinfo{person}{Philip~S Yu}.} \bibinfo{year}{2019}\natexlab{}.
\newblock \showarticletitle{Heterogeneous graph attention network}. In \bibinfo{booktitle}{\emph{The world wide web conference}}. \bibinfo{pages}{2022--2032}.
\newblock


\bibitem[\protect\citeauthoryear{Wang, Chang, Liu, Leskovec, and Li}{Wang et~al\mbox{.}}{2021a}]%
        {wang2021inductive}
\bibfield{author}{\bibinfo{person}{Yanbang Wang}, \bibinfo{person}{Yen{-}Yu Chang}, \bibinfo{person}{Yunyu Liu}, \bibinfo{person}{Jure Leskovec}, {and} \bibinfo{person}{Pan Li}.} \bibinfo{year}{2021}\natexlab{a}.
\newblock \showarticletitle{Inductive Representation Learning in Temporal Networks via Causal Anonymous Walks}. In \bibinfo{booktitle}{\emph{9th International Conference on Learning Representations}}. \bibinfo{publisher}{OpenReview.net}.
\newblock


\bibitem[\protect\citeauthoryear{Wang, Li, Bai, and Leskovec}{Wang et~al\mbox{.}}{2021c}]%
        {wang2021tedic}
\bibfield{author}{\bibinfo{person}{Yanbang Wang}, \bibinfo{person}{Pan Li}, \bibinfo{person}{Chongyang Bai}, {and} \bibinfo{person}{Jure Leskovec}.} \bibinfo{year}{2021}\natexlab{c}.
\newblock \showarticletitle{TEDIC: Neural modeling of behavioral patterns in dynamic social interaction networks}. In \bibinfo{booktitle}{\emph{Proceedings of the Web Conference 2021}}. \bibinfo{pages}{693--705}.
\newblock


\bibitem[\protect\citeauthoryear{Wang, Qin, Sun, Zhang, Hou, Hu, Cheng, Lei, and Zhang}{Wang et~al\mbox{.}}{2022c}]%
        {wang2022disenctr}
\bibfield{author}{\bibinfo{person}{Yifan Wang}, \bibinfo{person}{Yifang Qin}, \bibinfo{person}{Fang Sun}, \bibinfo{person}{Bo Zhang}, \bibinfo{person}{Xuyang Hou}, \bibinfo{person}{Ke Hu}, \bibinfo{person}{Jia Cheng}, \bibinfo{person}{Jun Lei}, {and} \bibinfo{person}{Ming Zhang}.} \bibinfo{year}{2022}\natexlab{c}.
\newblock \showarticletitle{DisenCTR: Dynamic graph-based disentangled representation for click-through rate prediction}. In \bibinfo{booktitle}{\emph{Proceedings of the 45th International ACM SIGIR Conference on Research and Development in Information Retrieval}}. \bibinfo{pages}{2314--2318}.
\newblock


\bibitem[\protect\citeauthoryear{Wang, Zhao, Shah, and Derr}{Wang et~al\mbox{.}}{2022d}]%
        {wang2022imbalanced}
\bibfield{author}{\bibinfo{person}{Yu Wang}, \bibinfo{person}{Yuying Zhao}, \bibinfo{person}{Neil Shah}, {and} \bibinfo{person}{Tyler Derr}.} \bibinfo{year}{2022}\natexlab{d}.
\newblock \showarticletitle{Imbalanced graph classification via graph-of-graph neural networks}. In \bibinfo{booktitle}{\emph{Proceedings of the 31st ACM International Conference on Information \& Knowledge Management}}. \bibinfo{pages}{2067--2076}.
\newblock


\bibitem[\protect\citeauthoryear{Wang, Zhou, Miao, Liu, and Wang}{Wang et~al\mbox{.}}{2022e}]%
        {wang2022adagcl}
\bibfield{author}{\bibinfo{person}{Yili Wang}, \bibinfo{person}{Kaixiong Zhou}, \bibinfo{person}{Rui Miao}, \bibinfo{person}{Ninghao Liu}, {and} \bibinfo{person}{Xin Wang}.} \bibinfo{year}{2022}\natexlab{e}.
\newblock \showarticletitle{AdaGCL: Adaptive Subgraph Contrastive Learning to Generalize Large-scale Graph Training}. In \bibinfo{booktitle}{\emph{Proceedings of the 31st ACM International Conference on Information \& Knowledge Management}}. \bibinfo{pages}{2046--2055}.
\newblock


\bibitem[\protect\citeauthoryear{Wu, Cao, Cheung, and Hamilton}{Wu et~al\mbox{.}}{2020a}]%
        {wu2020temp}
\bibfield{author}{\bibinfo{person}{Jiapeng Wu}, \bibinfo{person}{Meng Cao}, \bibinfo{person}{Jackie Chi~Kit Cheung}, {and} \bibinfo{person}{William~L. Hamilton}.} \bibinfo{year}{2020}\natexlab{a}.
\newblock \showarticletitle{TeMP: Temporal Message Passing for Temporal Knowledge Graph Completion}. In \bibinfo{booktitle}{\emph{Proceedings of the 2020 Conference on Empirical Methods in Natural Language Processing, {EMNLP} 2020, Online, November 16-20, 2020}}, \bibfield{editor}{\bibinfo{person}{Bonnie Webber}, \bibinfo{person}{Trevor Cohn}, \bibinfo{person}{Yulan He}, {and} \bibinfo{person}{Yang Liu}} (Eds.). \bibinfo{publisher}{Association for Computational Linguistics}, \bibinfo{pages}{5730--5746}.
\newblock


\bibitem[\protect\citeauthoryear{Wu, Zhang, Mao, Guo, Soflaei, and Huai}{Wu et~al\mbox{.}}{2020c}]%
        {wu2020dynamic}
\bibfield{author}{\bibinfo{person}{Junshuang Wu}, \bibinfo{person}{Richong Zhang}, \bibinfo{person}{Yongyi Mao}, \bibinfo{person}{Hongyu Guo}, \bibinfo{person}{Masoumeh Soflaei}, {and} \bibinfo{person}{Jinpeng Huai}.} \bibinfo{year}{2020}\natexlab{c}.
\newblock \showarticletitle{Dynamic graph convolutional networks for entity linking}. In \bibinfo{booktitle}{\emph{Proceedings of The ACM Web Conference 2020}}. \bibinfo{pages}{1149--1159}.
\newblock


\bibitem[\protect\citeauthoryear{Wu, Zhang, Yan, and Wipf}{Wu et~al\mbox{.}}{2022b}]%
        {wu2022handling}
\bibfield{author}{\bibinfo{person}{Qitian Wu}, \bibinfo{person}{Hengrui Zhang}, \bibinfo{person}{Junchi Yan}, {and} \bibinfo{person}{David Wipf}.} \bibinfo{year}{2022}\natexlab{b}.
\newblock \showarticletitle{Handling Distribution Shifts on Graphs: An Invariance Perspective}.
\newblock \bibinfo{journal}{\emph{International Conference on Learning Representations}} (\bibinfo{year}{2022}).
\newblock


\bibitem[\protect\citeauthoryear{Wu, Wang, Zhang, He, and Chua}{Wu et~al\mbox{.}}{2022a}]%
        {wu2022discovering}
\bibfield{author}{\bibinfo{person}{Yingxin Wu}, \bibinfo{person}{Xiang Wang}, \bibinfo{person}{An Zhang}, \bibinfo{person}{Xiangnan He}, {and} \bibinfo{person}{Tat{-}Seng Chua}.} \bibinfo{year}{2022}\natexlab{a}.
\newblock \showarticletitle{Discovering Invariant Rationales for Graph Neural Networks}. In \bibinfo{booktitle}{\emph{The Tenth International Conference on Learning Representations}}. \bibinfo{publisher}{OpenReview.net}.
\newblock


\bibitem[\protect\citeauthoryear{Wu, Pan, Chen, Long, Zhang, and Philip}{Wu et~al\mbox{.}}{2020b}]%
        {wu2020comprehensive}
\bibfield{author}{\bibinfo{person}{Zonghan Wu}, \bibinfo{person}{Shirui Pan}, \bibinfo{person}{Fengwen Chen}, \bibinfo{person}{Guodong Long}, \bibinfo{person}{Chengqi Zhang}, {and} \bibinfo{person}{S~Yu Philip}.} \bibinfo{year}{2020}\natexlab{b}.
\newblock \showarticletitle{A comprehensive survey on graph neural networks}.
\newblock \bibinfo{journal}{\emph{IEEE transactions on neural networks and learning systems}} \bibinfo{volume}{32}, \bibinfo{number}{1} (\bibinfo{year}{2020}), \bibinfo{pages}{4--24}.
\newblock


\bibitem[\protect\citeauthoryear{Xie, Zhu, Huang, Du, and Nie}{Xie et~al\mbox{.}}{2021}]%
        {xie2021graph}
\bibfield{author}{\bibinfo{person}{Qianqian Xie}, \bibinfo{person}{Yutao Zhu}, \bibinfo{person}{Jimin Huang}, \bibinfo{person}{Pan Du}, {and} \bibinfo{person}{Jian-Yun Nie}.} \bibinfo{year}{2021}\natexlab{}.
\newblock \showarticletitle{Graph neural collaborative topic model for citation recommendation}.
\newblock \bibinfo{journal}{\emph{ACM Transactions on Information Systems (TOIS)}} \bibinfo{volume}{40}, \bibinfo{number}{3} (\bibinfo{year}{2021}), \bibinfo{pages}{1--30}.
\newblock


\bibitem[\protect\citeauthoryear{Xu, Ruan, K{\"{o}}rpeoglu, Kumar, and Achan}{Xu et~al\mbox{.}}{2020}]%
        {xu2020inductive}
\bibfield{author}{\bibinfo{person}{Da Xu}, \bibinfo{person}{Chuanwei Ruan}, \bibinfo{person}{Evren K{\"{o}}rpeoglu}, \bibinfo{person}{Sushant Kumar}, {and} \bibinfo{person}{Kannan Achan}.} \bibinfo{year}{2020}\natexlab{}.
\newblock \showarticletitle{Inductive representation learning on temporal graphs}. In \bibinfo{booktitle}{\emph{8th International Conference on Learning Representations}}. \bibinfo{publisher}{OpenReview.net}.
\newblock


\bibitem[\protect\citeauthoryear{Yang, Zhou, Kalander, Huang, and King}{Yang et~al\mbox{.}}{2021b}]%
        {yang2021discrete}
\bibfield{author}{\bibinfo{person}{Menglin Yang}, \bibinfo{person}{Min Zhou}, \bibinfo{person}{Marcus Kalander}, \bibinfo{person}{Zengfeng Huang}, {and} \bibinfo{person}{Irwin King}.} \bibinfo{year}{2021}\natexlab{b}.
\newblock \showarticletitle{Discrete-time Temporal Network Embedding via Implicit Hierarchical Learning in Hyperbolic Space}. In \bibinfo{booktitle}{\emph{Proceedings of the 27th ACM SIGKDD Conference on Knowledge Discovery \& Data Mining}}. \bibinfo{pages}{1975--1985}.
\newblock


\bibitem[\protect\citeauthoryear{Yang, Ma, Zhang, Gao, Zhang, and Zhang}{Yang et~al\mbox{.}}{2023b}]%
        {yang2023interpretable}
\bibfield{author}{\bibinfo{person}{Qiang Yang}, \bibinfo{person}{Changsheng Ma}, \bibinfo{person}{Qiannan Zhang}, \bibinfo{person}{Xin Gao}, \bibinfo{person}{Chuxu Zhang}, {and} \bibinfo{person}{Xiangliang Zhang}.} \bibinfo{year}{2023}\natexlab{b}.
\newblock \showarticletitle{Interpretable Research Interest Shift Detection with Temporal Heterogeneous Graphs}. In \bibinfo{booktitle}{\emph{Proceedings of the Sixteenth ACM International Conference on Web Search and Data Mining}}. \bibinfo{pages}{321--329}.
\newblock


\bibitem[\protect\citeauthoryear{Yang, Hu, Shi, Ji, Li, and Nie}{Yang et~al\mbox{.}}{2021a}]%
        {yang2021hgat}
\bibfield{author}{\bibinfo{person}{Tianchi Yang}, \bibinfo{person}{Linmei Hu}, \bibinfo{person}{Chuan Shi}, \bibinfo{person}{Houye Ji}, \bibinfo{person}{Xiaoli Li}, {and} \bibinfo{person}{Liqiang Nie}.} \bibinfo{year}{2021}\natexlab{a}.
\newblock \showarticletitle{HGAT: Heterogeneous graph attention networks for semi-supervised short text classification}.
\newblock \bibinfo{journal}{\emph{ACM Transactions on Information Systems (TOIS)}} \bibinfo{volume}{39}, \bibinfo{number}{3} (\bibinfo{year}{2021}), \bibinfo{pages}{1--29}.
\newblock


\bibitem[\protect\citeauthoryear{Yang, Feng, Song, and Wang}{Yang et~al\mbox{.}}{2020}]%
        {yang2020factorizable}
\bibfield{author}{\bibinfo{person}{Yiding Yang}, \bibinfo{person}{Zunlei Feng}, \bibinfo{person}{Mingli Song}, {and} \bibinfo{person}{Xinchao Wang}.} \bibinfo{year}{2020}\natexlab{}.
\newblock \showarticletitle{Factorizable graph convolutional networks}.
\newblock \bibinfo{journal}{\emph{Advances in Neural Information Processing Systems}}  \bibinfo{volume}{33} (\bibinfo{year}{2020}), \bibinfo{pages}{20286--20296}.
\newblock


\bibitem[\protect\citeauthoryear{Yang, He, Zhang, Wu, Xin, Chen, and Wang}{Yang et~al\mbox{.}}{2023a}]%
        {yang2023generic}
\bibfield{author}{\bibinfo{person}{Zhengyi Yang}, \bibinfo{person}{Xiangnan He}, \bibinfo{person}{Jizhi Zhang}, \bibinfo{person}{Jiancan Wu}, \bibinfo{person}{Xin Xin}, \bibinfo{person}{Jiawei Chen}, {and} \bibinfo{person}{Xiang Wang}.} \bibinfo{year}{2023}\natexlab{a}.
\newblock \showarticletitle{A Generic Learning Framework for Sequential Recommendation with Distribution Shifts}. In \bibinfo{booktitle}{\emph{Proceedings of the 46th International ACM SIGIR Conference on Research and Development in Information Retrieval}}.
\newblock


\bibitem[\protect\citeauthoryear{Yao, Choi, Lee, Koh, and Finn}{Yao et~al\mbox{.}}{2022a}]%
        {yao2022wildtime}
\bibfield{author}{\bibinfo{person}{Huaxiu Yao}, \bibinfo{person}{Caroline Choi}, \bibinfo{person}{Yoonho Lee}, \bibinfo{person}{Pang~Wei Koh}, {and} \bibinfo{person}{Chelsea Finn}.} \bibinfo{year}{2022}\natexlab{a}.
\newblock \showarticletitle{Wild-Time: A Benchmark of in-the-Wild Distribution Shift over Time}. In \bibinfo{booktitle}{\emph{Proceedings of the Thirty-sixth Conference on Neural Information Processing Systems Datasets and Benchmarks Track}}.
\newblock


\bibitem[\protect\citeauthoryear{Yao, Wang, Li, Zhang, Liang, Zou, and Finn}{Yao et~al\mbox{.}}{2022b}]%
        {yao2022improving}
\bibfield{author}{\bibinfo{person}{Huaxiu Yao}, \bibinfo{person}{Yu Wang}, \bibinfo{person}{Sai Li}, \bibinfo{person}{Linjun Zhang}, \bibinfo{person}{Weixin Liang}, \bibinfo{person}{James Zou}, {and} \bibinfo{person}{Chelsea Finn}.} \bibinfo{year}{2022}\natexlab{b}.
\newblock \showarticletitle{Improving Out-of-Distribution Robustness via Selective Augmentation}. In \bibinfo{booktitle}{\emph{Proceeding of the Thirty-ninth International Conference on Machine Learning}}.
\newblock


\bibitem[\protect\citeauthoryear{You, Wang, Pal, Eksombatchai, Rosenburg, and Leskovec}{You et~al\mbox{.}}{2019}]%
        {you2019hierarchical}
\bibfield{author}{\bibinfo{person}{Jiaxuan You}, \bibinfo{person}{Yichen Wang}, \bibinfo{person}{Aditya Pal}, \bibinfo{person}{Pong Eksombatchai}, \bibinfo{person}{Chuck Rosenburg}, {and} \bibinfo{person}{Jure Leskovec}.} \bibinfo{year}{2019}\natexlab{}.
\newblock \showarticletitle{Hierarchical temporal convolutional networks for dynamic recommender systems}. In \bibinfo{booktitle}{\emph{The world wide web conference}}. \bibinfo{pages}{2236--2246}.
\newblock


\bibitem[\protect\citeauthoryear{Zhang, Li, Huang, Wu, Zhou, Yang, and Gao}{Zhang et~al\mbox{.}}{2022a}]%
        {zhang2022efraudcom}
\bibfield{author}{\bibinfo{person}{Ge Zhang}, \bibinfo{person}{Zhao Li}, \bibinfo{person}{Jiaming Huang}, \bibinfo{person}{Jia Wu}, \bibinfo{person}{Chuan Zhou}, \bibinfo{person}{Jian Yang}, {and} \bibinfo{person}{Jianliang Gao}.} \bibinfo{year}{2022}\natexlab{a}.
\newblock \showarticletitle{efraudcom: An e-commerce fraud detection system via competitive graph neural networks}.
\newblock \bibinfo{journal}{\emph{ACM Transactions on Information Systems (TOIS)}} \bibinfo{volume}{40}, \bibinfo{number}{3} (\bibinfo{year}{2022}), \bibinfo{pages}{1--29}.
\newblock


\bibitem[\protect\citeauthoryear{Zhang, Suzumura, and Zhang}{Zhang et~al\mbox{.}}{2021a}]%
        {zhang2021dyngraphtrans}
\bibfield{author}{\bibinfo{person}{Shilei Zhang}, \bibinfo{person}{Toyotaro Suzumura}, {and} \bibinfo{person}{Li Zhang}.} \bibinfo{year}{2021}\natexlab{a}.
\newblock \showarticletitle{DynGraphTrans: Dynamic Graph Embedding via Modified Universal Transformer Networks for Financial Transaction Data}. In \bibinfo{booktitle}{\emph{2021 IEEE International Conference on Smart Data Services (SMDS)}}. IEEE, \bibinfo{pages}{184--191}.
\newblock


\bibitem[\protect\citeauthoryear{Zhang, Zhang, Pfoser, and Zhao}{Zhang et~al\mbox{.}}{2021b}]%
        {zhang2021disentangled}
\bibfield{author}{\bibinfo{person}{Wenbin Zhang}, \bibinfo{person}{Liming Zhang}, \bibinfo{person}{Dieter Pfoser}, {and} \bibinfo{person}{Liang Zhao}.} \bibinfo{year}{2021}\natexlab{b}.
\newblock \showarticletitle{Disentangled dynamic graph deep generation}. In \bibinfo{booktitle}{\emph{Proceedings of the 2021 SIAM International Conference on Data Mining (SDM)}}. SIAM, \bibinfo{pages}{738--746}.
\newblock


\bibitem[\protect\citeauthoryear{Zhang, Wang, Zhang, Li, Qin, and Zhu}{Zhang et~al\mbox{.}}{2022b}]%
        {zhang2022dynamic}
\bibfield{author}{\bibinfo{person}{Zeyang Zhang}, \bibinfo{person}{Xin Wang}, \bibinfo{person}{Ziwei Zhang}, \bibinfo{person}{Haoyang Li}, \bibinfo{person}{Zhou Qin}, {and} \bibinfo{person}{Wenwu Zhu}.} \bibinfo{year}{2022}\natexlab{b}.
\newblock \showarticletitle{Dynamic graph neural networks under spatio-temporal distribution shift}. In \bibinfo{booktitle}{\emph{Advances in Neural Information Processing Systems}}.
\newblock


\bibitem[\protect\citeauthoryear{Zhang, Zhang, Wang, and Zhu}{Zhang et~al\mbox{.}}{2022c}]%
        {zhang2022learning}
\bibfield{author}{\bibinfo{person}{Zeyang Zhang}, \bibinfo{person}{Ziwei Zhang}, \bibinfo{person}{Xin Wang}, {and} \bibinfo{person}{Wenwu Zhu}.} \bibinfo{year}{2022}\natexlab{c}.
\newblock \showarticletitle{Learning to Solve Travelling Salesman Problem with Hardness-Adaptive Curriculum}. In \bibinfo{booktitle}{\emph{Thirty-Sixth {AAAI} Conference on Artificial Intelligence}}. \bibinfo{publisher}{{AAAI} Press}, \bibinfo{pages}{9136--9144}.
\newblock


\bibitem[\protect\citeauthoryear{Zhao, Zhu, Xu, Lizhiyu, Yu, Li, Yin, and Chen}{Zhao et~al\mbox{.}}{2023}]%
        {zhao2023time}
\bibfield{author}{\bibinfo{person}{Ziwei Zhao}, \bibinfo{person}{Xi Zhu}, \bibinfo{person}{Tong Xu}, \bibinfo{person}{Aakas Lizhiyu}, \bibinfo{person}{Yu Yu}, \bibinfo{person}{Xueying Li}, \bibinfo{person}{Zikai Yin}, {and} \bibinfo{person}{Enhong Chen}.} \bibinfo{year}{2023}\natexlab{}.
\newblock \showarticletitle{Time-interval Aware Share Recommendation via Bi-directional Continuous Time Dynamic Graphs}. In \bibinfo{booktitle}{\emph{Proceedings of the 46th International ACM SIGIR Conference on Research and Development in Information Retrieval}}. \bibinfo{pages}{822--831}.
\newblock


\bibitem[\protect\citeauthoryear{Zhou, Cui, Hu, Zhang, Yang, Liu, Wang, Li, and Sun}{Zhou et~al\mbox{.}}{2020}]%
        {zhou2020graph}
\bibfield{author}{\bibinfo{person}{Jie Zhou}, \bibinfo{person}{Ganqu Cui}, \bibinfo{person}{Shengding Hu}, \bibinfo{person}{Zhengyan Zhang}, \bibinfo{person}{Cheng Yang}, \bibinfo{person}{Zhiyuan Liu}, \bibinfo{person}{Lifeng Wang}, \bibinfo{person}{Changcheng Li}, {and} \bibinfo{person}{Maosong Sun}.} \bibinfo{year}{2020}\natexlab{}.
\newblock \showarticletitle{Graph neural networks: A review of methods and applications}.
\newblock \bibinfo{journal}{\emph{AI open}}  \bibinfo{volume}{1} (\bibinfo{year}{2020}), \bibinfo{pages}{57--81}.
\newblock


\bibitem[\protect\citeauthoryear{Zhou, Yang, Ren, Wu, and Zhuang}{Zhou et~al\mbox{.}}{2018}]%
        {zhou2018dynamic}
\bibfield{author}{\bibinfo{person}{Lekui Zhou}, \bibinfo{person}{Yang Yang}, \bibinfo{person}{Xiang Ren}, \bibinfo{person}{Fei Wu}, {and} \bibinfo{person}{Yueting Zhuang}.} \bibinfo{year}{2018}\natexlab{}.
\newblock \showarticletitle{Dynamic network embedding by modeling triadic closure process}. In \bibinfo{booktitle}{\emph{Proceedings of the AAAI conference on artificial intelligence}}, Vol.~\bibinfo{volume}{32}.
\newblock


\bibitem[\protect\citeauthoryear{Zhu, Ponomareva, Han, and Perozzi}{Zhu et~al\mbox{.}}{2021}]%
        {zhu2021shift}
\bibfield{author}{\bibinfo{person}{Qi Zhu}, \bibinfo{person}{Natalia Ponomareva}, \bibinfo{person}{Jiawei Han}, {and} \bibinfo{person}{Bryan Perozzi}.} \bibinfo{year}{2021}\natexlab{}.
\newblock \showarticletitle{Shift-robust gnns: Overcoming the limitations of localized graph training data}.
\newblock \bibinfo{journal}{\emph{Advances in Neural Information Processing Systems}}  \bibinfo{volume}{34} (\bibinfo{year}{2021}).
\newblock


\bibitem[\protect\citeauthoryear{Zhu, Lyu, Hu, Chen, and Liu}{Zhu et~al\mbox{.}}{2022}]%
        {zhu2022learnable}
\bibfield{author}{\bibinfo{person}{Yuecai Zhu}, \bibinfo{person}{Fuyuan Lyu}, \bibinfo{person}{Chengming Hu}, \bibinfo{person}{Xi Chen}, {and} \bibinfo{person}{Xue Liu}.} \bibinfo{year}{2022}\natexlab{}.
\newblock \showarticletitle{Learnable Encoder-Decoder Architecture for Dynamic Graph: A Survey}.
\newblock \bibinfo{journal}{\emph{arXiv preprint arXiv:2203.10480}} (\bibinfo{year}{2022}).
\newblock


\bibitem[\protect\citeauthoryear{Zitnik, Agrawal, and Leskovec}{Zitnik et~al\mbox{.}}{2018}]%
        {zitnik2018modeling}
\bibfield{author}{\bibinfo{person}{Marinka Zitnik}, \bibinfo{person}{Monica Agrawal}, {and} \bibinfo{person}{Jure Leskovec}.} \bibinfo{year}{2018}\natexlab{}.
\newblock \showarticletitle{Modeling polypharmacy side effects with graph convolutional networks}.
\newblock \bibinfo{journal}{\emph{Bioinformatics}} \bibinfo{volume}{34}, \bibinfo{number}{13} (\bibinfo{year}{2018}), \bibinfo{pages}{i457--i466}.
\newblock


\bibitem[\protect\citeauthoryear{Zitnik, Sosi{\v{c}}, Feldman, and Leskovec}{Zitnik et~al\mbox{.}}{2019}]%
        {zitnik2019evolution}
\bibfield{author}{\bibinfo{person}{Marinka Zitnik}, \bibinfo{person}{Rok Sosi{\v{c}}}, \bibinfo{person}{Marcus~W Feldman}, {and} \bibinfo{person}{Jure Leskovec}.} \bibinfo{year}{2019}\natexlab{}.
\newblock \showarticletitle{Evolution of resilience in protein interactomes across the tree of life}.
\newblock \bibinfo{journal}{\emph{Proceedings of the National Academy of Sciences}} \bibinfo{volume}{116}, \bibinfo{number}{10} (\bibinfo{year}{2019}), \bibinfo{pages}{4426--4433}.
\newblock


\end{thebibliography}
\end{document}